\documentclass[authorversion,sigconf]{acmart}

\copyrightyear{2022}
\acmYear{2022}
\setcopyright{acmlicensed}\acmConference[CCS '22]{Proceedings of the 2022 ACM SIGSAC Conference on Computer and Communications Security}{November 7--11, 2022}{Los Angeles, CA, USA}
\acmBooktitle{Proceedings of the 2022 ACM SIGSAC Conference on Computer and Communications Security (CCS '22), November 7--11, 2022, Los Angeles, CA, USA}
\acmPrice{15.00}
\acmDOI{10.1145/3548606.3560709}
\acmISBN{978-1-4503-9450-5/22/11}

\usepackage{algorithm}  
\usepackage[noend]{algpseudocode}  %

\usepackage[utf8]{inputenc} %
\usepackage{lipsum} %

\usepackage[T1]{fontenc}

\usepackage{amsmath,amsfonts,amsthm}

\ifdefined\isoverfull
\else
\fi

\usepackage{etoolbox}

\newbool{includeappendix}
\setbool{includeappendix}{true} %
\IfFileExists{headers/config/noappendix.config}{
	\setbool{includeappendix}{false}
}{}

\newif\ifincludeappendixx
\ifbool{includeappendix}{
	\includeappendixxtrue
}{
	\includeappendixxfalse
}

\usepackage{xr} %
\usepackage{filecontents}

\ifbool{includeappendix}{}{

	\externaldocument{appendix-labels}
}

\newcommand{\eg}{e.g., }
\newcommand{\ie}{i.e., }
\newcommand{\etc}{etc.}
\newcommand{\st}{s.t.\ }

\newcommand{\wrt}{{w.r.t.\ }}

\definecolor{my-full-blue}{HTML}{1F77B4}

\definecolor{my-full-orange}{HTML}{FF7F0E}

\definecolor{my-full-green}{HTML}{2CA02C}

\definecolor{my-full-red}{HTML}{d62728}

\definecolor{my-full-purple}{HTML}{9467bd}

\colorlet{my-blue}{my-full-blue!30}
\colorlet{my-orange}{my-full-orange!30}
\colorlet{my-green}{my-full-green!30}
\colorlet{my-red}{my-full-red!30}
\colorlet{my-purple}{my-full-purple!30}

\usepackage{listings}

\usepackage{textcomp}

\usepackage{xcolor}

\usepackage[scaled=0.8]{beramono}

\definecolor{ckeyword}{HTML}{7F0055}
\definecolor{ccomment}{HTML}{3F7F5F}
\definecolor{cstring}{HTML}{2A0099}

\lstdefinestyle{numbers}{
	numbers=left,
	framexleftmargin=20pt,
	numberstyle=\tiny,
	firstnumber=auto,
	numbersep=1em,
	xleftmargin=2em
}

\lstdefinestyle{layout}{
	frame=none,
	captionpos=b,
}

\lstdefinestyle{comment-style}{
	morecomment=[l]//,
	morecomment=[s]{/*}{*/},
	commentstyle={\color{ccomment}\itshape},
}

\lstdefinestyle{string-style}{
	morestring=[b]",%
	morestring=[b]',%
	stringstyle={\color{cstring}},
	showstringspaces=false,%
}

\lstdefinestyle{keyword-style}{
	keywordstyle={\ttfamily\bfseries},
	morekeywords={
		function,
		constructor,
		int,
		bool,
		return,
		returns,
		uint
	},
	morekeywords = [2]{},
	keywordstyle = [2]{\text},
	sensitive=true,
}

\lstdefinestyle{input-encoding}{
	inputencoding=utf8,
	extendedchars=true,
	literate=
	{ℝ}{$\reals$}1%
	{→}{$\rightarrow$}1%
	{α}{$\alpha$}1%
	{β}{$\beta$}1%
	{λ}{$\lambda$}1%
	{θ}{$\theta$}1%
	{ϕ}{$\phi$}1%
}

\lstdefinestyle{escaping}{
	moredelim={**[is][\color{blue}]{\%}{\%}},
	escapechar=|,
	mathescape=true
}

\lstdefinestyle{default-style}{
	basicstyle=\fontencoding{T1}\ttfamily\footnotesize,
	style=numbers,
	style=layout,
	style=comment-style,
	style=string-style,
	style=keyword-style,
	style=input-encoding,
	style=escaping,
	tabsize=2,
	upquote=true
}

\lstdefinelanguage{BASIC}{
	language=C++,
	style=default-style
}[keywords,comments,strings]%

\usepackage{csvsimple}

\usepackage[capitalize]{cleveref}

\makeatletter
\renewcommand\theHALG@line{\thealgorithm.\arabic{ALG@line}}
\makeatother

\crefformat{section}{\S#2#1#3}

\crefrangeformat{section}{\S#3#1#4\crefrangeconjunction\S#5#2#6}

\crefmultiformat{section}{\S#2#1#3}{\crefpairconjunction\S#2#1#3}{\crefmiddleconjunction\S#2#1#3}{\creflastconjunction\S#2#1#3}

\newcommand{\crefrangeconjunction}{--}

\crefname{theorem}{Thm.}{Thms.}

\crefname{algorithm}{Alg.}{Algs.}
\crefname{listing}{Lst.}{listings}
\crefname{line}{Line}{Lines}
\crefname{appendix}{App.}{App.}

\newbool{reftoextendeddone}
\setbool{reftoextendeddone}{false}

\newcommand{\reftoextended}{%
	\ifbool{reftoextendeddone}{\textsuperscript{\ref{footnote:extendedversion}}}{%
		\footnote{\label{footnote:extendedversion}Available in the extended version of this paper:\\ \url{\extendedpaperlink}}%
		\setbool{reftoextendeddone}{true}%
	}%
}
\newcommand{\apprefrange}[2]{%
	\ifbool{includeappendix}{\crefrange{#1}{#2}}{App.\reftoextended{}~\hyperref[footnote:extendedversion]{\ref*{#1}}--\hyperref[footnote:extendedversion]{\ref*{#2}}}%
}
\newcommand{\appref}[1]{%
	\ifbool{includeappendix}{\cref{#1}}{App.\reftoextended{}~\hyperref[footnote:extendedversion]{\ref*{#1}}}%
}
\newcommand{\Appref}[1]{%
	\ifbool{includeappendix}{\cref{#1}}{App.\reftoextended{}~\hyperref[footnote:extendedversion]{\ref*{#1}}}%
}

\usepackage{graphicx}
\usepackage{mathtools}%
\usepackage{enumitem}
\usepackage{bbm}    %
\usepackage{xspace}
\usepackage{textpos}	%
\usepackage[margin=4pt]{subcaption}	%

\usepackage{amsmath,amsfonts,bm}
\usepackage{amsthm}

\newtheorem{theorem}{Theorem}

\def\1{\bm{1}}

\DeclareMathAlphabet{\mathsfit}{\encodingdefault}{\sfdefault}{m}{sl}
\SetMathAlphabet{\mathsfit}{bold}{\encodingdefault}{\sfdefault}{bx}{n}

\DeclareMathOperator*{\argmax}{arg\,max}

\makeatletter
\renewcommand{\paragraph}{\textbf} 
    \def\@IEEEsectpunct{.\ }

    \def\parax{\@startsection%
        {paragraph}%
        {4}%
        {0\parindent}%
        {0.6ex plus 0.1ex minus 0.1ex}%
        {0ex}%
        {\normalfont\normalsize\itshape\bfseries}%
        *%
    }%
\makeatother

\newcommand{\para}[1]{\parax{#1}.\ } %

\usepackage{booktabs}
\usepackage[normalem]{ulem}

\newcommand{\enc}[1]{\ensuremath{\text{\texttt{Enc}}_{pk}({#1})}}
\newcommand{\dec}[1]{\ensuremath{\text{\texttt{Dec}}_{sk}({#1})}}

\newcommand{\msp}{\ensuremath{\mathcal{M}}}
\newcommand{\psp}{\ensuremath{\mathcal{P}}}
\newcommand{\csp}{\ensuremath{\mathcal{C}}}

\newcommand{\trash}{\ensuremath{\#}}

\usepackage{stackengine} 
\newcommand\oast{\stackMath\mathbin{\stackinset{c}{0ex}{c}{0ex}{\ast}{\bigcirc}}}

\newcommand{\indicator}[1]{\mathbbm{1}_{#1}}

\newcommand{\tool}{\textsc{Phoenix}\xspace}

\newcommand{\toolR}{\tool{}\textsc{R}\xspace}
\newcommand{\toolF}{\tool{}\textsc{F}\xspace}

\newcommand{\fhe}{FHE\xspace}
\newcommand{\Argmax}{\textsc{Argmax}\xspace}
\newcommand{\ArgmaxHE}{\textsc{ArgmaxHE}\xspace}

\newcommand{\lock}[0]{\mbox{\includegraphics[height=0.25cm]{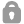}}\xspace}

\newenvironment{block}%
  {\vspace{1em}\list{}{\leftmargin=1em\rightmargin=1em}  \item[]  }%
	  {\endlist\vspace{1em}}

\usepackage{multirow}

\usepackage{balance}

\newcommand{\extendedpaperlink}{https://www.sri.inf.ethz.ch/publications/jovanovic2022phoenix}

\begin{document}

\title{Private and Reliable Neural Network Inference}

\author{Nikola Jovanović}
\email{nikola.jovanovic@inf.ethz.ch}
\affiliation{%
\institution{ETH Zurich}
\country{Switzerland}
}

\author{Marc Fischer}
\email{marc.fischer@inf.ethz.ch}
\affiliation{%
\institution{ETH Zurich}
\country{Switzerland}
}

\author{Samuel Steffen}
\email{samuel.steffen@inf.ethz.ch}
\affiliation{%
\institution{ETH Zurich}
\country{Switzerland}
}

\author{Martin Vechev}
\email{martin.vechev@inf.ethz.ch}
\affiliation{%
\institution{ETH Zurich}
\country{Switzerland}
}

\begin{abstract}
  Reliable neural networks (NNs) provide important inference-time reliability
guarantees such as fairness and robustness. Complementarily, privacy-preserving
NN inference protects the privacy of client data. So far these two emerging
areas have been largely disconnected, yet their combination will be increasingly
important.

In this work, we present the first system which enables privacy-preserving
inference on reliable NNs. Our key idea is to design efficient fully homomorphic
encryption (FHE) counterparts for the core algorithmic building blocks of
randomized smoothing, a state-of-the-art technique for obtaining reliable
models. The lack of required control flow in FHE makes this a demanding task, as
naïve solutions lead to unacceptable runtime.

We employ these building blocks to enable privacy-preserving NN inference with
robustness and fairness guarantees in a system called \tool. Experimentally, we
demonstrate that \tool achieves its goals without incurring prohibitive
latencies.

To our knowledge, this is the first work which bridges the areas of client data
privacy and reliability guarantees for NNs.

\end{abstract}

\ccsdesc[500]{Security and privacy~Privacy-preserving protocols}

\keywords{reliable machine learning; fully homomorphic encryption} 

\maketitle

\section{Introduction} \label{sec:intro}

In the \emph{Machine Learning as a Service} (MLaaS) \cite{GraepelLN12} setting,
a \emph{client} offloads the intensive computation required to perform inference
to an external \emph{server}. In practice, this inference is often performed on
sensitive data (\eg financial or medical records). However, as sending
unencrypted sensitive data to the server imposes major privacy risks on the
client, a long line of recent work in privacy-preserving machine
learning~\cite{GiladDLLNW16,ChouBLYHF18,BrutzkusGE19,LouJ19,LiuJLA17,JuvekarVC18,MishraLSZP20}
aims to protect the privacy of client data.

\para{Client Data Privacy using Encryption}
Typically, these methods rely on fully homomorphic encryption
(\fhe)~\cite{Gentry09}, which allows the evaluation of rich functions on
encrypted values (we discuss alternative approaches based on secure multiparty
computation and their drawbacks in~\cref{sec:related}). In particular, the
client encrypts their sensitive input $x$ using an \fhe scheme and a public
key~$pk$, and sends $\texttt{Enc}(x, pk)$ to the server. The server then
evaluates its model $f$ using \fhe capabilities to obtain the encrypted
prediction $\texttt{Enc}(f(x), pk)$, which is finally sent back to the client,
who can decrypt it using the corresponding secret key $sk$ to obtain the
prediction $f(x)$. As a result, the computation does not leak sensitive data to
the server.

FHE is becoming more practically viable and has already been introduced to
real-world systems \cite{SikeridisPRD17,MastersHSCB19,Barua21}. The reluctance
of clients to reveal valuable data to servers~\cite{Gill21,Williams21} has
recently been complemented by increasing legal obligations to maintain privacy
of customer data \cite{GDPR, CCPA}. Hence, it is expected that service
providers will be increasingly required to facilitate client data privacy.

\para{Reliable Machine Learning}
While the use of FHE to achieve inference-time client data privacy is well
understood, an independent and emerging line of work in (non-private) machine
learning targets building \emph{reliable} models that come with important
{guarantees}~\cite{DworkHPRZ12,yeom2020individual,RuossBFV20,GehrMDCTCV18,GowalDSBQJMK18,CohenRK19}. For example,
\mbox{\emph{fairness}}~\cite{DworkHPRZ12,yeom2020individual,RuossBFV20} ensures
that the model does not discriminate, a major concern when inputs represent
humans \eg in judicial, financial or recruitment applications. Similarly,
\emph{robustness}~\cite{GehrMDCTCV18,GowalDSBQJMK18,CohenRK19} ensures the model
is stable to small variations of the input (\eg different lighting conditions in
images)---a key requirement for critical use-cases often found in medicine such
as disease prediction.
Indeed, the importance of obtaining machine learning models with guarantees that
ensure societal accountability~\cite{Balakrishnan20} will only increase.

Given the prominence of these two emerging directions, data privacy and reliability, 
a natural question which arises is:

\begin{block}
	\emph{How do we retain reliability guarantees when transitioning machine learning models to a privacy-preserving setting?}
\end{block}

Addressing this question is difficult: to provide the desired reliability
guarantees, such models often rely on additional work performed at inference
time which cannot be directly lifted to state-of-the-art \fhe schemes. In
particular, the missing native support for control flow (\eg branching) and
evaluation of non-polynomial functions (\eg \Argmax) is a key challenge, as
naïve workarounds impose prohibitive computational overhead \cite{CheonKK20}.

\begin{figure}[t]
	\centering
	\includegraphics[height=2.8cm]{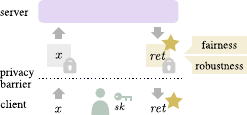}
	\begin{textblock}{2}(1.84,-1.52) \tool (\cref{fig:overview})
    \end{textblock}
	\caption{A high-level overview of private and reliable inference in an \fhe-based MLaaS setting using our system, \tool.}
	\label{fig:main}
\end{figure}

\para{This Work: Prediction Guarantees with Client Data Privacy}
We present \tool, which for the first time addresses the above question in a
comprehensive manner, illustrated in \cref{fig:main}. \tool enables
FHE-execution of \emph{randomized smoothing} (RS) \cite{CohenRK19}, a
state-of-the-art method for obtaining reliability guarantees on top of neural
network inference. \tool lifts the algorithmic building blocks of RS to \fhe,
utilizing recent advances in non-polynomial function approximation to solve the
challenging task of executing the \Argmax function in a private \fhe-based
setting.
Making progress on this challenge in the context of RS is significant as (i) RS
naturally enables guarantees such as fairness and robustness
(\cref{ssec:overview:pipeline}), and (ii) the core building blocks are general
and provide value outside of the RS context for establishing broader guarantees
(\cref{ssec:overview:guarantees}).

\begin{figure*}[t]
    \centering
    \includegraphics[width=\textwidth]{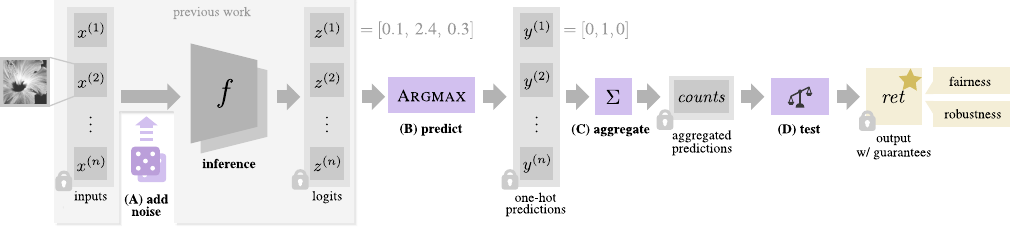}
    \caption{An illustration of \tool, which enables 
	     homomorphically encrypted neural network inference with guarantees. The
	     symbol~\protect\lock denotes encryption of the value in an \fhe scheme
	     by the client.}
    \label{fig:overview}
\end{figure*}

\para{Main Contributions} Our main contributions are:
\begin{itemize}[leftmargin=12pt]
    \item \ArgmaxHE, an efficient and precise approximation of the challenging
    \Argmax operator in RNS-CKKS \fhe \cite{CheonHKKS18Rns}
    (\cref{sec:argmaxhe}).
    \item \tool\footnote{Our implementation is available at: \url{https://github.com/eth-sri/phoenix}}, which by utilizing \ArgmaxHE and other novel building blocks,
    for the first time enables \fhe-based MLaaS with useful robustness and
    fairness guarantees. (\cref{sec:smoothing}, \cref{sec:error}).
    \item An extensive experimental evaluation of both instantiations of \tool,
    implemented in Microsoft SEAL~\cite{SEALCrypto} (\cref{sec:eval}).
\end{itemize} 

\section{Motivation} \label{sec:motivation}

We discuss the motivation for \tool, introducing examples that
illustrate the need for reliability guarantees, and motivate the importance of
obtaining such guarantees alongside client data privacy.

\para{Example: Fair Loan Eligibility} Consider an ML system deployed by a financial
institution in an MLaaS setting to automatically predict loan eligibility based
on clients' financial records. Recent work
\cite{BuolamwiniG18,CorbettPFGH17,KleinbergMR17,TatmanK17,DworkHPRZ12}
demonstrates that without any countermeasures, such a model may replicate biases
in training data and make unfair decisions. For example, a loan application
might be rejected based on a sensitive attribute of the client (\eg race, age,
or gender), or there could be pairs of individuals with only minor differences
who unjustly receive different outcomes (see \cref{ssec:background:fairness} for
more details on these two interpretations of fairness---\tool focuses on the
latter). Indeed, existing systems were shown to manifest unfair behavior in
practice. For instance, the recidivism prediction algorithm COMPAS was found to
unfairly treat black people~\cite{Angwin16}. Ideally, the loan eligibility
service would provide \emph{fairness guarantees}, ensuring that every prediction
for a client's input is fair. In case no fair decision can be made
automatically, the model refers the client to a human contact for manual
assessment. Providing such a guarantee greatly improves the accountability of
the system for both parties.
 
\para{Example: Robust Medical Image Analysis} Consider the problem of medical
image analysis, where patient data (MRI, X-ray or CT scans) is used for cancer
diagnosis, retinopathy detection, lung disease classification, \etc~
\cite{Apostolidis21}. ML algorithms are increasingly used for such applications
to reduce the worryingly high rate of medical error \cite{Makary16}. Such an
algorithm, deployed as a cloud service, may be used by medical providers to
assist in diagnosis, or even by patients as a form of second opinion. However,
it was observed that such algorithms are particularly non-robust \cite{Ma21} to
slight perturbations of input data, which could be due to a malicious actor, \eg
patients tampering with their reports to extract compensation \cite{Paschali18},
or naturally-occurring measurement errors. As such perturbations may lead to
needless costs or negative effects on patient's health \cite{Mangaokar20}, it is
in the interest of both parties that the cloud service provides \emph{robustness
guarantees}. That is, the returned diagnosis is guaranteed to be robust to input
data variations and can thus be deemed trustworthy. See
\cref{ssec:background:robustness} for a more formal discussion.

\para{Reliability Guarantees Alongside Data Privacy} As the above examples
illustrate, reliability guarantees can greatly benefit MLaaS deployments,
prompting regulators to call for their use \cite{AAMIBSI, FdaReg, EuReg}, and
researchers to actively investigate this area
\cite{DworkHPRZ12,yeom2020individual,RuossBFV20,GehrMDCTCV18,GowalDSBQJMK18,CohenRK19}.
However, ensuring reliable predictions does not address the important
complementary concern of \emph{client data privacy}, which becomes desirable and
commonly required \cite{GDPR, CCPA} when dealing with sensitive user
data, as is often the case in MLaaS setups. Latest privacy-preserving machine
learning
\cite{GiladDLLNW16,ChouBLYHF18,BrutzkusGE19,LouJ19,LiuJLA17,JuvekarVC18,MishraLSZP20}
methods such as FHE enable clients to utilize \eg loan eligibility and medical
diagnosis services without directly revealing their sensitive financial records
(\eg revenue) or medical imaging to the potentially untrusted service provider.
While desirable, achieving both of these goals at the same time is difficult:
methods that provide reliability guarantees often involve additional
inference-time logic, which makes lifting computation to FHE in order to protect
client data privacy significantly more challenging. \tool for the first time
takes a major step in addressing this challenge, enabling the deployment of
models which are both reliable and private. 

\section{Overview} \label{sec:overview}

We provide a high-level outline of \tool (\cref{ssec:overview:pipeline}),
presenting all algorithmic building blocks needed to enable reliability
guarantees via randomized smoothing. Further, we discuss alternative
instantiations of \tool that enable different properties
(\cref{ssec:overview:guarantees}), its threat model
(\cref{ssec:overview:threat-model}), and why it needs to be deployed on the
server (\cref{ssec:overview:server}).

\subsection{Overview of \tool}
\label{ssec:overview:pipeline}

\tool (\cref{fig:overview}) leverages randomized smoothing to produce ML models
with inference-time reliability guarantees. Key to enabling this in \fhe is to
combine existing techniques for evaluating neural networks in \fhe (marked
``previous work'' in \cref{fig:overview}) with new \fhe counterparts which
handle additional inference-time operations (indicated by (A)--(D) in
\cref{fig:overview}).
In that way, \tool also enables \fhe-execution of any future procedure which
relies on one or more of these fundamental components (as we discuss shortly in
\cref{ssec:overview:guarantees}).

\para{Inference}
The first ingredient is \emph{neural network inference}: Consider a neural
network $f$ classifying an input  $x^{(i)}$ into one of $c$ classes. Concretely,
$f$ outputs a vector of unnormalized scores $z^{(i)}$, called \emph{logits}, and
$f$ is said to classify $x^{(i)}$ to the class with the largest logit.

In the private MLaaS setting, the inputs $x^{(i)}$ are encrypted by the client
using an \fhe scheme (denoted by \lock). Ignoring the addition of noise
(discussed shortly), the inputs are propagated through $f$ utilizing the
homomorphic operations of the \fhe scheme to obtain the logits~$z^{(i)}$. In
\cref{fig:overview}, this is illustrated for $c=3$ where $x^{(1)}$ is classified
to the second class with logit value $z^{(1)}_2 = 2.4$. See
\cref{ssec:background:fhenn} for a more detailed discussion of neural network
inference in \fhe.

\para{Randomized Smoothing}
A core paradigm employed by \tool is \emph{randomized smoothing} (RS, detailed
in \cref{ssec:background:smoothing}). To obtain reliability guarantees, we
\emph{add Gaussian random noise (A)} to many copies of a single input $x^{(1)} =
\dots = x^{(n)}$, and perform neural network inference on these noisy copies to
get a set of logit vectors $z^{(i)}$.

Existing privacy-preserving machine learning systems would typically directly
return the logits~$z^{(i)}$ to the client for decryption
\cite{GiladDLLNW16,BrutzkusGE19,DathathriSCLLMMM19,LouJ21}. However, to apply
RS, \tool includes a \emph{predict step (B)}, which evaluates the \Argmax
function in \fhe. In particular, this step transforms the logit vectors
$z^{(i)}$ of unnormalized scores into \emph{one-hot} vectors $y^{(i)}$ of size
$c$, where a single non-zero entry $y^{(i)}_j$ encodes the index $j$ of the
largest component $z_j^{(i)}$. For example, predicting the second class is
represented by $[0, 1, 0]$. 
  
Next, we \emph{aggregate the predictions (C)} in a statistically sound way
(details given in \cref{sec:smoothing}) to determine the index $k$ of the class
most often predicted by the network, as well as the number of such predictions.

Finally, we perform a homomorphic version of a carefully calibrated
\emph{statistical test (D)}, to obtain $\mathit{ret}$, the ciphertext returned to the client, containing the final prediction with reliability guarantees.

\para{Obtaining Specific Guarantees} Depending on how it is employed, RS is able
to produce different guarantees. In \cref{ssec:smoothing:smoothing}, we discuss
\emph{local robustness} (\toolR), i.e., guaranteeing that minor input
perturbations such as slight changes of lighting conditions for images do not
change the model prediction. Further, in \cref{ssec:smoothing:fairness} we adapt
RS to support inference-time \emph{individual fairness} guarantees (\toolF),
ensuring that similar individuals are treated similarly. 

\subsection{Further Instantiations} \label{ssec:overview:guarantees}

The building blocks (A--D) of RS enabled by \tool are general. Thus, \tool can
also be instantiated in configurations beyond those we consider. In particular,
depending on the number of models used for inference, which subset of
predictions is aggregated, and where noise is added, the output may satisfy
different properties.

Namely, variants of RS have been proposed for various scenarios
\citep{SalmanLRZZBY19, FischerBV20, YangDHSRL20, ChingCAKDG20, FischerBV21,
KumarLF21, BojchevskiKG20, Peychev21}. For example, \citet{Peychev21} combines
RS with generative models to achieve provably fair representation learning, and
\citet{BojchevskiKG20} considers robust reasoning over graphs in the presence of
perturbations on edge weights. To migrate these and similar techniques to \fhe,
the building blocks (A--D) can likely be reused with slight adaptions.

Further, \tool could be instantiated to enhance the privacy properties of
PATE~\cite{PapernotAEGT17}, a state-of-the-art differentially private
learning~\cite{DworkMNS06} framework. In particular, executing PATE in an
FHE-based private setup would preserve the confidentiality of the unlabeled
dataset, making PATE more broadly applicable to scenarios where this data is
private and sensitive. On a technical level, PATE uses an ensemble of models at
the inference step, adding noise~(A) to their aggregated predictions (B--C).
This makes \tool particularly suitable for the task of lifting it to FHE.

Finally, as another example, addition of noise (A) and \Argmax (B) are also
fundamental components of the uncertainty quantification approach proposed by
\citet{Wang19}. 

We leave extending \tool to these settings to future work.

\subsection{Threat Model and Security Notion}\label{ssec:overview:threat-model}

In this work, we consider a combination of two standard threat models: the
two-party semi-honest attacker model from private NN inference (attackers
(i)--(ii) below), and the standard active attacker model from robustness
(attacker (iii) below). In particular, we consider three adversaries: (i) a
semi-honest attacker on the server-side, which tries to learn the plaintext
client input data $x$; (ii)~a semi-honest attacker sitting on the network
between client and server, also trying to learn~$x$; and (iii) an active
adversary on the client-side, which tries to imperceptibly corrupt client's data
$x$. As we will discuss in \cref{ssec:background:robustness}, adversary (iii)
models both naturally occurring measurement errors and deliberate tampering.

The core security notion achieved by \tool is \emph{client data privacy} (see
\cref{thm:privacy} in \cref{ssec:error:privacy}). In particular, we do not aim
to achieve the orthogonal notions of training set privacy or model privacy,
which are concerned with maintaining the privacy of server's data at training
time or maintaining privacy of model data at inference time. Note that attacker
(iii) is irrelevant for ensuring client data privacy. In addition, \tool
satisfies reliability guarantees (see~\cref{thm:reliability} in
\cref{ssec:error:reliability}); only attacker (iii) is relevant for these
properties.

\subsection{Necessity of Computing on the Server}
\label{ssec:overview:server}

\tool executes RS fully on the server. Due to the active client-side adversary
((iii) in \cref{ssec:overview:threat-model}), offloading RS to clients
would make it impossible for the service provider to guarantee reliability, as
such guarantees rely on careful parameter choices and sound execution of the
certification procedure.
For example, in the case of fairness, setting unsuitable parameters of RS on the
client side (maliciously or by accident, both of which are captured by the
active client-side adversary) would allow the clients to use the service to
obtain unfair predictions, nullifying the efforts of the service provider to
prove fairness compliance to external auditors. See
\cref{ssec:error:reliability} for a thorough correctness analysis of guarantees
produced by \tool.

An additional privacy benefit of server-side computation, orthogonal to client
data privacy, is in reducing the threat of \emph{model stealing
attacks}~\cite{TramerZJRR16,ReithST19}, where clients aim to recover the details
of the proprietary model $f$ given the logit vector $z$ of unnormalized scores.
Most prior work on neural network inference in FHE is vulnerable to such attacks
as it returns the encryption of $z$ to the client
\cite{GiladDLLNW16,BrutzkusGE19,DathathriSCLLMMM19,LouJ21}, who then needs to
apply \Argmax on the decrypted vector to obtain the prediction. Merely
approximating the score normalization, \ie executing the softmax function in
FHE~\cite{LeeKLCEDLLYKN21}, was shown to not offer meaningful
protection~\cite{TruongMWP20}. In contrast, applying the approximation of
\Argmax to $z$ in \fhe on the server (as done in \tool) allows the server to
return a one-hot prediction vector to the client. As illustrated by the model
access hierarchy of \citet{JagielskiCBKP20}, this greatly reduces the risk of
model stealing.

Finally, client-side RS would increase the communication cost, as well as the
effort for transitioning existing systems to inference with guarantees or later
modifying the parameters or offered guarantees.

\section{Background} \label{sec:background}

We next introduce the background necessary to present \tool. In
\cref{ssec:background:fhe} and \cref{ssec:background:fhenn} we recap fully
homomorphic encryption and how it is commonly used for privacy-preserving neural
network inference. In \cref{ssec:background:robustness} and
\cref{ssec:background:fairness} we formally introduce the two types of
reliability guarantees considered in \tool: local robustness and individual
fairness. Finally, in \cref{ssec:background:smoothing} we discuss randomized
smoothing, a technique commonly used to provide such guarantees. \tool lifts
this technique to fully homomorphic encryption to ensure predictions enjoy both
client data privacy as well as reliability guarantees.

\subsection{Leveled Fully Homomorphic Encryption with RNS-CKKS}
\label{ssec:background:fhe}

An asymmetric encryption scheme with public key $pk$ and secret key $sk$ is
\emph{homomorphic} \wrt some operation $\ast$ if there exists an operation
$\oast$, efficiently computable without knowing $sk$, \st for each pair of
messages $x, y \in \mathcal{M}$, where $\mathcal{M}$ is a field, it holds that
\[
    \dec{\enc{x} \oast \enc{y}} = x \ast y.
\]
Here, for a ciphertext space $\csp$, the functions 
\mbox{$\texttt{Enc}_{pk}\colon \msp \to \csp$} and
\mbox{$\texttt{Dec}_{sk}\colon \csp \to \msp$} denote the (often randomized)
encryption using $pk$, and decryption using $sk$, respectively. \emph{Fully homomorphic encryption (FHE)} schemes
\cite{Gentry09,BrakerskiGV12,FanV12,CheonKKS17,DucasM15,ChillottiGGI16,ChillottiJP21,CheonHKKS18Rns}
are those homomorphic \wrt all field operations in~$\mathcal{M}$. As such,
FHE can be used to evaluate arbitrary arithmetic circuits (\ie computations
of fixed size and structure) on underlying plaintexts. Note that as such
circuits do not allow for control flow (\eg branching or loops), reflecting
arbitrary computation in FHE can be challenging and/or very expensive.

\para{RNS-CKKS} In this work, we focus on RNS-CKKS~\cite{CheonHKKS18Rns}---an
efficient implementation of the CKKS~\cite{CheonKKS17} FHE scheme.
The plaintexts and ciphertexts of RNS-CKKS consist of elements of the polynomial
ring of degree $N$ with integer coefficients modulo $Q = \prod_{l=0}^L Q_l$,
where $Q_l$ are distinct primes, and $N$ and $L$ are parameters. A
\emph{message} is the atomic unit of information that can be represented in the
scheme. We consider the message space $\msp = \mathbb{R}^M$ (technically,
$\mathbb{R}$ is represented as fixed-point numbers), where $M=\frac{N}{2}$
denotes the number of so-called \emph{slots} in a message. To encrypt, all slots
in a message $x \in \msp$ are multiplied by the initial scale $\Delta \in
\mathbb{R}$, and the result is encoded to get a plaintext $\texttt{Ecd}(x) \in
\psp$, and then encrypted using a public key $pk$ to obtain a ciphertext
$\enc{x} \in \csp$.

Below, we list the homomorphic evaluation primitives of RNS-CKKS. Here, all
binary operations ${\oast \colon \csp \times \csp \to \csp}$ have a variant
${\oast_p \colon \csp \times \msp \to \csp}$, where one argument is a
plaintext.
 
\begin{itemize} 
    \item \emph{Addition/subtraction}: $\oplus$ and $\ominus$ perform slot-wise
    (vectorized) addition of the messages underlying the operands.
    \item \emph{Multiplication}: $\otimes$ performs slot-wise multiplication.
    \item \emph{Rotation}: \texttt{RotL}: $\csp \times \mathbb{N} \to \csp$ and
    \texttt{RotR}: $\csp \times \mathbb{N} \to \csp$ cyclically rotate the slots
    of the underlying message by the given number of steps left and
    right, respectively.
\end{itemize}
For example, for $M = 3$ and $\msp = \mathbb{R}^3$, it is
\[
    \texttt{RotR}(\enc{[a, b, c]}, 1) \otimes_p [2, 3, 1]\ =\ \enc{[2c, 3a,
b]}.
\]

RNS-CKKS is a \emph{leveled} FHE scheme: only computation up to multiplicative
depth $L$ is supported. That is, any computation path can include at most $L$ 
multiplications $\otimes$ or $\otimes_p$ (but arbitrarily many other
operations). Bootstrapping~\cite{Gentry09,CheonHKKS18Bootstrap} is designed to
overcome this limitation, however, it is still largely impractical
\cite{LeeKLCEDLLYKN21, LeeLLKKNC21}, so the RNS-CKKS scheme is most often used
in leveled mode.

Typical choices of $N$ are $N \geq 2^{15}$, enabling powerful
single-instruction-multiple-data (SIMD) batching of computation. This, with the
support for fixed-point computation on real numbers, makes RNS-CKKS the arguably
most popular choice for private ML
applications~\cite{DathathriSCLLMMM19,DathathriKSDLM20,IshiyamaSY20,
LeeKLCEDLLYKN21,LeeLLKKNC21,LouJ21}. Unfortunately, choosing a large $L$ incurs
a significant performance penalty. Therefore, representing the computation in a
way that minimizes multiplicative depth is a key challenge for practical
deployments of RNS-CKKS. \tool tackles this challenge when designing an efficient
approximation of \Argmax for FHE, which we will present in \cref{sec:argmaxhe}.

\subsection{Privacy-Preserving Neural Network Inference}
\label{ssec:background:fhenn}

We next describe how FHE is used to perform privacy-preserving inference on
neural network models.

\para{Setting} We focus on the classification task, where for given inputs $x
\in \mathbb{R}^d$ from $c$ classes, a classifier $f \colon \mathbb{R}^d \to
\mathbb{R}^c$ maps inputs to class scores (discussed shortly). We model $f$ as a
neural network with $k$ successively applied \emph{layers}. As a first step, in this work we focus on
fully-connected ReLU networks: each layer $f_i\colon \mathbb{R}^{n_{i-1}} \to
\mathbb{R}^{n_{i}}$ of the network either represents a linear transformation
$f_i(t) = W_i t + b_i$ of the input $t \in \mathbb{R}^{n_{i-1}}$, where $W_i \in
\mathbb{R}^{n_{i} \times n_{i-1}}$ and $b_i \in \mathbb{R}^{n_{i}}$, or a non-linear
ReLU activation function $f_i(t) = max(0, t)$, where $max$ is applied
componentwise. Layers $f_1$ and $f_k$ are generally linear, with $n_0=d$, and
$n_{k}=c$. The output of the full network \mbox{$z = f(x) = f_k \circ \ldots
\circ f_1(x) \in \mathbb{R}^c$} is a vector of unnormalized scores, called
\emph{logits}, for each class (see also \cref{fig:overview}). To simplify the
discussion, we assume that the elements of $z$ are unique. We say that for an
input $x$ the class with the highest logit is the class \emph{predicted by}
$f(x)$. For $[c] := \{1, \ldots, c\}$, we define $F\colon \mathbb{R}^d \to [c]$
to compute the class $F(x)$  predicted by $f(x)$.

\para{Inference \& Batching}
We now discuss how neural network inference (that is, the evaluation of $f(x)$)
is commonly encoded in RNS-CKKS~\cite{DathathriSCLLMMM19, DathathriKSDLM20,
LouJ21}. For an input \mbox{$x \in \mathbb{R}^d$}, we write $[x_1, \ldots, x_d,
0^{M-d}]$ to denote either an RNS-CKKS message underlying the plaintext
$\texttt{Ecd}(x) \in \psp$, or the corresponding ciphertext $\enc{x} \in \csp$,
where the distinction is clear from the context. Here, $0^{M-d}$ denotes
\mbox{$M-d$} zeroes padding the message to $M$ slots.

To efficiently evaluate a fully-connected layer $f_i({s}) = W_is + b_i$ on
ciphertext $s$ and plaintexts $W_i$ and $b_i$, we use the hybrid matrix-vector
product (MVP) algorithm of~\citet{JuvekarVC18}. This algorithm uses a sequence
of $\oplus_p$, $\otimes_p$, and $\texttt{RotL}$ operations to compute an MVP. It
requires $\texttt{RotL}$ to produce valid cyclic rotations of $s_{1}, \ldots,
s_{d}$ in the first $d$ (not $M$) slots of the ciphertext. To satisfy this, the
MVP algorithm is applied on a \emph{duplicated} version $s'$ of the input
message~$s$, namely $s'= [s_1, \ldots, s_d, s_1, \ldots, s_d, 0^{M-2d}]$,
obtained by \mbox{$s' \leftarrow s \oplus \texttt{RotR}([s_1, \ldots, s_d,
0^{M-d}], d)$}.

Recall that all operations to be performed in \fhe need to be reduced to the
primitives of the RNS-CKKS scheme. As a result, evaluating
non-polynomial functions such as ReLU is practically infeasible. Like
previous work~\citet{DathathriSCLLMMM19}, we replace ReLU with the \emph
{learnable square} activation function $c_2 x^2 + c_1 x$, where coefficients
$c_2$ and $c_1$ are optimized during training.

It is common to use $M \gg 2d$ and leverage SIMD batching to simultaneously
perform inference on a batch of $B = M / 2d$ (assuming $2d$ divides $M$)
duplicated inputs. Using rotations and $\oplus$ we put $B$ duplicated inputs in
one ciphertext, obtaining
\begin{equation} \label{eq:xbatch}
    \bar{x} = [x^{(1)}_1, \ldots, x^{(1)}_{2d}, ~\cdots~, x^{(B)}_1, \ldots, x^{(B)}_{2d}],
\end{equation}
which consists of $B$ \emph{blocks} of size $2d$ each. Propagating this through
the network, we obtain a batch of logits
\begin{equation*}
    \bar{z} = [z^{(1)}_1, \ldots, z^{(1)}_{c}, \trash^{2d-c}, ~\cdots~, z^{(B)}_1, \ldots, z^{(B)}_{c}, \trash^{2d-c}],
\end{equation*}
where we use $\trash$ to denote irrelevant slots which contain arbitrary values,
produced as a by-product of inference. 

In \cref{sec:smoothing}, we will describe how \tool relies on all introduced
techniques for private inference in reliable neural networks.

\subsection{Local Robustness}
\label{ssec:background:robustness}

The discovery of {adversarial examples}
\cite{SzegedyZSBEGF2013,BiggioCMNSLGR2013}, seemingly benign inputs that cause
mispredictions of neural networks, lead to the investigation of robustness of
neural networks. This is commonly formalized as \emph{local robustness}: a model
$F$ is locally-robust in the $\ell_p$ ball of radius $R$ around input $x$ if,
for some class $k$,
\begin{equation} \label{eq:consistency-guarantee}
    \forall \delta, \|\delta\|_p < R\colon \quad {F(x + \delta)} = k.
\end{equation}
 
In an attempt to mitigate adversarial examples, various heuristic methods
\cite{GoodfellowSS2015,KurakinGB2017} produce \emph{empirically robust}
networks. However, such models have been repeatedly successfully attacked
\cite{AthalyeCW18, CarliniPBRTGMK19, TramerCBM20} and are hence not suited to
safety-critical scenarios. In contrast, \emph{robustness certification} methods
\cite{GehrMDCTCV18, GowalDSBQJMK18, CohenRK19, SinghGPV19,
XuETAL20} provide rigorous mathematical guarantees that $F$ is locally-robust
around some input~$x$ according to \cref{eq:consistency-guarantee}, for some
choice of  $p$ and $R$.

In \tool, we consider $\ell_2$ as well as $\ell_1$ local robustness
certification (\ie ${p = 2}$ and ${p=1}$ in \cref{eq:consistency-guarantee}).
However, \tool can likely be extended to other settings, as we will discuss in
\cref{ssec:background:smoothing}.

\para{Attack Vectors \& Natural Changes}
As discussed above, local robustness offers protection against data
perturbations. In practice, there are two different sources of such
perturbations.
First, an adversary may deliberately and imperceptibly perturb client data
before or after it is acquired by the client, \eg compromising client's data
collection, storage, or loading procedure~\cite{BagdasaryanS21}. Second, even in
the absence of deliberate perturbations, ML models are susceptible to volatile
behavior under natural changes such as measurement noise or lighting
conditions---in fact, this is a key motivation for robustness
\cite{EngstromTTSM19,HendrycksD19,WongK21,LaidlawSF21, HendrycksZBSS21, TaoriDSCRS20}.
The active client-side attacker introduced in \cref{ssec:overview:threat-model}
models both these sources of perturbations.

\subsection{Individual Fairness}
\label{ssec:background:fairness}
The notion of \emph{fairness} of ML models has no universal definition in the
literature \cite{CatonH20}, however group fairness \cite{DworkHPRZ12,
HardtPNS16} and individual fairness \cite{DworkHPRZ12} are the most prevalent notions
in the research community. \tool focuses on individual fairness, whose
guarantees could be perceived as stronger, as models that obey group fairness
for certain groups can still discriminate against other population subgroups or
individual users \cite{KearnsNRW18}.

Individual fairness states a model should treat similar individuals
similarly, most often formalized as \emph{metric fairness}
\cite{DworkHPRZ12, RuossBFV20, YurochkinBS20, MukherjeeYBS20}:
\begin{equation} \label{eq:metric_fairness}
  \forall x_{1}, x_{2} \in \mathcal{X}\colon \quad d_{y}(F(x_{1}), F(x_{2})) \leq L d_{x}(x_{1}, x_{2}),
\end{equation}
where $L \in \mathbb{R}^{> 0}$, and $F \colon \mathcal{X} \to \mathcal{Y}$ maps
the input space $\mathcal{X}$ with metric $d_{x}$ to the output space
$\mathcal{Y}$ with metric $d_{y}$. Intuitively, for individuals $x_1$ and $x_2$
close \wrt $d_x$, the difference in the output of $F$ must not exceed a certain
upper bound, proportional to the input distance.

\subsection{Randomized Smoothing}
\label{ssec:background:smoothing}

We now recap randomized smoothing, which we will use in \cref{sec:smoothing} to
provide fairness and robustness guarantees.

\para{Smoothed Classifier}
\emph{Randomized smoothing (RS)}~\cite{CohenRK19} is a procedure that at
inference time transforms a base classifier $F \colon \mathbb{R}^d \to [c]$ (\eg
a trained neural network) into a \emph{smoothed classifier} $G \colon
\mathbb{R}^d \to [c]$ as follows. For given input $x$, the classifier $G$
returns the most probable prediction of $F$ under isotropic Gaussian
perturbations of $x$. Formally, for $\varepsilon \sim \mathcal{N}(0,
\sigma^2I_d)$, where $I_d$ is the $d$-dimensional identity matrix and $\sigma
\in \mathbb{R}^{\geq 0}$ is the standard deviation, we define
\begin{equation} \label{eq:smoothed}
    G(x) := \argmax_{j \in [c]} p_j
	\quad\text{with}\quad
	p_j := \mathbb{P}( F(x+\varepsilon) = j).
\end{equation}

The smoothed classifier $G$ behaves similarly to $F$, but with the additional
property that slight changes in the input only result in slight changes of the
output. In particular, $G$ is guaranteed to be locally-robust (see
\cref{eq:consistency-guarantee}) in the $\ell_2$ ball of radius $R$ around input
$x$, where $R$ can be determined as a function of $x$, $F$ and~$\sigma$. In
\cref{sec:smoothing}, we will show how RS can also be used to enforce individual
fairness.

\begin{algorithm}[t]
    \caption{Certification with Randomized Smoothing~\cite{FischerBV21}}
    \label{alg:certify}
    \small

\begin{algorithmic}[1]
    \State \textit{\# a subroutine to evaluate $f$ at $x$ under Gaussian noise}
    \State \textbf{function} \textsc{SampleUnderNoise}($f$, $x$, $n$, $\sigma$)
    \label{lin:certify-sample} \State \quad $\mathit{counts}$ $\leftarrow$ $[{0,
    \ldots, 0}]$ of size $c$ \State \quad \textbf{for} $i \leftarrow 1$
    \textbf{to} $n$ \textbf{do} \State \quad \quad $\varepsilon \leftarrow$ sample
    from $\mathcal{N}(0, \sigma^2I_d)$ \label{lin:certify-gauss} \State \quad \quad $x' \leftarrow x +
    \varepsilon$ \label{lin:certify-noise} \State \quad \quad $z \leftarrow f(x')$
    \label{lin:certify-infer} \State \quad \quad $y \leftarrow$
    \textsc{Argmax}($z$) \Comment{one-hot vector} \label{lin:certify-argmax-1-hot}
    \State \quad \quad $\mathit{counts} \leftarrow \mathit{counts} + y$
    \label{lin:certify-aggregate} \State \quad \textbf{return} $\mathit{counts}$ \label{lin:certify-sample-under-noise-return}
\vspace{0.08in}
    \State \textit{\# evaluate and certify the robustness of $g$ at $x$} \State
    \textbf{function} \textsc{Certify}($f$, $x$, $\sigma$, $n$, $n_{0}$, $\tau$,
    $\alpha$) \label{lin:certify-certify} \State \quad $\mathit{counts}_{0}$
    $\leftarrow$ \textsc{SampleUnderNoise}($f$, $x$, $n_{0}$, $\sigma$) \label{lin:certify-sample-under-noise-1} \State
    \quad $k \leftarrow$ \textsc{ArgmaxIdx}($\mathit{counts}_{0}$)
    \Comment{integer index} \label{lin:certify-argmax2} \State \quad
    $\mathit{counts}$ $\leftarrow$ \textsc{SampleUnderNoise}($f$, $x$, $n$,
    $\sigma$) \label{lin:certify-sample-under-noise-2} \State \quad $pv \leftarrow $\textsc{BinPValue}($counts[k]$, $n$, $\tau$) \label{lin:certify-binp} \State \quad \textbf{if} $pv \leq
    \alpha$ \textbf{return} $k$ \textbf{else} \textbf{abstain} \label{lin:certify-return-abstain}
\end{algorithmic}
\end{algorithm}

\para{Probabilistic Certification}
Unfortunately, exactly computing $p_j$ in \cref{eq:smoothed}, required to
determine both $G(x)$ and~$R$, is typically intractable (for most $F$). However,
using Monte Carlo sampling, $p_j$ can be lower-bounded by $\underline{p_j}$ with
confidence $1-\alpha$. From this, for fixed target radius $R \in
\mathbb{R}^{\geq 0}$, we can say that $G(x) = j$ if \mbox{$\underline{p_j} > \tau =
\Phi(R/\sigma)$} with the same confidence $1-\alpha$, where $\Phi$ denotes the
standard Gaussian CDF.
\textsc{SampleUnderNoise} in \cref{alg:certify} follows this approach to sample
$n$ Gaussian perturbations of input $x$ with standard deviation $\sigma$, and
count the number of predictions of $F$ for each class.

For a target $R$ and error upper bound $\alpha$, we can check whether this
sampling-based approximation satisfies \cref{eq:consistency-guarantee} with
probability at least $1-\alpha$ using a statistical test. In \cref{alg:certify},
we show the function \textsc{Certify} performing this step, inspired by
\citet{FischerBV21}. 

To ensure statistical soundness, the function first uses
$n_0$ samples of $F(x + \varepsilon)$ to guess the most likely class $k$
predicted by $G$. Note that this \emph{preliminary counting} procedure can use
any heuristic to guess $k$, as this does not affect the algorithm soundness. We
exploit this in \cref{ssec:smoothing:smoothing}, where we introduce the
\emph{soft preliminary counting} heuristic.

Given $k$, \textsc{Certify} further performs a one-sided binomial test
(\cref{lin:certify-binp}) on a fresh set of~$n$ samples, to show (up to error
probability $\alpha$) that the probability $p_k$ for $G$ to predict $k$ is at
least $\tau$, and thus the certified radius is at least $R$. If the test fails,
the function returns an ``abstain'' decision. Otherwise, the function returns
the final prediction $k$, guaranteed to satisfy \cref{eq:consistency-guarantee}
with probability at least $1 - \alpha$. See \cite[Theorem~1 and
Proposition~2]{CohenRK19} for a proof. 

Note that in \cref{fig:overview}, we do not make a distinction between
preliminary (\cref{lin:certify-sample-under-noise-1}) and main counting
(\cref{lin:certify-sample-under-noise-2}). While these are conceptually
independent procedures, \tool heavily relies on batching
(\cref{ssec:background:fhenn}) to efficiently execute them jointly (discussed in
\cref{sec:smoothing}).

RS relies on several algorithmic building blocks such as \Argmax and statistical
testing, which are challenging to perform in FHE. \tool overcomes this
(\crefrange{sec:argmaxhe}{sec:smoothing}) and is instantiated to obtain local
robustness and individual fairness guarantees via RS.

\para{Generalizations} RS has been extended beyond $\ell_2$ certification in the
setting of classification to a wide range of perturbation types
\cite{SalmanLRZZBY19, FischerBV20, YangDHSRL20} and settings
\cite{ChingCAKDG20, FischerBV21, KumarLF21}. For example, \citet{YangDHSRL20}
show that sampling uniform noise $\varepsilon \sim \mathcal{U}(-\eta, \eta)$ in
place of Gaussian noise (\cref{lin:certify-gauss} of \cref{alg:certify}) and
executing the rest of RS as usual leads to an $\ell_1$ robustness certificate
with radius $R = 2 \eta (\tau - 0.5)$. Further, any $R$ certified in the
$\ell_2$ case directly implies $\ell_\infty$ robustness with radius
$R/\sqrt{d}$, where $d$ is the input dimension. In our experimental evaluation
in \cref{sec:eval} we explore both $\ell_2$ and $\ell_1$ certification.

\section{Approximating the \Argmax operator in FHE} \label{sec:argmaxhe} 

Recall the key algorithmic building blocks of \tool visualized in
\cref{fig:overview}.  While some steps can easily be performed in FHE, a key
challenge is posed by the \Argmax step (B): as FHE does not allow evaluation of
control flow (\eg branching), and ciphertext comparison (\eg checking
inequality) is not directly supported by the homomorphic primitives of the
RNS-CKKS FHE scheme, we cannot implement \Argmax in the canonical way. Instead,
we aim for an approximation. Formally, our goal is to approximate the following
function on an RNS-CKKS ciphertext logit vector: 
\begin{equation} \label{eq:argmax}
    [z_1, \ldots, z_c, 0^{M-c}] \to [y_1, \ldots, y_c, \trash^{M-c}],
\end{equation}
where $y_i = 1$ for the index $i$ corresponding to the largest value among $z_i$
(and $0$ elsewhere). Recall that we assume $z_i$ are unique. 

In the following, we present the first efficient approximation of \Argmax for
RNS-CKKS (we discuss related work in \cref{sec:related}). Our key insight is
that \Argmax can be expressed via simple arithmetic operations and the sign
function, for which efficient RNS-CKKS approximations exist. For careful
parameter choices and appropriate conditions (introduced shortly), we achieve
both efficiency and the ability to bound the error of our approximation,
guaranteeing the soundness of reliability guarantees returned by \tool.

\para{Sign Function Approximation} To efficiently realize \textsc{Argmax}, we
rely on a polynomial approximation of the sign function. In particular, we
utilize recent work \cite{CheonKK20} that efficiently approximates
\begin{equation*}
    \text{\textsc{Sgn}}(x) \vcentcolon=
    \begin{cases} 
        \phantom{-}1 & \text{if~} x>0\\
        \phantom{-}0 & \text{if~} x=0\\
        -1 & \text{if~} x<0
     \end{cases}
\end{equation*}
in RNS-CKKS for $x \in [-1, 1]$. The proposed approximation \textsc{SgnHE}
composes odd polynomials $p_n$ and $q_n$ parametrized by integer $n$:
\begin{equation} \label{eq:sgnhe}
\text{\textsc{Sgn}}(x) \sim \text{\textsc{SgnHE($d_q, d_p, n, x$)}} = p_n^{d_p} \circ q_n^{d_q} 
\end{equation}
For desired precision parameters $\phi$ and $\psi$, we can choose integers
$d_q(\phi)$ and $d_p(\psi)$ such that the approximation is
\emph{$(\phi,\psi)$-close}, meaning that for $x \in [\phi, 1]$ (resp. $x \in
[-1, -\phi]$), the output of \textsc{SgnHE} is guaranteed to be in $[1 -
2^{-\psi}, 1]$ (resp. $[-1, -1 + 2^{-\psi}]$). Note that for inputs $x \in [-1,
-\phi] \cup [\phi, 1]$, $\textsc{SngHE}$ never outputs $0$.

As suggested in~\cite{CheonKK20}, we choose \mbox{$n = 4$} to efficiently
evaluate the polynomials using the Baby-Step Giant-Step algorithm~\cite{HanK20}.
This results in a multiplicative depth of \textsc{SgnHE} of $4(d_p + d_q)$.
We remark that future iterations of \tool could benefit from ongoing work on
optimizing non-polynomial function approximation in FHE, such as the latest
extension of \citet{LeeLKN22}. 

\begin{algorithm}[t]
    \caption{Approximation of \textsc{Argmax} for RNS-CKKS}
    \label{alg:argmax}
    \small
    \begin{algorithmic}[1]
        \State \textbf{function} \textsc{ArgmaxHE} 
        \State \quad \textbf{Inputs}: $z = [z_1, \ldots, z_c, 0^{M-c}]$, $d_{q}^{(1)}$, $d_{p}^{(1)}$, $d_{q}^{(2)}$, $d_{p}^{(2)}$
        \State \quad \textbf{Output}: $y = [y_1, \ldots, y_c, \trash^{M-c}]$ as in \cref{eq:argmax}
        \State \quad $z \leftarrow z \oplus \texttt{RotR}(z, c)$ \label{lin:argmax:logit-dup}
        \State \quad $\mathit{scores} \leftarrow [0^M]$ \label{lin:argmax:score-init}
        \State \quad $z_{rot} \leftarrow z$
        \State \quad \textbf{for} $\mathit{offset} \leftarrow 1$ \textbf{to} $c-1$ \textbf{do}
        \State \quad \quad $z_{rot} \leftarrow \texttt{RotL}(z_{rot}, 1)$
        \State \quad \quad $\mathit{diff} \leftarrow z \ominus z_{rot}$ \label{lin:argmax:diff}
        \State \quad \quad $\mathit{signs} \leftarrow
        \text{\textsc{SgnHE($d_{q}^{(1)}$, $d_{p}^{(1)}$, 4, $\mathit{diff}$)}}$  \label{lin:argmax:sign-1}
        \State \quad \quad $\mathit{scores} \leftarrow \mathit{scores} \oplus \mathit{signs}$
        \State \quad $\mathit{scores} \leftarrow (\mathit{scores} \otimes_p
        [\frac{1}{2c-2}^M]) \ominus_p [\frac{c-2}{2c-2}^M]$ \label{lin:argmax:scores}
        \State \quad $\mathit{scores} \leftarrow \mathit{scores} \otimes_p [1^{c}, 0^{M-c}]$ \label{lin:argmax:mask}
        \State \quad $y \leftarrow \text{\textsc{SgnHE($d_{q}^{(2)}$,
        $d_{p}^{(2)}$, 4, $\mathit{scores}$)}}$  \label{lin:argmax:sign-2}
        \State \quad $y \leftarrow (y \otimes_p [\frac{1}{2}^M]) \oplus_p [\frac{1}{2}^M]$ \label{lin:argmax:transform-y}
        \State \quad return $y$
    \end{algorithmic}
\end{algorithm} 

\para{Argmax Approximation} In \cref{alg:argmax} we present \textsc{ArgmaxHE},
which efficiently approximates \textsc{Argmax} via \textsc{SgnHE}. Our
construction is similar to the ideas of \cite{SebertPZGS21,IliashenkoZ21}
(discussed in \cref{sec:related}), but specifically targets the RNS-CKKS scheme.
The procedure takes as input a logit ciphertext $z$ and produces an output
ciphertext $y$, generalizing naturally to batched computation. \textsc{ArgmaxHE}
relies on \textsc{SgnHE} and is parameterized by $d_{q}^{(1)}$, $d_{p}^{(1)}$,
$d_{q}^{(2)}$, $d_{p}^{(2)}$ used for \textsc{SgnHE}.

\begin{figure}
    \centering
    \includegraphics[width=0.66\linewidth]{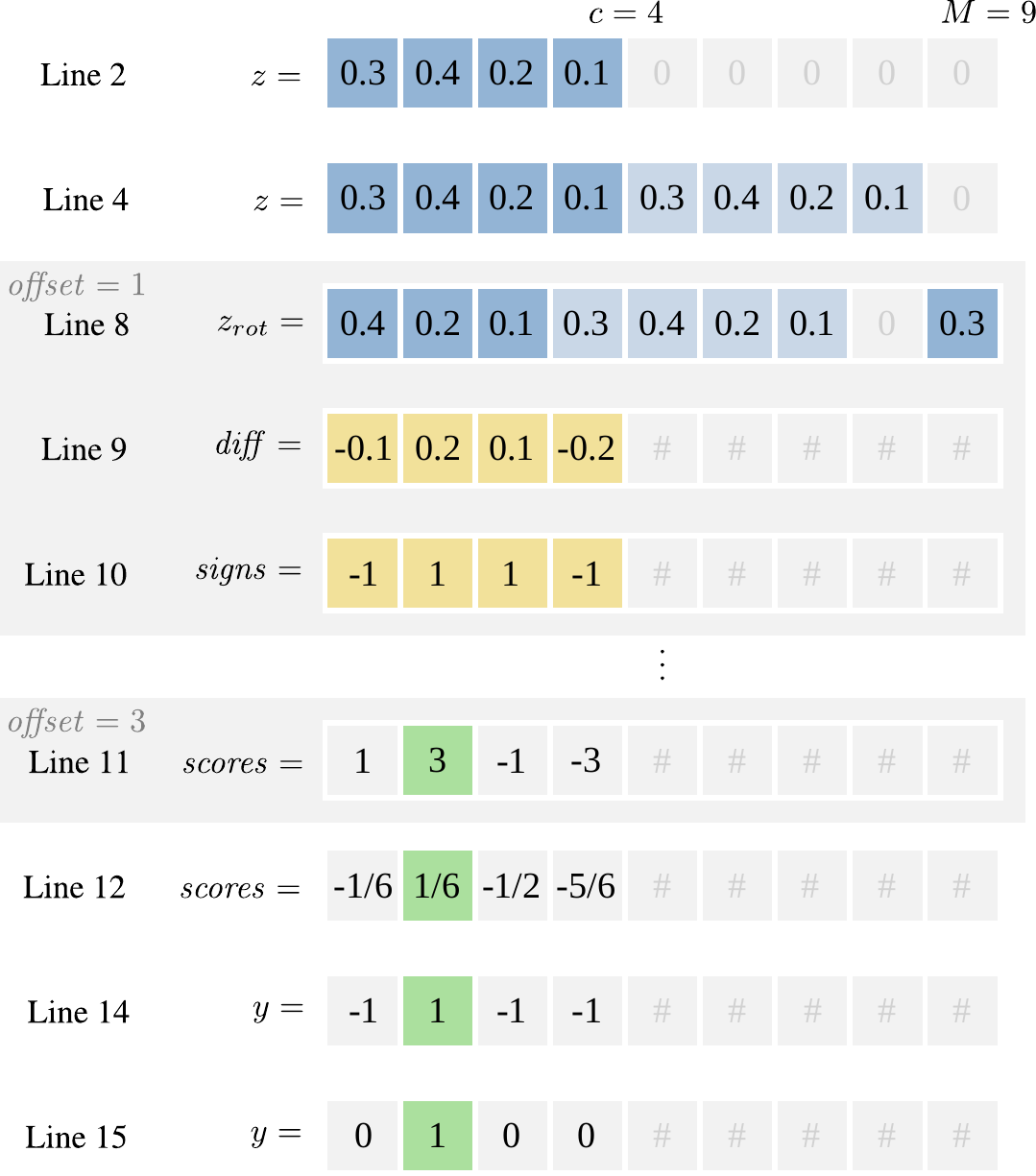}
    \caption{Example run of \cref{alg:argmax}.}
    \label{fig:argmax-example}
\end{figure}

In \cref{fig:argmax-example}, we illustrate how \cref{alg:argmax} processes a
toy example. The algorithm first duplicates the logits
(\cref{lin:argmax:logit-dup}) and initializes $\mathit{scores}$ to $0$. Then,
for each offset \mbox{$\mathit{offset} \in [1, c-1]$} we construct
$\mathit{diff}$ (\cref{lin:argmax:diff}), such that $\mathit{diff}_i = z_{i} -
z_{j}$, where \mbox{$j = (i+\mathit{offset}-1)\ \%\ (c+1)$} takes the values
$i+1, i+2, \ldots, i-1$ in successive iterations. For example, in
\cref{fig:argmax-example}, the first iteration with $\mathit{offset} = 1$
computes in $\mathit{diff}$ the slot-wise difference between $z$ and $z_{rot}$
to obtain $\mathit{diff}_0 = 0.3 - 0.4 = -0.1$. Applying \textsc{SgnHE}
(\cref{lin:argmax:sign-1}) to these differences maps them to $\{-1, 1\}$ (see
$\mathit{signs}$ in \cref{fig:argmax-example}). Summing these up yields
\mbox{$\mathit{scores}_i = \sum_{j \neq i} \left( 2 \cdot \indicator{z_i > z_j}
- 1\right)$} for each logit, where \mbox{$\indicator{\phi} = 1$} if $\phi$
holds, and $0$ otherwise.
We next argue that assuming $c$ is even and $z_i$ are unique, the first $c$
slots of $\mathit{scores}$ are exactly a permutation of \mbox{$S := [-(c-1),
-(c-3), \ldots, -1, 1, \ldots, c-3, c-1]$}. Assume w.l.o.g.\ that the logits are
sorted increasingly as $z_1 < z_2 < \cdots < z_c$. Then, for $i \in \{1, \ldots,
c\}$, it is $\mathit{scores}_i = \sum_{j < i} 1 + \sum_{j > i} (-1) = (2i - c -
1)$, giving the array $S$ above. As the logits are not necessarily sorted, the
resulting scores for $i\in \{1, \ldots, c\}$ are a permutation of $S$.
By construction, the index $k$ at which we find $c-1$ is the index of the
largest logit $z_k$. For example, in \cref{fig:argmax-example}, the value of
$\mathit{scores}$ computed in the last iteration ($\mathit{offset} = 3$)
contains the value~$3$ at index $k = 1$.

In order to identify~$k$, we transform the first $c$ slots of $\mathit{scores}$
to $[-1, \ldots, -1/(2c-2), 1/(2c-2)]$ (\cref{lin:argmax:scores}), multiply
$\mathit{scores}$ with $[1^c, 0^{M-c}]$ to clear the unused values which
stabilizes the ciphertext (\cref{lin:argmax:mask}), and use another
\textsc{SgnHE} (\cref{lin:argmax:sign-2}) to map all of the first $c$ slots but
slot $k$ to $-1$. The final transformation (\cref{lin:argmax:transform-y})
produces the desired one-hot vector.

\para{Input Requirement and Error Bound} As we previously noted, \textsc{SgnHE}
provably approximates \textsc{Sgn} well if, for small $\phi$, its input is in \mbox{$[-1, -\phi]
\cup [\phi, 1]$}. In particular, the two invocations of
\textsc{SgnHE} in \cref{lin:argmax:sign-1} and \cref{lin:argmax:sign-2} require
the inputs $\mathit{diff}$ and $\mathit{scores}$, respectively, to satisfy this
requirement in order to bound the error of \cref{alg:argmax}. 

In \cref{ssec:error:reliability}, we will introduce two conditions on the logit
vectors $z \in \mathbb{R}^c$ returned by the neural network $f$ in \tool, which
(assuming appropriate parameter choices, also discussed in
\cref{ssec:error:reliability}) ensure that the above input requirement provably
holds. By analyzing the probability of these conditions to be violated, we will
derive an upper bound on the overall error probability of \tool, \ie the
probability that a returned guarantee does not hold. We will further discuss the
values of this bound in practice, concluding that the introduced conditions are
not restrictive.

\para{Multiplicative Depth} By modifying the polynomials involved in the
evaluation of \textsc{SgnHE}, we can absorb the applications of the $\otimes_p$
operation in \cref{lin:argmax:scores} and \cref{lin:argmax:transform-y},
reducing the multiplicative depth. As a result, the \textsc{ArgmaxHE} function
in \cref{alg:argmax} has a multiplicative depth of
\mbox{$\lambda_{\mathit{argmax}} = 4 \cdot ( d_q^{(1)}+d_p^{(1)} +
d_q^{(2)}+d_p^{(2)} ) + 1$}, where $1$ arises from \cref{lin:argmax:mask}.

\section{Private Randomized Smoothing}
\label{sec:smoothing} Next, we present how to evaluate RS (see
\cref{ssec:background:smoothing}) in the RNS-CKKS FHE scheme, leveraging
\ArgmaxHE from \cref{sec:argmaxhe}. In particular, we discuss how to
homomorphically execute \cref{alg:certify} and obtain local robustness
guarantees (\toolR, \cref{ssec:smoothing:smoothing}), and how to extend this
approach to guarantee individual fairness (\toolF,
\cref{ssec:smoothing:fairness}).

\subsection{Robustness Guarantees} \label{ssec:smoothing:smoothing}

\toolR instantiates \tool for robustness certification by evaluating
\textsc{Certify} on the server, lifted to \fhe. In the following, we present the
main challenges in achieving this and how \tool overcomes them. First, we
discuss how we implement the \textsc{SampleUnderNoise} subroutine, including the
considerations needed to efficiently execute neural network inference. Then, we
introduce the idea of using \emph{soft preliminary counting} to improve
efficiency, before we describe how to implement the statistical test of
\textsc{Certify}.

\para{Sampling under Noise} The client provides a single encrypted input $x =
[x_0, \ldots, x_d, 0^{M-d}]$ to \toolR. As we require $n$ copies of $x$ in
\cref{lin:certify-noise}, we use a sequence of \texttt{RotR} and $\oplus$
operations to create $n$ duplicated copies $\{x^{(1)}, \ldots, x^{(n)}\}$ of
$x$, packed in $n/B$ ciphertexts as in \cref{eq:xbatch}, where $B = M/(2d)$ is
the batch size. 
Note that all RS parameters, including $n$, are set by the server, \mbox{as noted in \cref{ssec:overview:server}}. 

Next, we independently sample $n$ Gaussian noise vectors
according to \cref{lin:certify-gauss}, and use $\oplus_p$ to homomorphically add
the noise to batched duplicated inputs. We next apply batched neural network
inference, described shortly, to obtain batched logit vectors~$z^{(i)}$, further
reduced to a single ciphertext $\bar{z}$.

\para{Inference} Recall that in the pipeline of \cref{fig:overview}, the inputs
$x^{(i)}$ are encrypted using FHE scheme. To homomorphically evaluate the
neural network $f$ in the inference step (\cref{lin:certify-infer} of
\cref{alg:certify}), we rely on methods from related work discussed in
\cref{ssec:background:fhenn}. Namely, we directly adopt the described approaches
to evaluate $n_\mathit{lin}$ linear and $n_\mathit{act}$ ReLU layers with
multiplicative depth of $\lambda_{\mathit{inf}} = 2 (n_\mathit{lin} +
n_\mathit{act}) - 1$, and heavily utilize SIMD batching, adapting it to our
use-case to reduce the number of expensive FHE operations.

In particular, as commonly $2d \ll M$, we can batch the inputs $x^{(i)}$ in
$M/(2d)$ ciphertexts. Note that this is in contrast to most MLaaS setups, where
inference is typically only performed on a single input at a time and SIMD
batching therefore cannot be effectively utilized~\cite{DathathriSCLLMMM19,
LouJ19}. Next, as inference transforms each $2d$-dimensional input into a
$c$-dimensional logit vector, and commonly $c \ll 2d$, we can further batch the
logit vectors $z^{(i)}$, again using rotations, $\oplus$, and masking of
$\trash$ values using $\otimes_p$ (which adds $\lambda_{\mathit{batching}} = 1$
to the multiplicative depth). This reduces the number of ciphertext inputs to
the costly prediction step (discussed shortly), where it is typically possible
to place all logit vectors into a single ciphertext $\bar{z}$.

\para{Prediction and Aggregation} For the prediction step
(\cref{lin:certify-argmax-1-hot}), we apply \ArgmaxHE (\cref{sec:argmaxhe}) to
$\bar{z}$ to obtain $\bar{y}$, a ciphertext of batched one-hot prediction
vectors. To aggregate them (\cref{lin:certify-aggregate}), we use a simple
variation of the \emph{RotateAndSum} algorithm~\cite{JuvekarVC18}, which repeatedly
applies rotations and $\oplus$. \textsc{SampleUnderNoise} returns the
ciphertext
\begin{equation*}
    \mathit{counts} = [\mathit{cnt}_1, \ldots, \mathit{cnt}_c, \trash^{M-c}]
\end{equation*}
containing the number of times each class was predicted.

\para{Soft Preliminary Counting}
Determining the most likely class $k$ based on $\mathit{counts}_0$ requires two
invocations of \textsc{ArgmaxHE}
(\cref{lin:certify-argmax-1-hot,lin:certify-argmax2}), significantly increasing
the depth of \textsc{Certify}. 

Recall from \cref{ssec:background:smoothing} that using a heuristic during
preliminary counting does not affect the soundness of RS. Following this, we
introduce the \emph{soft preliminary counting} heuristic. Namely, when computing
$\mathit{counts}_0$, instead of invoking \textsc{ArgmaxHE} to compute one-hot
predictions $y$, we simply rescale $z$ by $1/n$ (by accordingly scaling the
weights of $f$) to obtain a rough approximation of $y$. Consequently, the result
returned in \cref{lin:certify-sample-under-noise-return} holds the average of
all logit vectors.

Avoiding another invocation of \ArgmaxHE greatly reduces the multiplicative
depth of \toolR, which would otherwise be prohibitive. At the same time, soft
counting does not affect the soundness of the guarantees, and as we demonstrate
in \cref{sec:eval}, it rarely impacts the decisions of \textsc{Certify}. See
\appref{app:softcount} for further discussion.

\para{Statistical Testing}
We next describe how to evaluate the statistical test in
\crefrange{lin:certify-binp}{lin:certify-return-abstain}, which provides a robustness guarantee.

Let $p_k$ denote the probability that the smoothed model~$G$
(\cref{eq:smoothed}) predicts $k$ as computed in \cref{lin:certify-argmax2}. The
one-sided binomial test attempts to prove that $p_k > \tau$ with error
probability at most $\alpha$. To this end, it computes the $p$-value \mbox{$pv =
1 - \textsc{BinCDF}(\mathit{counts}[k]-1; n, \tau)$}, the probability of
predicting the class $k$ at least $\mathit{counts}[k]$ out of $n$ times under
the worst case $p_k=\tau$ of the null hypothesis $p_k \leq \tau$, where
$\textsc{BinCDF}$ denotes the binomial CDF. 

\begin{figure}
    \centering
    \includegraphics[width=0.65\linewidth]{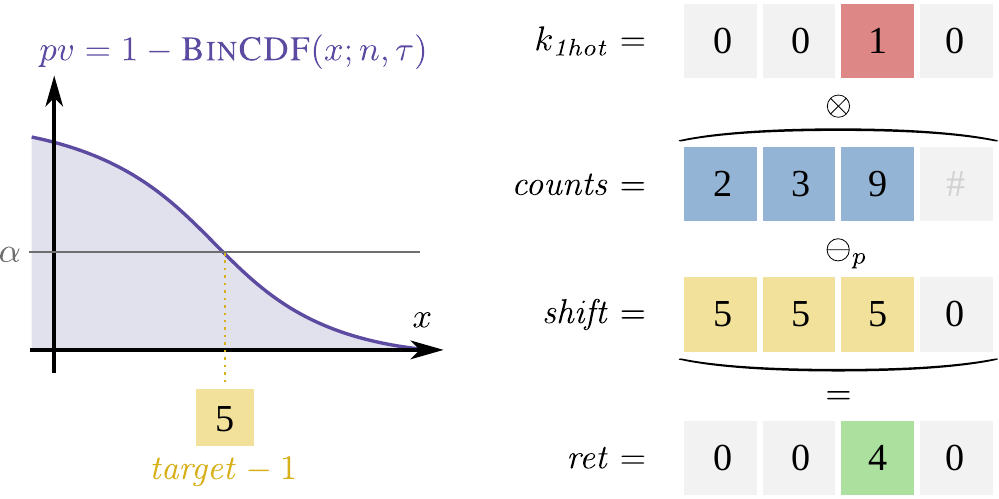}
    \caption{The statistical test of \tool, illustrated on an example.} 
    \label{fig:bintest}
\end{figure}

Computing $\textsc{BinCDF}$ on $\mathit{counts}[k]$ involves approximating the
incomplete beta function, which is not easily reducible to FHE primitives. To
overcome this, we rephrase the test using the following observation: For fixed
$n$ and $\tau$, the \mbox{p-value} $pv$ is strictly decreasing \wrt
$\mathit{counts}[k]$. Thus, we employ binary search to find $\mathit{target}$,
the smallest value of $\mathit{counts}[k]$ for which $pv \leq \alpha$, \ie the
test passes. We visualize this in \cref{fig:bintest}, where $k = 3$, and
$\mathit{target} = 6$.

Let $k_\mathit{1hot}$ denote the one-hot vector encoding the predicted class,
obtained by multiplying the result of \textsc{ArgmaxHE} in
\cref{lin:certify-argmax2} with $[1^c, 0^{M-c}]$ using $\otimes_p$, such that
the non-zero slot is at index $k$. The server returns a ciphertext
$\mathit{ret}$ to the client, computed as follows:
\begin{equation} \label{eq:ret}
    \mathit{ret} \leftarrow k_{\mathit{1hot}} \otimes (\mathit{counts} \ominus_p \mathit{shift}),
\end{equation}
where \mbox{$\mathit{shift} = [(\mathit{target}-1)^c, 0^{M-c}]$} is the
plaintext with $\mathit{target}-1$ in first $c$ slots, and $0$ in rest. Note
that the use of $\otimes_p$ to obtain $k_{1hot}$ and the use of $\otimes$ in
\cref{eq:ret} add $\lambda_{ret} = 2$ to multiplicative depth. We illustrate the
computation of $\mathit{ret}$ in \cref{fig:bintest}, where $M = 4$ and $c = 3$. 

\para{Interpreting the Result} The client decrypts and decodes $\mathit{ret}$
using the secret key, rounds all its entries to the closest integer value, and
finds $Z$, the unique non-zero value in first $c$ slots of~$\mathit{ret}$. If
such a $Z$ does not exist (\ie $\mathit{counts}[k] = target-1$), we define $Z =
0$. In \cref{fig:bintest} it is $Z=4$ at slot $k=3$.
There are two possible outcomes.

\begin{itemize}
    \item \emph{Abstention}: If $Z\leq 0$, the prediction is to be discarded, as
    robustness cannot be guaranteed.
    \item \emph{Guarantee}: If $Z>0$, the model predicts $k$ and is guaranteed
    to be robust in the $\ell_2$ ball of radius $R$ around the input $x$
    (\cref{eq:consistency-guarantee}) with probability at least $1-\xi$
    (discussed below), where $R$ and $\xi$ can be communicated to the client
    unencrypted.
\end{itemize}
Note that if necessary, the server can avoid revealing the predicted class in
the \emph{abstention} case, and the value of $Z$, by appropriately rescaling $Z$
within $\mathit{ret}$ and applying another \textsc{SgnHE}, such that
$\mathit{ret}$ is one-hot if the outcome is guaranteed to be robust, and the
all-zero vector otherwise. In our implementation, we omit this step to avoid
increasing the multiplicative depth.

The non-homomorphic version of RS (\cref{alg:certify}) ensures robustness with
probability at least $1-\alpha$ if it does not return ``abstain''. As the
homomorphic version in \toolR includes various approximations, it introduces an
additional approximation error, resulting in a total error probability bound
$\xi$ (discussed in~\cref{sec:error}).

\para{Multiplicative Depth} Summing up previously derived depths of inference,
batching, \textsc{ArgmaxHE}, and the calculations in \cref{eq:ret}, we obtain
the total multiplicative depth of \toolR: 
\begin{align*}
    \lambda &= \lambda_{\mathit{inf}} + \lambda_{\mathit{batching}} + \lambda_{\mathit{argmax}} + \lambda_{\mathit{ret}} \\ 
    &= 2(n_\mathit{lin} + n_\mathit{act}) + 4(d_q^{(1)}+d_p^{(1)} + d_q^{(2)}+d_p^{(2)}) + 3.
\end{align*}
An extensive experimental evaluation given in \cref{ssec:eval:smoothing}
demonstrates that this leads to practical latency.

\subsection{Individual Fairness Guarantees}
\label{ssec:smoothing:fairness}
Next, we present how \toolF instantiates \tool for individual fairness
guarantees by applying a generalization of RS.

\para{Metric Fairness as Robustness}
Inspired by \citet{YurochkinBS20,MukherjeeYBS20} and \citet{RuossBFV20}, we
transform the definition of metric fairness (\cref{eq:metric_fairness}) into a
variant of local robustness (\cref{eq:consistency-guarantee}), allowing us to
utilize RS to obtain guarantees. Similar to \cite{YurochkinBS20,MukherjeeYBS20},
we consider input space metrics of the form
\begin{equation*} 
  d_{x}(x_{1}, x_{2}) := \left(x_{1} - x_{2} \right)^{\top} \Theta \left(x_{1} - x_{2}\right),
\end{equation*}
where $\Theta$ is a symmetric positive definite matrix. We focus on classifiers,
thus we have $\mathcal{Y} = [c]$ in \cref{eq:metric_fairness}, and $d_{y}(y_{1}, y_{2})
:= \indicator{y_{1} \neq y_{2}}$, \ie $d_{y}(y_{1}, y_{2}) = 1$ if $y_{1} \neq
y_{2}$, and $0$ otherwise.

For fixed $x$, to show that $F$ is individually fair, we need to show:
\begin{equation}
  \label{eq:lipp_fair_simple}
  \forall x' \in \mathcal{X}\colon \quad \indicator{F(x) \neq F(x')} \leq L \left(x - x'\right)^{\top} \Theta \left(x - x'\right).
\end{equation}
We let $\|x\|_{\Theta} := \sqrt{x^{\top} \Theta x}$ denote the Mahalanobis norm and
rewrite \cref{eq:lipp_fair_simple} as
\begin{align}
  & \forall x' \in \mathcal{X}\colon \quad F(x) \neq F(x') \implies \|x-x'\|_{\Theta}^{2} \geq 1/L \nonumber \\
  \iff & \forall x' \in \mathcal{X}\colon \quad \|x-x'\|_{\Theta}^{2} < 1/L \implies F(x) = F(x') \nonumber \\
  \iff &\forall \delta \; \text{with} \; \|\delta\|_{\Theta} < \sqrt{1/L}\colon \quad F(x) = F(x + \delta),
  \label{eq:robustness2}
\end{align}
which represents a local robustness constraint (\cref{eq:consistency-guarantee})
with radius $R=\sqrt{1/L}$, and the $\ell_2$ norm generalized to the
$\|\cdot\|_{S}$ norm.

In other words, by achieving Mahalanobis norm robustness, we can achieve
equivalent individual fairness guarantees.

\para{Similarity Constraint} To set $\Theta$ and $L$, we choose the
\emph{similarity constraint} $\theta \in \mathbb{R}^d$, encoding the
problem-specific understanding of similarity. Concretely, $\theta_i$ is chosen
to quantify the minimum difference in attribute $i$ (\eg client age in our
example of fair loan eligibility prediction from \cref{sec:motivation}) for
which two individuals are considered dissimilar, assuming all other attributes
are fixed.

The similarity constraint should be set by a data regulator \cite{DworkHPRZ12,
McNamaraOW17}. In our analysis, following prior work on fairness certification
\cite{RuossBFV20,john2020verifying}, we consider the problem of choosing the
similarity constraint orthogonal to this investigation and assume it to be
given.

Given $\theta$, we set $\Theta = diag(1/\theta^2)$, where squaring is applied
componentwise, and $L=1$, which implies $R = \sqrt{1/L}=1$. Intuitively, for two
individuals differing only in attribute $i$ by $\delta < s_i$, we will have
$\|\delta\|_{\Theta} < 1 = R$, \ie to ensure fairness, the prediction of the
model $F$ should be the same for these individuals, as per
\cref{eq:lipp_fair_simple}.

\para{Mahalanobis Randomized Smoothing}
To produce individual fairness guarantees via Mahalanobis norm robustness,
\toolF leverages a generalization of RS from \citet[Theorem A.1]{FischerBV20}.
In particular, the noise added before the inference step is sampled from
$\mathcal{N}(0, \Sigma)$ instead of $\mathcal{N}(0, \sigma^2 I_d)$, where
$\Sigma$ is a covariance matrix. If the algorithm does not return the
``abstain'' decision, the returned class $k$ satisfies the following robustness
guarantee:
\[
  \forall \delta \text{ with } \|\delta\|_{\Sigma^{-1}} < R: \quad G(x + \delta) = k,
\]
where $G$ denotes the smoothed classifier to be used instead of $F$ (see
\cref{ssec:background:smoothing}), and $R= \Phi^{-1}(\tau)$, for $\tau$ and
$\Sigma$ being parameters of RS.
 
Given $\Theta$ and $L$, instantiating this with $\Sigma = \Theta^{-1}$ and $\tau
= \Phi(\sqrt{1/L})$ implies $R=\sqrt{1/L}$ and recovers \cref{eq:robustness2},
directly leading to individual fairness guarantees with respect to our
similarity constraint. 

Therefore, by changing the distribution of the noise applied to the inputs in
\toolR as part of RS, we obtain \toolF. An experimental evaluation of \toolR and
\toolF is given in \cref{sec:eval}.

\section{Security and Reliability Properties} \label{sec:error}

We now provide a more formal discussion of the properties guaranteed by \tool.
After presenting its security notion (\cref{ssec:error:privacy}), we analyze its
robustness and fairness guarantees (\cref{ssec:error:reliability}).

\subsection{Client Data Privacy} \label{ssec:error:privacy}

As stated in \cref{thm:privacy} below, any passive attacker sitting on the
server or on the network between client and server (see also
\cref{ssec:overview:threat-model}) cannot learn anything about the client data.

\begin{theorem}[Client Data Privacy]\label{thm:privacy}
    A passive polynomial-time attacker with access to all of the server's state
    cannot distinguish two different plaintext inputs $x \neq x'$ of the client.
\end{theorem}

The theorem directly follows from the confidentiality property of
RNS-CKKS~\cite{CheonHKKS18Rns}, assuming hardness of the Ring LWE problem (note
that in \tool, decrypted values are never shared with the server, hence the
attack of \citet{Li2021} does not apply).

\subsection{Robustness and Fairness} \label{ssec:error:reliability}

Recall that in non-private RS (\cref{alg:certify}), a non-abstention result is
robust (\cref{eq:consistency-guarantee}) with probability at least $1-\alpha$,
for some \emph{algorithmic error} probability $\alpha$. In principle, \tool
inherits this guarantee. However, when lifting \cref{alg:certify} to FHE, \tool
introduces an additional \emph{approximation error}, as \textsc{SgnHE} (used
within \textsc{ArgmaxHE}) is only an approximation of \textsc{Sgn} (see
\cref{sec:argmaxhe}).
We analyze this approximation error and bound the total error to
prove \cref{thm:reliability} below.

\begin{theorem}[Reliability] \label{thm:reliability} With high confidence, a
    non-ab\-sten\-tion result returned by \toolR or \toolF satisfies local
    robustness (\cref{eq:consistency-guarantee}) or individual fairness (in the
    form of \cref{eq:robustness2}), respectively, with probability at least
    $1-\xi$, where $\xi$ is computed as described below.
\end{theorem}

Note that \cref{thm:reliability} ignores the additional error introduced by
noise in the RNS-CKKS scheme. However, in line with previous work, \tool
leverages existing error reduction techniques~\cite{BossuatMTH20}  to make these
errors negligible in practice, as we elaborate on in~\cref{ssec:eval:warmup}.
In our evaluation (\cref{sec:eval}), we observe low values of $\xi \leq 0.012$.

In order to prove \cref{thm:reliability}, we next introduce two conditions on
the output logits of the involved neural network. Then, we discuss how we can
upper bound the approximation error probability of \tool by computing the
probability of violating these conditions.

\para{Range and Difference Conditions}
Let $Z_\textit{min}, Z_\textit{max}, D$ be some constants. For all logit vectors
$z \in \mathbb{R}^c$ output by the neural network~$f$ used in \tool, we
introduce the following conditions:
\begin{itemize} 
    \item $\forall i\colon\ z_i \in [Z_\textit{min}, Z_\textit{max}]\ $ (\emph{range
    condition}), and
    \item $\forall i \neq j\colon\ |z_i-z_j| \geq D\ $ (\emph{difference
    condition}).
\end{itemize}

\para{Parameter Selection}
\tool jointly selects \mbox{$(Z_\textit{min}, Z_\textit{max}, D)$}, the
parameters of the \textsc{ArgmaxHE} function \mbox{$(d_{q}^{(1)}$,
$d_{p}^{(1)}$, $d_{q}^{(2)}$, $d_{p}^{(2)})$}, and the precision settings
$(\phi, \psi)$ for each \textsc{SgnHE} invocation such that \emph{zero}
approximation error is introduced if the range and difference conditions hold
(see also~\cref{sec:argmaxhe}). Importantly, choosing $(Z_\textit{min},
Z_\textit{max}, D)$ together with $(\phi, \psi)$ parameters of the first
\textsc{SgnHE} (\cref{lin:argmax:sign-1}) enables us to guarantee that the input
requirement of \textsc{SgnHE} holds, \ie all components of $\mathit{diff}$ are
in $[-1, -\phi] \cup [\phi, 1]$.

While wider $[Z_\textit{min}, Z_\textit{max}]$ and lower $D$ lead to fewer
violations of conditions, this increases the multiplicative depth of the
computation in FHE. In practice, \tool selects a reasonable tradeoff allowing
for both tight error bounds and practical latency. See \appref{app:params} for
full details of the parameter selection procedure.

\para{Bounding the Violation Probability}
To prove \cref{thm:reliability}, it hence suffices to compute an upper bound on
the probability of condition violations. Note that as we empirically observe in
our evaluation~(\cref{sec:eval}), the following derivation is rather loose and
the approximation error gracefully absorbs violations up to some
degree. 

The true range violation probability $\beta_R^\star$ (and analogously $\beta_D
^\star$) can only be computed if the true input distribution is known. However,
as common in machine learning, the true distribution is not known and \tool only
has access to samples from this distribution. To estimate upper bounds $\beta_R$
and $\beta_D$ of $\beta_R^\star$ and $\beta_D^\star$ respectively, \tool relies
on samples from a test set.

First, $\beta_R^\star$ is estimated by $\hat{\beta}_R$, the ratio of examples
$x$ from a test set for which \emph{at least one} of the output logits $z$
violates the range condition. Then, we use the
Clopper-Pearson~\cite{ClopperPearson} confidence interval with p-value $p =
0.05$ to obtain an upper bound $\beta_R$ of $\beta_R^\star$, which holds with
confidence $1 - p$.

We similarly compute an upper bound $\beta_D$ of $\beta_D^\star$: we first
calculate $\hat{\beta}_D$ as the ratio of test set examples for which at least
$\zeta = 1 \%$ of the logits $z$ violate the difference condition, and then
derive $\beta_D$ from $\hat{\beta}_D$ as before, again with confidence $1 - p$.
To ensure soundness when less than a fraction $\zeta$ of the logits~$z$ violate
the condition, we replace $\tau$ with $\tau + \zeta$ in \cref{alg:certify},
following~\cite[Lemma~2]{FischerBV20}.

\begin{proof}[Proof of \cref{thm:reliability}]
Given $\alpha$, $\beta_R$ and $\beta_D$, the total error probability~$\xi$ of
\tool can be computed using the union bound. In particular, with confidence
${1-2p = 0.9}$ (accounting for the two Clopper-Pearson intervals), any
non-abstention value returned by \toolR or \toolF satisfies robustness or
fairness, respectively, except with probability at most ${\xi = \alpha + \beta_R
+ \beta_D}$.
\end{proof}

By using more samples during the estimation of $\beta_R$ and $\beta_D$, we can
decrease $p$ and hence increase the confidence in the theorem.

\para{Unsound Parameters}
In \appref{app:reducing}, we study the effect of selecting unsound parameters
\mbox{$d_{q}^{(1)}$, $d_{p}^{(1)}$, $d_{q}^{(2)}$, $d_{p}^{(2)}$} of
\textsc{ArgmaxHE}. That is, in order to reduce the multiplicative depth of the
FHE computation, we decrease the parameters to purposefully violate the core
\mbox{$(\phi, \psi)$-closeness} property of \textsc{SgnHE} (thus invalidating
\cref{thm:reliability}).

We discover that reducing these parameters below theoretically sound values can
in practice still lead to results consistent with the ones obtained in a
non-private setting, which can be desirable in certain use-cases where formal
guarantees are not essential.

\section{Experimental Evaluation} \label{sec:eval}

Here, we present a thorough experimental evaluation of \tool, posing two main research questions.
\begin{enumerate}[leftmargin=32pt]
    \item [(RQ1)] \emph{Consistency}: Do the results produced by \tool match the
    results of the same procedure evaluated in a standard, non-private setting?
    \item [(RQ2)] \emph{Efficiency}: Are the latency and communication costs of
    \tool sufficiently small to make transitioning real deployments to a private
    FHE-based setting feasible?
\end{enumerate}
In order to answer RQ1, we first discuss reasons for inconsistencies and
substantiate why these are rare in practice (\cref{ssec:eval:warmup}), and then
confirm this empirically (\cref{ssec:eval:smoothing}). Next, in
\cref{ssec:eval:latency}, we discuss the efficiency of our system and answer
RQ2.

\para{Experimental Setup} To instantiate FHE in our experiments, we use the
Microsoft SEAL 3.6.6 library~\cite{SEALCrypto} with a security level of 128 bits
and the following cryptographic parameters: polynomial ring degree $N=2^{17}$,
initial scale $\Delta={2^{50}}$ and coefficient modulo \mbox{$Q = \prod_{l=0}^L
Q_l$} \st $Q_0 \approx 2^{60}$ and $Q_l \approx 2^{50}$ for all $l \geq 1$,
where $L$ is the maximum multiplicative depth of our program (see
\cref{ssec:background:fhe}). All experiments were run on a 2-socket machine with
Intel Xeon Gold 6242 2.80GHz CPU with 64 cores, with parallelization employed to
reduce the latency of batched inference and the main loop in \cref{alg:argmax}.

For each run of \tool performed in FHE, we execute the equivalent procedure in a
non-encrypted setting (\emph{non-private evaluation}) with the same set of noise
samples. This allows us to inspect potential inconsistencies caused by the
encryption and the approximations introduced by \tool (RQ1). Note that
non-private evaluation does not use the soft preliminary counting heuristic in
RS.

We remark that FHE primitives are several orders of magnitude slower than their
unencrypted equivalents \cite{ViandJH21}. While efficient implementations can
lead to practical latency, extensive testing of \tool on a large number of
inputs is impractical. To circumvent this issue and expand the scope of our
experiments, we introduce a mock implementation of the SEAL library
(\emph{MockSEAL}), which replicates all FHE operations in cleartext (\ie without
encryption). Note that in contrast to non-private evaluation, MockSEAL
faithfully simulates all approximations introduced by \tool, such as \ArgmaxHE
or soft preliminary counting. Thus, MockSEAL is a reliable proxy for SEAL (up to
FHE noise, discussed shortly).

\subsection{Causes and Prevalence of Inconsistencies} \label{ssec:eval:warmup}
Inconsistent results of \tool (\ie ones that do not match the non-private
evaluation) are almost nonexistent in practice, as we later confirm
experimentally (\cref{ssec:eval:smoothing}). We can identify three causes of
such inconsistencies, discussed below.

\para{(i) Harmful Violations} Recall the range and difference conditions
introduced in \cref{sec:error}. If satisfied, these ensure that the input
requirements of each invocation of \textsc{SgnHE} hold and the latter is hence a
good approximation of \textsc{Sgn} (see $(\phi, \psi)$-\emph{close}
in~\cref{sec:argmaxhe}). When proving \cref{thm:reliability}, we conservatively
assumed that any violation of these conditions leads to an error. Refining the
analysis of \cref{alg:argmax} in~\cref{sec:error} and~\appref{app:params}, we
can conclude that as long as $\max(z) - \min(z) \leq Z_\textit{max} -
Z_\textit{min}$ and $z_1 - z_2 \geq D$, where $z_1$ and $z_2$ are the two
largest logits, a violation will not cause an inconsistency---we say that such a
violation is \emph{harmless}. Otherwise, it is \emph{harmful} and may cause an
inconsistency. See~\appref{app:harm} for more details.

We note that harmful range violations are more severe in practice than harmful
difference violations: range violations are likely to compromise the whole
inference as they lead to inputs to \textsc{SgnHE} outside $[-1, 1]$, leading to
outputs arbitrarily diverging from the output of \textsc{Sgn}. However,
difference violations can at most affect one sample in the RS procedure, which
in turn often does not change the final result (the predicted class or the
abstention), as we substantiate and demonstrate in our empirical results
(\cref{ssec:eval:smoothing}).

\para{(ii) FHE Noise} RNS-CKKS performs \emph{approximate} arithmetic, meaning that
errors are introduced during computation due to various sources of noise
(see~\cite{KimPP20} for details). \tool leverages \emph{scale propagation}
by~\citet{BossuatMTH20} to eliminate the dominant noise component. As shown
in~\cite{DathathriKSDLM20}, for appropriate parameter choices, the impact of
other components is rarely relevant in practice.

\para{(iii) Soft Preliminary Counting} As previously discussed, the soft
preliminary counting heuristic used by \tool in RS can never produce an invalid
guarantee, \ie it does not impact the error probability as derived in
\cref{sec:error}. However, it can still lead to inconsistencies with the
non-private evaluation in cases where the heuristic guess for the most likely
class $k$ was incorrect. We further elaborate on this distinction in
\appref{app:softcount}.

\subsection{Evaluation of Consistency}
\label{ssec:eval:smoothing}

In our consistency experiments, we use the following models.%

\emph{MNIST}: A 2-layer fully-connected network with hidden layer size 32,
trained on the MNIST \cite{LeCunCB10} dataset of handwritten digit images of
size \mbox{$d = 28 \times 28$} to be classified in one of $c=10$ classes. As
usual when aiming to apply RS, we add Gaussian noise $\varepsilon \sim
\mathcal{N}(0, \sigma^2 I)$ to examples during training, setting $\sigma=0.5$.

\emph{Adult}: A network with the same architecture as MNIST above but trained on
the continuous features of the Adult dataset~\citep{Kohavi96}, in a binary
classification task ($c=2$) whose goal is to predict whether a person has a
yearly income of more or less than $\$50000$.

\emph{CIFAR}: A network trained on the more challenging
CIFAR-10~\cite{Krizhevsky09} dataset of color images ($d = 32
\times 32 \times 3$) featuring $c=10$ classes of objects. We use the
architecture proposed in prior work (network A in \citet{JuvekarVC18}, accuracy not reported): a
3-layer fully-connected network with hidden layers of size $128$. We use
$\sigma=0.25$.

We explore the consistency of \toolR ($\ell_2$ robustness) on MNIST and CIFAR
models, and \toolF (individual fairness) on the Adult model. To broaden the
scope of our investigation, we explore two additional settings by training
variants of the MNIST model: (i) the use of larger $\sigma \in \{0.75, 1.0,
1.5\}$ in RS, while setting $\tau$ such that the certified radius $R$ remains
unchanged; and (ii) $\ell_1$ certification via RS by adding uniform noise
$\varepsilon \sim \mathcal{U}(-\eta, \eta)$ (as described in
\cref{ssec:background:smoothing}), where we set $\eta=2.0$.

\tool can readily be used with other networks, including ones produced by
optimizing compilers such as CHET~\cite{DathathriSCLLMMM19}, which greatly
improves the inference latency. Unfortunately, the source code of CHET is not
publicly available at the time of this writing.

\begin{table}[t]\centering

    \caption{Overview of the models used in consistency experiments.}
    \label{table:nets}
    
    \newcommand{\twocol}[1]{\multicolumn{2}{c}{#1}}
    \renewcommand{\arraystretch}{1.2}

    \resizebox{\columnwidth}{!} {
    \begin{tabular}{@{}rrrrrrrr@{}} \toprule
    & Acc ($\%$) & $R$ & Cert ($\%$)&  $Z_\textit{min}$ & $Z_\textit{max}$ & D &$\xi$\\ \midrule
    MNIST  & 96.8 & 0.34 & 86.1 & -80 & 60 & 0.0056& 0.0028 \\
    CIFAR   & 52.7 & 0.17 & 47.7& -10 & 45 & 0.0022 & 0.0120  \\
    Adult  & 81.6 & n/a & 95.7& -180 & 180 & 0.0072 & 0.0014 \\
    \midrule 
    MNIST{$_{\sigma=0.75}$}   & 95.5  & 0.34 & 78.8& -40 & 50 & 0.0036 & 0.0016\\
    MNIST{$_{\sigma=1.0}$}   & 94.3 & 0.34 & 68.1& -30 & 45 & 0.0030& 0.0016   \\
    MNIST{$_{\sigma=1.5}$}   & 91.3 & 0.34 & 44.0& -25 & 30 & 0.0022 & 0.0018\\
    MNIST{$_{\eta=2.0}$}  & 93.25 & 0.5 & 62.1 & -35 & 50 & 0.0034& 0.0032  \\
    \bottomrule %
    \end{tabular}}
  \end{table} 
  
\begin{table*}[t]\centering

    \caption{Evaluation of private inference with guarantees with \tool on
    models from \cref{table:nets}. We show the validity of results obtained in
    FHE, and the consistency (in \%) with the non-private evaluation at
    prediction, count, and result level (see \cref{ssec:eval:smoothing}).
    Further, we show the fraction of (harmful; H) range violations (RV) and
    difference violations (DV). P- denotes the violations during the process of
    preliminary counting.}
    \label{table:results}
    
    \newcommand{\twocol}[1]{\multicolumn{2}{c}{#1}}
    \renewcommand{\arraystretch}{1.2}

    \resizebox{0.77\linewidth}{!} {
    \begin{tabular}{@{}rrrrrrrrr@{}} \toprule
    Model & Valid1Hot & PredOK & CountsOK & ResultOK & RV (H) & DV (H) & P-RV (H) & P-DV (H) \\ \midrule
    MNIST--MockSEAL & 100.00& 99.52 & 100.00 & 100.00 & 0.00 (0.00) & 92.08 (5.05) & 0.00 (0.00) & 2.23 (0.02) \\ 
    MNIST--SEAL & 100.00 & 100.00 & 100.00 & 100.00 & 0.00 (0.00) & 97.06 (7.84) & 0.00 (0.00) & 5.88 (0.00) \\
    CIFAR--MockSEAL& 99.99 & 94.71 & 99.97 & 99.98 & 0.22 (0.00) & 77.49 (11.68) & 0.05 (0.00) & 2.75 (0.22) \\
    CIFAR--SEAL& 100.00& 97.00 & 100.00& 100.00& 0.00 (0.00) & 82.00 (11.00) & 0.00 (0.00) & 3.00 (0.00) \\
    Adult--MockSEAL & 99.98 & 99.89 & 98.42 & 99.97 & 0.00 (0.00) & 29.41 (29.41) & 0.00 (0.00) & 0.12 (0.12) \\
    Adult--SEAL & 100.00 & 100.00 & 95.00 & 100.00 & 0.00 (0.00) & 58.00 (58.00) & 0.00 (0.00) & 0.00 (0.00) \\
    \midrule 
    MNIST{$_{\sigma=0.75}$}--MockSEAL & 100.00& 98.82 & 99.96 & 100.00& 0.00 (0.00) & 89.32 (6.57) & 0.00 (0.00) & 2.09 (0.00) \\
    MNIST{$_{\sigma=1.0}$}--MockSEAL & 100.00& 97.93 & 99.98 & 100.00& 0.00 (0.00) & 91.58 (9.28) & 0.00 (0.00) & 2.53 (0.08) \\
    MNIST{$_{\sigma=1.5}$}--MockSEAL & 99.99 & 95.42 & 99.96 & 99.99 & 0.00 (0.00) & 94.35 (12.88) & 0.00 (0.00) & 3.09 (0.07) \\
    MNIST{$_{\eta=2.0}$}--MockSEAL  & 99.99 & 97.35 & 99.98 & 99.99 & 0.00 (0.00) & 96.4 (13.56) & 0.00 (0.00) & 3.49 (0.03) \\
    \bottomrule 
    \end{tabular}}
  \end{table*}

\para{Guarantees of Obtained Models}
All trained models are summarized in \cref{table:nets}. We report the standard
accuracy, as well as certified accuracy at given radius $R$ (\ie the ratio of
examples where the outcome was not ``abstain''). Note that $R$ is not applicable
to the case of Adult, as we certify individual fairness according to a specific
similarity constraint $\theta \in \mathbb{R}^d$ (deferred to \appref{app:eval}).

Further, we report the values of $Z_\textit{min}, Z_\textit{max}$ and $D$,
chosen according to the procedure in \appref{app:params}, as well as the error
probability upper bound $\xi$, which is the sum of algorithmic error of RS and
the approximation error of \tool (see \cref{ssec:error:reliability}). More
details on parameter values are given in \appref{app:eval}.

We remark that incorporating RS, compared to standard FHE inference: (i)
increases the multiplicative depth, and (ii) requires training with gaussian
noise which is known to reduce accuracy~\cite{CohenRK19}. Thus, the depth of
models used in \tool is currently limited, leading to moderate accuracy on more
challenging datasets; we believe future advances in FHE will enable \tool to
also be instantiated for deeper networks.
 
\para{Methodology and Metrics} The consistency results are given in
\cref{table:results}. We use MockSEAL with the full test set for all models.
Additionally, we evaluate the main MNIST, CIFAR and Adult models using SEAL on
a subset of 100 test set examples. These examples are manually selected to
include an approximately equal number of ``abstain'' decisions, ``guaranteed
class $k$'' decisions where $k$ is correct, and same decisions where $k$ is
wrong. Note that picking this subset randomly would include mostly easy
examples, where the network is confident and small violations do not affect the
result. 

We report four consistency metrics. First, as harmful violations can lead to
unbounded outputs of \textsc{SgnHE}, the masking vector $k_{1hot}$ (see
\cref{eq:ret}) may have more than one non-zero component. Column
\emph{Valid1Hot} in \cref{table:results} quantifies how often ciphertext
$\mathit{ret}$ returned by server respects the one-hot property.

Columns \emph{PredOK}, \emph{CountsOK} and \mbox{\emph{ResultOK}} indicate
how often the FHE procedure is consistent with the non-private evaluation at
various points in the procedure. \emph{PredOK} is true if the class
predicted by soft preliminary counting ($k_{1hot}$ in \cref{eq:ret}) equals the
class predicted in the non-private case. \emph{CountsOK} is true if main counts
($\mathit{counts}$ in \cref{eq:ret}) are consistent. Finally, \emph{ResultOK}
reports the consistency of the decision (``abstain'' or ``guaranteed class
$k$'').

The remaining four columns show the statistics of range and difference
violations, stating the percentage of examples where at least one range
violation (\emph{RV}) or difference violation (\emph{DV}) occurred, where
\emph{P-} indicates that the violation happens on the averaged logit vector
during soft preliminary counting. Harmful violations are shown in parentheses.
Note that for the special case $c=2$ (Adult, \toolF), all violations are
harmful.

\para{Results} We observe near-perfect results of our implementation across all
settings (including large noise and $\ell_1$ certification), positively
answering RQ1, with nearly 100\% of cases following the protocol and being
consistent with the non-private evaluation (columns \emph{Valid1Hot} and
\emph{ResultOK}). We observe that range violations are nonexistent (except in
one case, where still none are harmful), and that difference violations are
common but rarely harmful for \toolR. For \toolF, harmful difference violations
are more common. Still, as noted in \cref{ssec:eval:warmup} this affects at most
one prediction, rarely compromising the final result. While increasing $\sigma$
leads to slightly more difference violations, range violations are still
nonexistent and there is no significant degradation of consistency.

As noted in \cref{ssec:eval:warmup}, procedures based on aggregation of
predictions with noise such as RS are naturally resilient to inaccuracies in
intermediate steps. This makes these procedures particularly suited to FHE with
RNS-CKKS, which often requires introducing various approximations (such as
\textsc{ArgmaxHE} used in this work). Namely, even predicting a different most
likely class~$k$ (\ie \emph{PredOK} is false) or producing inconsistent main
counts $\mathit{counts}$ (\ie \emph{CountsOK} is false) often does not impact
the consistency of the final result (\emph{ResultOK}). Intuitively, the
prediction $k$ of the preliminary counting heuristic is more likely to be
inconsistent in the non-private case when the model is unsure about its
predictions. However, such cases usually lead to the ``abstain'' decision, which
does not reveal $k$. Similarly, unconfident main counts leading to abstention
still abstain with small variations, and same holds for extremely confident
counts, where the predicted class is unlikely to change with small inaccuracies.

\begin{table*}[t]\centering

  \caption{The evaluation of performance. Each row represents a parameter
  variation of the baseline experimental setting (MNIST model). For each of the
  main components of \toolR, we report the mean latency of 10 runs, as well as
  the relative cost in isolation (see \cref{ssec:eval:latency}).}
  \label{table:latency}
  
  \newcommand{\twocol}[1]{\multicolumn{2}{c}{#1}}
  \renewcommand{\arraystretch}{1.2}

  \resizebox{0.65 \linewidth}{!}{
  \begin{tabular}{@{}rrrrrrr@{}} \toprule

  & (i)~Duplication & (ii)~Inference & \multirow{2}{*}{(iii)~\textsc{ArgmaxHE}}
  & (iv)~Aggregation && \multirow{2}{*}{Total} \\ 
  
  & + Noise & + Reduction &  & + BinTest &&  \\ \midrule

  Baseline & 37s (1.0) & 662s (191.0) & 157s (\phantom{0}84.5) & <1s (1.4) && 856s (277.9) \\ \midrule
  $1/2$ Data Size & 41s (1.0) & 608s (\phantom{0}98.6) & 157s (\phantom{0}72.8) & <1s (1.2) && 806s (173.6) \\
  $2 \times$ Data Size & 33s (1.0) & 740s (408.7) & 156s (100.2) & <1s (1.6) && 929s (890.1) \\
  $-1$ Layers & 31s (1.0) & 233s (\phantom{0}62.0) & 159s (\phantom{0}84.5) & <1s (1.4) && 424s (148.9) \\
  $+1$ Layers & 43s (1.0) & 1734s (466.5) & 158s (\phantom{0}84.5) & <1s (1.4) && 1935s (553.4) \\
  $1/2$ Width & 37s (1.0) & 431s (119.6) & 159s (\phantom{0}84.5) & <1s (1.4) && 628s (206.5) \\
  $2 \times$ Width & 37s (1.0) & 1177s (337.5) & 158s (\phantom{0}84.5) & <1s (1.4) && 1371s (424.4) \\

  \bottomrule 
  \end{tabular}}

\end{table*}

\subsection{Evaluation of Efficiency} \label{ssec:eval:latency} 

To answer RQ2, we analyze the communication cost and latency (wall clock time)
of processing a single client input in \tool.

\para{Communication Cost} As \tool uses a non-interactive FHE-based protocol,
the analysis of the communication cost is straightforward. Namely, each
client-server interaction requires the client to upload exactly one RNS-CKKS
ciphertext $x$ of fixed size (dependent on FHE parameters) and receives exactly
one ciphertext $ret$ as the result, as well as at most two unencrypted values:
the error bound $\xi$ and the robustness specification ($R \in \mathbb{R}$ for
\toolR, and $\theta \in \mathbb{R}^d$ for \toolF). Note that in FHE, where all
data for one client-server interaction is encrypted under the same key, we
cannot (unlike in non-private ML inference pipelines) apply batching to combine
multiple queries of different clients with different keys.
\para{Latency} The latency of models investigated in
\cref{ssec:eval:smoothing} is as follows: $15$~min for \toolR MNIST models,
$7$~min for the \toolF Adult model, and around $110$~min for the more
challenging \toolR CIFAR model. This does not include the per-client one-time
cost (up to $15$~min in all cases) needed to initialize the RNS-CKKS context.

These latencies are in the same order of magnitude as those reported in
literature for neural network evaluation in \fhe. As discussed above, FHE
primitives are known to be significantly slower than equivalent operations on
the CPU~\cite{ViandJH21}. Furthermore, \tool can directly benefit from work on
speeding up privacy-preserving machine learning with FHE, which we discuss in
\cref{sec:related}.

To study the efficiency of the individual steps of \tool, we perform a more
detailed analysis, focusing on \toolR with the MNIST model from
\cref{table:results}. Namely, we split the execution into four
\emph{components}: (i)~addition of noise to duplicated inputs, (ii)~neural
network inference with reduction of logit vectors (see "Inference" in
\cref{ssec:smoothing:smoothing}) , (iii)~the application of \textsc{ArgmaxHE},
and (iv)~counting (including preliminary counting) followed by statistical
testing. For each component, the ``Baseline'' row in \cref{table:latency}
reports its latency during a complete run (with parameters
from~\cref{ssec:eval:smoothing}).

Additionally, to understand the effect of different parameters on runtime, we
measure latency for variants of the baseline. In particular, we inspect the
effect of doubling (halving) the input data size, increasing (decreasing) the
number of layers of the network by 1, and doubling (halving) the width of the
first hidden layer.

\para{Results} The results of our latency evaluation are given in
\cref{table:latency}. In each cell, we report the mean latency of 10 independent
runs (the values in parentheses will be explained shortly).

We observe that most of the runtime is spent on neural network inference (ii),
being roughly $4$x more expensive than \textsc{ArgmaxHE}. 
Regarding the effect of parameter variation, we can observe that the number of
layers has the most significant impact on runtime. In particular, increasing the
layers by 1 approximately leads to doubled latency. However, recall that \tool
can readily adapt to advances in speeding up FHE such as purpose-build hardware,
an ongoing research area we discuss in \cref{sec:related}.

\para{Relative Costs} 
The latency of an atomic RNS-CKKS FHE operation (\emph{op}) depends on the ring
degree $N$ and the depth of the downstream computation after the op. Thus, while
the latencies in \cref{table:latency} reflect the costs of the components in the
actual instantiation of \tool, they do not accurately capture the relative costs
of components in isolation (\ie at equal depth). We hence more accurately
compute those relative costs as follows. First, we derive a cost for each atomic
op by measuring their relative latency for fixed parameters $N=2^{17}, L=1$. As
an example, this results in a cost of $1.0$ for $\oplus_p$, the most efficient
op, and $4.73$ for $\otimes$. Next, we count the type and the number of ops used
in each component of \tool and compute a weighted sum, resulting in a total
component cost which we further normalize \wrt the cheapest component. 

The obtained relative costs are reported in parentheses in
\cref{table:latency}. While inference still dominates, our measurements
suggest that it is (in the baseline run) only approximately twice as expensive
as \textsc{ArgmaxHE} (as opposed to $4$x suggested by the latency results).

\para{Summary}
We conclude that the performance of \tool is satisfactory, positively answering
RQ2. Moreover, our \textsc{ArgmaxHE} approximation is efficient, allowing for
broader use outside of the context of reliable inference---\eg as noted in
\cref{ssec:overview:server}, using \textsc{ArgmaxHE} in standard FHE inference
reduces the risk of model stealing.

\section{Related Work} \label{sec:related}

\para{Private Inference via FHE or MPC}
Many recent works consider the problem of neural network inference while
preserving the privacy of client data. To this end, like \tool, many works
instantiate FHE
schemes~\cite{GiladDLLNW16,ChouBLYHF18,BadawiJLMJTNAC21,BrutzkusGE19,LouJ19,LeeKLCEDLLYKN21,LeeLLKKNC21,LouJ21,deepreduce,BouraGGJ20,LuHHMQ20},
while others utilize secure multiparty computation~(MPC) or combine it with
FHE~\cite{PaymanZ17,LiuJLA17,RouhaniRK18,JuvekarVC18,MishraLSZP20,SafeNET,CryptoNAS}.
Both of these research directions are active and growing, with a universally
acknowledged non-comparable set of advantages and drawbacks. As noted in prior
work~\cite{BadawiJLMJTNAC21,BrutzkusGE19,ChouBLYHF18,GiladDLLNW16,deepreduce,LeeLLKKNC21,LeeKLCEDLLYKN21,LouJ19,LouJ21},
MPC-based works have several shortcomings: these approaches typically (i) have
high communication overhead; (ii) require continuous client presence during
inference; (iii) reveal the architecture and model details to the client; (iv)
come with more offline preprocessing (on both client and server); and (v) do not
support large-batch inference. Here, (iii) is often undesirable in a MLaaS
setting. Further, (i), (ii) and (v) are significantly amplified in the setting
of RS, where a large number of inference passes are made for a single client
query.

\para{Reliable Inference} 
A long line of research considers building models with reliability guarantees in
a non-private setting. The two types of guarantees we focus on in \tool are
robustness~\cite{GehrMDCTCV18,GowalDSBQJMK18,WongK18,CohenRK19,FischerBV21}, and
fairness~\cite{DworkHPRZ12,mcnamara2019costs,yeom2020individual,john2020verifying,YurochkinBS20,RuossBFV20}.

\para{Speeding up FHE}
Orthogonal to our work, recent research on optimizing compilers
\cite{BoemerLCW19,DathathriSCLLMMM19,DathathriKSDLM20,ChowdharyDLS21}, GPU
acceleration \cite{DaiS15}, and specialized hardware accelerators for
FHE~\cite{RiaziLPD20,FeldmannSKDDPS22} has demonstrated significant speedups for
FHE. These results suggest that sub-second latencies for \tool are within reach,
which would allow us to investigate more challenging settings in future work.
Specialized hardware will also greatly increase the demand for
privacy-preserving ML solutions, making tools that enable real-world scenarios
in private settings (such as \tool) more valuable.

\para{{Argmax} in FHE} Independent of reliability guarantees for 
ML models, several works discuss lifting \textsc{Argmax} to \fhe.
While \tool targets a non-interactive setting,~\citet{BostPTG15} and
\citet{ChooDDZJPW21} provide interactive protocols for private \textsc{Argmax}.
Works \cite{SebertPZGS21,IliashenkoZ21} discuss the non-interactive evaluation
of \textsc{Argmax} in generally incompatible TFHE~\cite{ChillottiGGI16} and
BGV/BFV~\cite{BrakerskiGV12,FanV12} schemes, thus the proposed constructions
cannot be directly applied to RNS-CKKS. Namely, BGV/BFV operate on integers,
while TFHE operates on bits, has no native support for batched inference, and
offers the concept of ``sign bootstrapping'' leveraged by \citet{SebertPZGS21}
to implement \textsc{Argmax}.
\citet{CheonKKLL19} propose a construction to evaluate \textsc{Argmax} in CKKS,
but this construction is prohibitively expensive~\cite{CheonKK20}. Finally,
\citet{CheonKK20} discuss implementing the \textsc{Max} function via the sign
function approximation, but focus on the simple case of two inputs assumed to be
in $[0,1]$, and thus do not support the significantly more complex general case
of $c>2$ inputs from an unknown domain. In addition, while the FHE engineering
community shows great interest in efficiently evaluating \textsc{Argmax} in
CKKS~\cite{GithubSeal1, GithubSeal2, GithubTenseal1}, no implementation is
publicly available to date.

\section{Conclusion}

We presented \tool, the first system for private inference on reliable neural
networks, instantiated with two important guarantees: local robustness and
individual fairness. The key idea is constructing efficient \fhe-counterparts
for core parts of randomized smoothing. Our evaluation demonstrated
acceptable latency costs of \tool. We believe our work is an important
step in enabling the use of reliable ML models in privacy-preserving settings.

\message{^^JLASTBODYPAGE \thepage^^J}

\clearpage
\balance
\bibliographystyle{ACM-Reference-Format}

\balance

\bibliography{references}


\begin{thebibliography}{124}


\ifx \showCODEN    \undefined \def \showCODEN     #1{\unskip}     \fi
\ifx \showDOI      \undefined \def \showDOI       #1{#1}\fi
\ifx \showISBNx    \undefined \def \showISBNx     #1{\unskip}     \fi
\ifx \showISBNxiii \undefined \def \showISBNxiii  #1{\unskip}     \fi
\ifx \showISSN     \undefined \def \showISSN      #1{\unskip}     \fi
\ifx \showLCCN     \undefined \def \showLCCN      #1{\unskip}     \fi
\ifx \shownote     \undefined \def \shownote      #1{#1}          \fi
\ifx \showarticletitle \undefined \def \showarticletitle #1{#1}   \fi
\ifx \showURL      \undefined \def \showURL       {\relax}        \fi
\providecommand\bibfield[2]{#2}
\providecommand\bibinfo[2]{#2}
\providecommand\natexlab[1]{#1}
\providecommand\showeprint[2][]{arXiv:#2}

\bibitem[Apostolidis and Papakostas(2021)]%
        {Apostolidis21}
\bibfield{author}{\bibinfo{person}{Kyriakos~D. Apostolidis} {and}
  \bibinfo{person}{George~A. Papakostas}.} \bibinfo{year}{2021}\natexlab{}.
\newblock \showarticletitle{A Survey on Adversarial Deep Learning Robustness in
  Medical Image Analysis}. In \bibinfo{booktitle}{\emph{Electronics}}.
\newblock


\bibitem[Athalye et~al\mbox{.}(2018)]%
        {AthalyeCW18}
\bibfield{author}{\bibinfo{person}{Anish Athalye}, \bibinfo{person}{Nicholas
  Carlini}, {and} \bibinfo{person}{David~A. Wagner}.}
  \bibinfo{year}{2018}\natexlab{}.
\newblock \showarticletitle{Obfuscated Gradients Give a False Sense of
  Security: Circumventing Defenses to Adversarial Examples}. In
  \bibinfo{booktitle}{\emph{ICML}}.
\newblock


\bibitem[Badawi et~al\mbox{.}(2021)]%
        {BadawiJLMJTNAC21}
\bibfield{author}{\bibinfo{person}{Ahmad~Al Badawi}, \bibinfo{person}{Chao
  Jin}, \bibinfo{person}{Jie Lin}, \bibinfo{person}{Chan~Fook Mun},
  \bibinfo{person}{Sim~Jun Jie}, \bibinfo{person}{Benjamin Hong~Meng Tan},
  \bibinfo{person}{Xiao Nan}, \bibinfo{person}{Khin Mi~Mi Aung}, {and}
  \bibinfo{person}{Vijay~Ramaseshan Chandrasekhar}.}
  \bibinfo{year}{2021}\natexlab{}.
\newblock \showarticletitle{Towards the AlexNet Moment for Homomorphic
  Encryption: HCNN, the First Homomorphic {CNN} on Encrypted Data With GPUs}.
  In \bibinfo{booktitle}{\emph{TETCI}}.
\newblock


\bibitem[Bagdasaryan and Shmatikov(2021)]%
        {BagdasaryanS21}
\bibfield{author}{\bibinfo{person}{Eugene Bagdasaryan} {and}
  \bibinfo{person}{Vitaly Shmatikov}.} \bibinfo{year}{2021}\natexlab{}.
\newblock \showarticletitle{Blind Backdoors in Deep Learning Models}. In
  \bibinfo{booktitle}{\emph{USENIX}}.
\newblock


\bibitem[Balakrishnan et~al\mbox{.}(2020)]%
        {Balakrishnan20}
\bibfield{author}{\bibinfo{person}{Tara Balakrishnan}, \bibinfo{person}{Michael
  Chui}, \bibinfo{person}{Bryce Hall}, {and} \bibinfo{person}{Nicolaus Henke}.}
  \bibinfo{year}{2020}\natexlab{}.
\newblock \showarticletitle{The state of AI}.
\newblock
\newblock
\shownote{\url{https://www.mckinsey.com/business-functions/mckinsey-analytics/our-insights/global-survey-the-state-of-ai-in-2020}}.


\bibitem[Barua(2021)]%
        {Barua21}
\bibfield{author}{\bibinfo{person}{Hrishav~Bakul Barua}.}
  \bibinfo{year}{2021}\natexlab{}.
\newblock \showarticletitle{Data science and Machine learning in the Clouds: A
  Perspective for the Future}. In \bibinfo{booktitle}{\emph{arXiv}}.
\newblock


\bibitem[Biggio et~al\mbox{.}(2013)]%
        {BiggioCMNSLGR2013}
\bibfield{author}{\bibinfo{person}{Battista Biggio}, \bibinfo{person}{Igino
  Corona}, \bibinfo{person}{Davide Maiorca}, \bibinfo{person}{Blaine Nelson},
  \bibinfo{person}{Nedim {\v{S}}rndi{\'c}}, \bibinfo{person}{Pavel Laskov},
  \bibinfo{person}{Giorgio Giacinto}, {and} \bibinfo{person}{Fabio Roli}.}
  \bibinfo{year}{2013}\natexlab{}.
\newblock \showarticletitle{Evasion attacks against machine learning at test
  time}. In \bibinfo{booktitle}{\emph{ECML PKDD}}.
\newblock


\bibitem[Boemer et~al\mbox{.}(2019)]%
        {BoemerLCW19}
\bibfield{author}{\bibinfo{person}{Fabian Boemer}, \bibinfo{person}{Yixing
  Lao}, \bibinfo{person}{Rosario Cammarota}, {and} \bibinfo{person}{Casimir
  Wierzynski}.} \bibinfo{year}{2019}\natexlab{}.
\newblock \showarticletitle{NGraph-HE: A Graph Compiler for Deep Learning on
  Homomorphically Encrypted Data}. In \bibinfo{booktitle}{\emph{CF}}.
\newblock


\bibitem[Bojchevski et~al\mbox{.}(2020)]%
        {BojchevskiKG20}
\bibfield{author}{\bibinfo{person}{Aleksandar Bojchevski},
  \bibinfo{person}{Johannes Klicpera}, {and} \bibinfo{person}{Stephan
  G{\"{u}}nnemann}.} \bibinfo{year}{2020}\natexlab{}.
\newblock \showarticletitle{Efficient Robustness Certificates for Discrete
  Data: Sparsity-Aware Randomized Smoothing for Graphs, Images and More}. In
  \bibinfo{booktitle}{\emph{ICML}}.
\newblock


\bibitem[Bossuat et~al\mbox{.}(2021)]%
        {BossuatMTH20}
\bibfield{author}{\bibinfo{person}{Jean-Philippe Bossuat},
  \bibinfo{person}{Christian Mouchet}, \bibinfo{person}{Juan
  Troncoso-Pastoriza}, {and} \bibinfo{person}{Jean-Pierre Hubaux}.}
  \bibinfo{year}{2021}\natexlab{}.
\newblock \showarticletitle{{Efficient Bootstrapping for Approximate
  Homomorphic Encryption with Non-sparse Keys}}. In
  \bibinfo{booktitle}{\emph{EUROCRYPT}}.
\newblock


\bibitem[Bost et~al\mbox{.}(2015)]%
        {BostPTG15}
\bibfield{author}{\bibinfo{person}{Raphael Bost}, \bibinfo{person}{Raluca~Ada
  Popa}, \bibinfo{person}{Stephen Tu}, {and} \bibinfo{person}{Shafi
  Goldwasser}.} \bibinfo{year}{2015}\natexlab{}.
\newblock \showarticletitle{Machine Learning Classification over Encrypted
  Data}. In \bibinfo{booktitle}{\emph{NDSS}}.
\newblock


\bibitem[Boura et~al\mbox{.}(2020)]%
        {BouraGGJ20}
\bibfield{author}{\bibinfo{person}{Christina Boura}, \bibinfo{person}{Nicolas
  Gama}, \bibinfo{person}{Mariya Georgieva}, {and} \bibinfo{person}{Dimitar
  Jetchev}.} \bibinfo{year}{2020}\natexlab{}.
\newblock \showarticletitle{{CHIMERA:} Combining Ring-LWE-based Fully
  Homomorphic Encryption Schemes}. In \bibinfo{booktitle}{\emph{JMC}}.
\newblock


\bibitem[Brakerski et~al\mbox{.}(2012)]%
        {BrakerskiGV12}
\bibfield{author}{\bibinfo{person}{Zvika Brakerski}, \bibinfo{person}{Craig
  Gentry}, {and} \bibinfo{person}{Vinod Vaikuntanathan}.}
  \bibinfo{year}{2012}\natexlab{}.
\newblock \showarticletitle{(Leveled) Fully Homomorphic Encryption without
  Bootstrapping}. In \bibinfo{booktitle}{\emph{ITCS}}.
\newblock


\bibitem[Brutzkus et~al\mbox{.}(2019)]%
        {BrutzkusGE19}
\bibfield{author}{\bibinfo{person}{Alon Brutzkus}, \bibinfo{person}{Ran
  Gilad-Bachrach}, {and} \bibinfo{person}{Oren Elisha}.}
  \bibinfo{year}{2019}\natexlab{}.
\newblock \showarticletitle{Low Latency Privacy Preserving Inference}. In
  \bibinfo{booktitle}{\emph{ICML}}.
\newblock


\bibitem[Buolamwini and Gebru(2018)]%
        {BuolamwiniG18}
\bibfield{author}{\bibinfo{person}{Joy Buolamwini} {and}
  \bibinfo{person}{Timnit Gebru}.} \bibinfo{year}{2018}\natexlab{}.
\newblock \showarticletitle{Gender shades: Intersectional accuracy disparities
  in commercial gender classification}. In \bibinfo{booktitle}{\emph{FAccT}}.
\newblock


\bibitem[Carlini et~al\mbox{.}(2019)]%
        {CarliniPBRTGMK19}
\bibfield{author}{\bibinfo{person}{Nicholas Carlini}, \bibinfo{person}{Anish
  Athalye}, \bibinfo{person}{Nicolas Papernot}, \bibinfo{person}{Wieland
  Brendel}, \bibinfo{person}{Jonas Rauber}, \bibinfo{person}{Dimitris Tsipras},
  \bibinfo{person}{Ian~J. Goodfellow}, \bibinfo{person}{Aleksander Madry},
  {and} \bibinfo{person}{Alexey Kurakin}.} \bibinfo{year}{2019}\natexlab{}.
\newblock \showarticletitle{On Evaluating Adversarial Robustness}. In
  \bibinfo{booktitle}{\emph{arXiv}}.
\newblock


\bibitem[Caton and Haas(2020)]%
        {CatonH20}
\bibfield{author}{\bibinfo{person}{Simon Caton} {and}
  \bibinfo{person}{Christian Haas}.} \bibinfo{year}{2020}\natexlab{}.
\newblock \showarticletitle{Fairness in Machine Learning: {A} Survey}. In
  \bibinfo{booktitle}{\emph{arXiv}}.
\newblock


\bibitem[Cheon et~al\mbox{.}(2018a)]%
        {CheonHKKS18Bootstrap}
\bibfield{author}{\bibinfo{person}{Jung~Hee Cheon}, \bibinfo{person}{Kyoohyung
  Han}, \bibinfo{person}{Andrey Kim}, \bibinfo{person}{Miran Kim}, {and}
  \bibinfo{person}{Yongsoo Song}.} \bibinfo{year}{2018}\natexlab{a}.
\newblock \showarticletitle{Bootstrapping for Approximate Homomorphic
  Encryption}. In \bibinfo{booktitle}{\emph{EUROCRYPT}}.
\newblock


\bibitem[Cheon et~al\mbox{.}(2018b)]%
        {CheonHKKS18Rns}
\bibfield{author}{\bibinfo{person}{Jung~Hee Cheon}, \bibinfo{person}{Kyoohyung
  Han}, \bibinfo{person}{Andrey Kim}, \bibinfo{person}{Miran Kim}, {and}
  \bibinfo{person}{Yongsoo Song}.} \bibinfo{year}{2018}\natexlab{b}.
\newblock \showarticletitle{A full {RNS} variant of approximate homomorphic
  encryption}. In \bibinfo{booktitle}{\emph{SAC}}.
\newblock


\bibitem[Cheon et~al\mbox{.}(2017)]%
        {CheonKKS17}
\bibfield{author}{\bibinfo{person}{Jung~Hee Cheon}, \bibinfo{person}{Andrey
  Kim}, \bibinfo{person}{Miran Kim}, {and} \bibinfo{person}{Yongsoo Song}.}
  \bibinfo{year}{2017}\natexlab{}.
\newblock \showarticletitle{Homomorphic Encryption for Arithmetic of
  Approximate Numbers}. In \bibinfo{booktitle}{\emph{ASIACRYPT}}.
\newblock


\bibitem[Cheon et~al\mbox{.}(2020)]%
        {CheonKK20}
\bibfield{author}{\bibinfo{person}{Jung~Hee Cheon}, \bibinfo{person}{Dongwoo
  Kim}, {and} \bibinfo{person}{Duhyeong Kim}.} \bibinfo{year}{2020}\natexlab{}.
\newblock \showarticletitle{Efficient Homomorphic Comparison Methods with
  Optimal Complexity}. In \bibinfo{booktitle}{\emph{ASIACRYPT}}.
\newblock


\bibitem[Cheon et~al\mbox{.}(2019)]%
        {CheonKKLL19}
\bibfield{author}{\bibinfo{person}{Jung~Hee Cheon}, \bibinfo{person}{Dongwoo
  Kim}, \bibinfo{person}{Duhyeong Kim}, \bibinfo{person}{Hun{-}Hee Lee}, {and}
  \bibinfo{person}{Keewoo Lee}.} \bibinfo{year}{2019}\natexlab{}.
\newblock \showarticletitle{Numerical Method for Comparison on Homomorphically
  Encrypted Numbers}. In \bibinfo{booktitle}{\emph{ASIACRYPT}}.
\newblock


\bibitem[Chiang et~al\mbox{.}(2020)]%
        {ChingCAKDG20}
\bibfield{author}{\bibinfo{person}{Ping-yeh Chiang}, \bibinfo{person}{Michael
  Curry}, \bibinfo{person}{Ahmed Abdelkader}, \bibinfo{person}{Aounon Kumar},
  \bibinfo{person}{John Dickerson}, {and} \bibinfo{person}{Tom Goldstein}.}
  \bibinfo{year}{2020}\natexlab{}.
\newblock \showarticletitle{{Detection as Regression: Certified Object
  Detection with Median Smoothing}}. In \bibinfo{booktitle}{\emph{NeurIPS}}.
\newblock


\bibitem[Chillotti et~al\mbox{.}(2016)]%
        {ChillottiGGI16}
\bibfield{author}{\bibinfo{person}{Ilaria Chillotti}, \bibinfo{person}{Nicolas
  Gama}, \bibinfo{person}{Mariya Georgieva}, {and} \bibinfo{person}{Malika
  Izabach{\`{e}}ne}.} \bibinfo{year}{2016}\natexlab{}.
\newblock \showarticletitle{Faster Fully Homomorphic Encryption: Bootstrapping
  in Less Than 0.1 Seconds}. In \bibinfo{booktitle}{\emph{{ASIACRYPT}}}.
\newblock


\bibitem[Chillotti et~al\mbox{.}(2021)]%
        {ChillottiJP21}
\bibfield{author}{\bibinfo{person}{Ilaria Chillotti}, \bibinfo{person}{Marc
  Joye}, {and} \bibinfo{person}{Pascal Paillier}.}
  \bibinfo{year}{2021}\natexlab{}.
\newblock \showarticletitle{Programmable Bootstrapping Enables Efficient
  Homomorphic Inference of Deep Neural Networks}. In
  \bibinfo{booktitle}{\emph{{CSCML}}}.
\newblock


\bibitem[Choquette{-}Choo et~al\mbox{.}(2021)]%
        {ChooDDZJPW21}
\bibfield{author}{\bibinfo{person}{Christopher~A. Choquette{-}Choo},
  \bibinfo{person}{Natalie Dullerud}, \bibinfo{person}{Adam Dziedzic},
  \bibinfo{person}{Yunxiang Zhang}, \bibinfo{person}{Somesh Jha},
  \bibinfo{person}{Nicolas Papernot}, {and} \bibinfo{person}{Xiao Wang}.}
  \bibinfo{year}{2021}\natexlab{}.
\newblock \showarticletitle{CaPC Learning: Confidential and Private
  Collaborative Learning}. In \bibinfo{booktitle}{\emph{arXiv}}.
\newblock


\bibitem[Chou et~al\mbox{.}(2018)]%
        {ChouBLYHF18}
\bibfield{author}{\bibinfo{person}{Edward Chou}, \bibinfo{person}{Josh Beal},
  \bibinfo{person}{Daniel Levy}, \bibinfo{person}{Serena Yeung},
  \bibinfo{person}{Albert Haque}, {and} \bibinfo{person}{Li Fei{-}Fei}.}
  \bibinfo{year}{2018}\natexlab{}.
\newblock \showarticletitle{Faster CryptoNets: Leveraging Sparsity for
  Real-World Encrypted Inference}. In \bibinfo{booktitle}{\emph{arXiv}}.
\newblock


\bibitem[Chowdhary et~al\mbox{.}(2021)]%
        {ChowdharyDLS21}
\bibfield{author}{\bibinfo{person}{Sangeeta Chowdhary}, \bibinfo{person}{Wei
  Dai}, \bibinfo{person}{Kim Laine}, {and} \bibinfo{person}{Olli Saarikivi}.}
  \bibinfo{year}{2021}\natexlab{}.
\newblock \showarticletitle{{EVA} Improved: Compiler and Extension Library for
  {CKKS}}. In \bibinfo{booktitle}{\emph{{WAHC}}}.
\newblock


\bibitem[Clopper and Pearson(1934)]%
        {ClopperPearson}
\bibfield{author}{\bibinfo{person}{C.~J. Clopper} {and} \bibinfo{person}{E.~S.
  Pearson}.} \bibinfo{year}{1934}\natexlab{}.
\newblock \showarticletitle{{The use of confidence or fiducial limits
  illustrated in the case of the binomial}}. In
  \bibinfo{booktitle}{\emph{Biometrika}}.
\newblock


\bibitem[Cohen et~al\mbox{.}(2019)]%
        {CohenRK19}
\bibfield{author}{\bibinfo{person}{Jeremy~M. Cohen}, \bibinfo{person}{Elan
  Rosenfeld}, {and} \bibinfo{person}{J.~Zico Kolter}.}
  \bibinfo{year}{2019}\natexlab{}.
\newblock \showarticletitle{Certified Adversarial Robustness via Randomized
  Smoothing}. In \bibinfo{booktitle}{\emph{{ICML}}}.
\newblock


\bibitem[Comission(2022)]%
        {EuReg}
\bibfield{author}{\bibinfo{person}{European Comission}.}
  \bibinfo{year}{2022}\natexlab{}.
\newblock \bibinfo{title}{Europe fit for the Digital Age: Commission proposes
  new rules and actions for excellence and trust in Artificial Intelligence}.
\newblock
\newblock
\newblock
\shownote{\url{https://ec.europa.eu/commission/presscorner/detail/en/IP_21_1682},
  accessed: 2022-04-15}.


\bibitem[Corbett-Davies et~al\mbox{.}(2017)]%
        {CorbettPFGH17}
\bibfield{author}{\bibinfo{person}{Sam Corbett-Davies}, \bibinfo{person}{Emma
  Pierson}, \bibinfo{person}{Avi Feller}, \bibinfo{person}{Sharad Goel}, {and}
  \bibinfo{person}{Aziz Huq}.} \bibinfo{year}{2017}\natexlab{}.
\newblock \showarticletitle{Algorithmic decision making and the cost of
  fairness}. In \bibinfo{booktitle}{\emph{ACM SIGKDD}}.
\newblock


\bibitem[Dai and Sunar(2015)]%
        {DaiS15}
\bibfield{author}{\bibinfo{person}{Wei Dai} {and} \bibinfo{person}{Berk
  Sunar}.} \bibinfo{year}{2015}\natexlab{}.
\newblock \showarticletitle{cuHE: {A} Homomorphic Encryption Accelerator
  Library}. In \bibinfo{booktitle}{\emph{{IACR} ePrint}}.
\newblock


\bibitem[Dathathri et~al\mbox{.}(2020)]%
        {DathathriKSDLM20}
\bibfield{author}{\bibinfo{person}{Roshan Dathathri},
  \bibinfo{person}{Blagovesta Kostova}, \bibinfo{person}{Olli Saarikivi},
  \bibinfo{person}{Wei Dai}, \bibinfo{person}{Kim Laine}, {and}
  \bibinfo{person}{Madan Musuvathi}.} \bibinfo{year}{2020}\natexlab{}.
\newblock \showarticletitle{{EVA: An encrypted vector arithmetic language and
  compiler for efficient homomorphic computation}}. In
  \bibinfo{booktitle}{\emph{ACM SIGPLAN}}.
\newblock


\bibitem[Dathathri et~al\mbox{.}(2019)]%
        {DathathriSCLLMMM19}
\bibfield{author}{\bibinfo{person}{Roshan Dathathri}, \bibinfo{person}{Olli
  Saarikivi}, \bibinfo{person}{Hao Chen}, \bibinfo{person}{Kim Laine},
  \bibinfo{person}{Kristin Lauter}, \bibinfo{person}{Saeed Maleki},
  \bibinfo{person}{Madan Musuvathi}, {and} \bibinfo{person}{Todd Mytkowicz}.}
  \bibinfo{year}{2019}\natexlab{}.
\newblock \showarticletitle{{CHET: An Optimizing Compiler for Fully-Homomorphic
  Neural-Network Inferencing}}. In \bibinfo{booktitle}{\emph{ACM SIGPLAN}}.
\newblock


\bibitem[Ducas and Micciancio(2015)]%
        {DucasM15}
\bibfield{author}{\bibinfo{person}{L{\'{e}}o Ducas} {and}
  \bibinfo{person}{Daniele Micciancio}.} \bibinfo{year}{2015}\natexlab{}.
\newblock \showarticletitle{{FHEW:} Bootstrapping Homomorphic Encryption in
  Less Than a Second}. In \bibinfo{booktitle}{\emph{{EUROCRYPT}}}.
\newblock


\bibitem[Dwork et~al\mbox{.}(2012)]%
        {DworkHPRZ12}
\bibfield{author}{\bibinfo{person}{Cynthia Dwork}, \bibinfo{person}{Moritz
  Hardt}, \bibinfo{person}{Toniann Pitassi}, \bibinfo{person}{Omer Reingold},
  {and} \bibinfo{person}{Richard~S. Zemel}.} \bibinfo{year}{2012}\natexlab{}.
\newblock \showarticletitle{Fairness through awareness}. In
  \bibinfo{booktitle}{\emph{{ITCS}}}.
\newblock


\bibitem[Dwork et~al\mbox{.}(2006)]%
        {DworkMNS06}
\bibfield{author}{\bibinfo{person}{Cynthia Dwork}, \bibinfo{person}{Frank
  McSherry}, \bibinfo{person}{Kobbi Nissim}, {and} \bibinfo{person}{Adam
  Smith}.} \bibinfo{year}{2006}\natexlab{}.
\newblock \showarticletitle{Calibrating Noise to Sensitivity in Private Data
  Analysis}. In \bibinfo{booktitle}{\emph{TCC}}.
\newblock


\bibitem[Engstrom et~al\mbox{.}(2019)]%
        {EngstromTTSM19}
\bibfield{author}{\bibinfo{person}{Logan Engstrom}, \bibinfo{person}{Brandon
  Tran}, \bibinfo{person}{Dimitris Tsipras}, \bibinfo{person}{Ludwig Schmidt},
  {and} \bibinfo{person}{Aleksander Madry}.} \bibinfo{year}{2019}\natexlab{}.
\newblock \showarticletitle{Exploring the Landscape of Spatial Robustness}. In
  \bibinfo{booktitle}{\emph{ICML}}.
\newblock


\bibitem[Fan and Vercauteren(2012)]%
        {FanV12}
\bibfield{author}{\bibinfo{person}{Junfeng Fan} {and} \bibinfo{person}{Frederik
  Vercauteren}.} \bibinfo{year}{2012}\natexlab{}.
\newblock \showarticletitle{Somewhat Practical Fully Homomorphic Encryption}.
  In \bibinfo{booktitle}{\emph{IACR ePrint}}.
\newblock


\bibitem[FDA(2019)]%
        {FdaReg}
\bibfield{author}{\bibinfo{person}{FDA}.} \bibinfo{year}{2019}\natexlab{}.
\newblock \bibinfo{title}{US Food \& Drug Administration, Proposed Regulatory
  Framework for Modifications to Artificial Intelligence/Machine Learning -
  [AL/ML]-Based Software as a Medical Device - [SaMD]}.
\newblock
\newblock


\bibitem[Feldmann et~al\mbox{.}(2022)]%
        {FeldmannSKDDPS22}
\bibfield{author}{\bibinfo{person}{Axel Feldmann}, \bibinfo{person}{Nikola
  Samardzic}, \bibinfo{person}{Aleksandar Krastev}, \bibinfo{person}{Srinivas
  Devadas}, \bibinfo{person}{Ronald~G. Dreslinski}, \bibinfo{person}{Chris
  Peikert}, {and} \bibinfo{person}{Daniel S{\'{a}}nchez}.}
  \bibinfo{year}{2022}\natexlab{}.
\newblock \showarticletitle{An Architecture to Accelerate Computation on
  Encrypted Data}. In \bibinfo{booktitle}{\emph{{IEEE} Micro}}.
\newblock


\bibitem[Fischer et~al\mbox{.}(2020)]%
        {FischerBV20}
\bibfield{author}{\bibinfo{person}{Marc Fischer}, \bibinfo{person}{Maximilian
  Baader}, {and} \bibinfo{person}{Martin~T. Vechev}.}
  \bibinfo{year}{2020}\natexlab{}.
\newblock \showarticletitle{Certified Defense to Image Transformations via
  Randomized Smoothing}. In \bibinfo{booktitle}{\emph{NeurIPS}}.
\newblock


\bibitem[Fischer et~al\mbox{.}(2021)]%
        {FischerBV21}
\bibfield{author}{\bibinfo{person}{Marc Fischer}, \bibinfo{person}{Maximilian
  Baader}, {and} \bibinfo{person}{Martin~T. Vechev}.}
  \bibinfo{year}{2021}\natexlab{}.
\newblock \showarticletitle{Scalable Certified Segmentation via Randomized
  Smoothing}. In \bibinfo{booktitle}{\emph{{ICML}}}.
\newblock


\bibitem[Gehr et~al\mbox{.}(2018)]%
        {GehrMDCTCV18}
\bibfield{author}{\bibinfo{person}{Timon Gehr}, \bibinfo{person}{Matthew
  Mirman}, \bibinfo{person}{Dana Drachsler-Cohen}, \bibinfo{person}{Petar
  Tsankov}, \bibinfo{person}{Swarat Chaudhuri}, {and} \bibinfo{person}{Martin
  Vechev}.} \bibinfo{year}{2018}\natexlab{}.
\newblock \showarticletitle{Ai2: Safety and robustness certification of neural
  networks with abstract interpretation}. In \bibinfo{booktitle}{\emph{IEEE
  S\&P}}.
\newblock


\bibitem[Gentry(2009)]%
        {Gentry09}
\bibfield{author}{\bibinfo{person}{Craig Gentry}.}
  \bibinfo{year}{2009}\natexlab{}.
\newblock \showarticletitle{Fully homomorphic encryption using ideal lattices}.
  In \bibinfo{booktitle}{\emph{STOC}}.
\newblock


\bibitem[Ghodsi et~al\mbox{.}(2020)]%
        {CryptoNAS}
\bibfield{author}{\bibinfo{person}{Zahra Ghodsi}, \bibinfo{person}{Akshaj~Kumar
  Veldanda}, \bibinfo{person}{Brandon Reagen}, {and} \bibinfo{person}{Siddharth
  Garg}.} \bibinfo{year}{2020}\natexlab{}.
\newblock \showarticletitle{{CryptoNAS: Private Inference on a ReLU Budget}}.
  In \bibinfo{booktitle}{\emph{NeurIPS}}.
\newblock


\bibitem[Gilad-Bachrach et~al\mbox{.}(2016)]%
        {GiladDLLNW16}
\bibfield{author}{\bibinfo{person}{Ran Gilad-Bachrach}, \bibinfo{person}{Nathan
  Dowlin}, \bibinfo{person}{Kim Laine}, \bibinfo{person}{Kristin Lauter},
  \bibinfo{person}{Michael Naehrig}, {and} \bibinfo{person}{John Wernsing}.}
  \bibinfo{year}{2016}\natexlab{}.
\newblock \showarticletitle{Cryptonets: Applying neural networks to encrypted
  data with high throughput and accuracy}. In \bibinfo{booktitle}{\emph{ICML}}.
\newblock


\bibitem[Gill(2021)]%
        {Gill21}
\bibfield{author}{\bibinfo{person}{Jagreet~Kaur Gill}.}
  \bibinfo{year}{2021}\natexlab{}.
\newblock \showarticletitle{Overview of Privacy-Preserving AI with a
  Case-Study}.
\newblock
\newblock
\shownote{\url{https://www.akira.ai/blog/privacy-preserving-ai/}, accessed:
  2022-05-02}.


\bibitem[GitHub(2020)]%
        {GithubSeal1}
\bibfield{author}{\bibinfo{person}{GitHub}.} \bibinfo{year}{2020}\natexlab{}.
\newblock \bibinfo{title}{{Microsoft SEAL Repository, Issue \#215}}.
\newblock
\newblock
\newblock
\shownote{\url{github.com/microsoft/SEAL/issues/215}, accessed: 2022-04-15}.


\bibitem[GitHub(2021a)]%
        {GithubSeal2}
\bibfield{author}{\bibinfo{person}{GitHub}.} \bibinfo{year}{2021}\natexlab{a}.
\newblock \bibinfo{title}{{Microsoft SEAL Repository, Issue \#397}}.
\newblock
\newblock
\newblock
\shownote{\url{github.com/microsoft/SEAL/issues/397}, accessed: 2022-04-15}.


\bibitem[GitHub(2021b)]%
        {GithubTenseal1}
\bibfield{author}{\bibinfo{person}{GitHub}.} \bibinfo{year}{2021}\natexlab{b}.
\newblock \bibinfo{title}{{OpenMined TenSEAL Repository, Issue \#277}}.
\newblock
\newblock
\newblock
\shownote{\url{github.com/OpenMined/TenSEAL/issues/277}, accessed: 2022-04-15}.


\bibitem[Goodfellow et~al\mbox{.}(2015)]%
        {GoodfellowSS2015}
\bibfield{author}{\bibinfo{person}{Ian Goodfellow}, \bibinfo{person}{Jonathon
  Shlens}, {and} \bibinfo{person}{Christian Szegedy}.}
  \bibinfo{year}{2015}\natexlab{}.
\newblock \showarticletitle{Explaining and Harnessing Adversarial Examples}. In
  \bibinfo{booktitle}{\emph{ICLR}}.
\newblock


\bibitem[Gowal et~al\mbox{.}(2018)]%
        {GowalDSBQJMK18}
\bibfield{author}{\bibinfo{person}{Sven Gowal}, \bibinfo{person}{Krishnamurthy
  Dvijotham}, \bibinfo{person}{Robert Stanforth}, \bibinfo{person}{Rudy Bunel},
  \bibinfo{person}{Chongli Qin}, \bibinfo{person}{Jonathan Uesato},
  \bibinfo{person}{Timothy Mann}, {and} \bibinfo{person}{Pushmeet Kohli}.}
  \bibinfo{year}{2018}\natexlab{}.
\newblock \showarticletitle{On the effectiveness of interval bound propagation
  for training verifiably robust models}. In \bibinfo{booktitle}{\emph{arXiv}}.
\newblock


\bibitem[Graepel et~al\mbox{.}(2012)]%
        {GraepelLN12}
\bibfield{author}{\bibinfo{person}{Thore Graepel}, \bibinfo{person}{Kristin~E.
  Lauter}, {and} \bibinfo{person}{Michael Naehrig}.}
  \bibinfo{year}{2012}\natexlab{}.
\newblock \showarticletitle{{ML} Confidential: Machine Learning on Encrypted
  Data}. In \bibinfo{booktitle}{\emph{{ICISC}}}.
\newblock


\bibitem[Han and Ki(2020)]%
        {HanK20}
\bibfield{author}{\bibinfo{person}{Kyoohyung Han} {and}
  \bibinfo{person}{Dohyeong Ki}.} \bibinfo{year}{2020}\natexlab{}.
\newblock \showarticletitle{Better bootstrapping for approximate homomorphic
  encryption}. In \bibinfo{booktitle}{\emph{RSA}}.
\newblock


\bibitem[Hardt et~al\mbox{.}(2016)]%
        {HardtPNS16}
\bibfield{author}{\bibinfo{person}{Moritz Hardt}, \bibinfo{person}{Eric Price},
  {and} \bibinfo{person}{Nati Srebro}.} \bibinfo{year}{2016}\natexlab{}.
\newblock \showarticletitle{Equality of Opportunity in Supervised Learning}. In
  \bibinfo{booktitle}{\emph{NeurIPS}}.
\newblock


\bibitem[Hendrycks and Dietterich(2019)]%
        {HendrycksD19}
\bibfield{author}{\bibinfo{person}{Dan Hendrycks} {and}
  \bibinfo{person}{Thomas~G. Dietterich}.} \bibinfo{year}{2019}\natexlab{}.
\newblock \showarticletitle{Benchmarking Neural Network Robustness to Common
  Corruptions and Perturbations}. In \bibinfo{booktitle}{\emph{ICLR}}.
\newblock


\bibitem[Hendrycks et~al\mbox{.}(2021)]%
        {HendrycksZBSS21}
\bibfield{author}{\bibinfo{person}{Dan Hendrycks}, \bibinfo{person}{Kevin
  Zhao}, \bibinfo{person}{Steven Basart}, \bibinfo{person}{Jacob Steinhardt},
  {and} \bibinfo{person}{Dawn Song}.} \bibinfo{year}{2021}\natexlab{}.
\newblock \showarticletitle{Natural Adversarial Examples}. In
  \bibinfo{booktitle}{\emph{CVPR}}.
\newblock


\bibitem[Iliashenko and Zucca(2021)]%
        {IliashenkoZ21}
\bibfield{author}{\bibinfo{person}{Ilia Iliashenko} {and}
  \bibinfo{person}{Vincent Zucca}.} \bibinfo{year}{2021}\natexlab{}.
\newblock \showarticletitle{Faster homomorphic comparison operations for {BGV}
  and {BFV}}. In \bibinfo{booktitle}{\emph{PoPETs}}.
\newblock


\bibitem[Ishiyama et~al\mbox{.}(2020)]%
        {IshiyamaSY20}
\bibfield{author}{\bibinfo{person}{Takumi Ishiyama}, \bibinfo{person}{Takuya
  Suzuki}, {and} \bibinfo{person}{Hayato Yamana}.}
  \bibinfo{year}{2020}\natexlab{}.
\newblock \showarticletitle{Highly Accurate {CNN} Inference Using Approximate
  Activation Functions over Homomorphic Encryption}. In
  \bibinfo{booktitle}{\emph{{IEEE} BigData}}.
\newblock


\bibitem[Jagielski et~al\mbox{.}(2020)]%
        {JagielskiCBKP20}
\bibfield{author}{\bibinfo{person}{Matthew Jagielski},
  \bibinfo{person}{Nicholas Carlini}, \bibinfo{person}{David Berthelot},
  \bibinfo{person}{Alex Kurakin}, {and} \bibinfo{person}{Nicolas Papernot}.}
  \bibinfo{year}{2020}\natexlab{}.
\newblock \showarticletitle{High Accuracy and High Fidelity Extraction of
  Neural Networks}. In \bibinfo{booktitle}{\emph{{USENIX}}}.
\newblock


\bibitem[Jha et~al\mbox{.}(2021)]%
        {deepreduce}
\bibfield{author}{\bibinfo{person}{Nandan~Kumar Jha}, \bibinfo{person}{Zahra
  Ghodsi}, \bibinfo{person}{Siddharth Garg}, {and} \bibinfo{person}{Brandon
  Reagen}.} \bibinfo{year}{2021}\natexlab{}.
\newblock \showarticletitle{DeepReDuce: ReLU Reduction for Fast Private
  Inference}. In \bibinfo{booktitle}{\emph{ICML}}.
\newblock


\bibitem[John et~al\mbox{.}(2020)]%
        {john2020verifying}
\bibfield{author}{\bibinfo{person}{Philips~George John},
  \bibinfo{person}{Deepak Vijaykeerthy}, {and} \bibinfo{person}{Diptikalyan
  Saha}.} \bibinfo{year}{2020}\natexlab{}.
\newblock \showarticletitle{Verifying Individual Fairness in Machine Learning
  Models}. In \bibinfo{booktitle}{\emph{UAI}}.
\newblock


\bibitem[Julia~Angwin and Kirchner(2016)]%
        {Angwin16}
\bibfield{author}{\bibinfo{person}{Surya~Mattu Julia~Angwin, Jeff~Larson} {and}
  \bibinfo{person}{Lauren Kirchner}.} \bibinfo{year}{2016}\natexlab{}.
\newblock \showarticletitle{Machine Bias}.
\newblock
\newblock
\shownote{\url{https://www.propublica.org/inproceedings/machine-bias-risk-assessments-in-criminal-sentencing},
  accessed: 2022-04-15}.


\bibitem[Juvekar et~al\mbox{.}(2018)]%
        {JuvekarVC18}
\bibfield{author}{\bibinfo{person}{Chiraag Juvekar}, \bibinfo{person}{Vinod
  Vaikuntanathan}, {and} \bibinfo{person}{Anantha Chandrakasan}.}
  \bibinfo{year}{2018}\natexlab{}.
\newblock \showarticletitle{GAZELLE: A Low Latency Framework for Secure Neural
  Network Inference}. In \bibinfo{booktitle}{\emph{USENIX}}.
\newblock


\bibitem[Kearns et~al\mbox{.}(2018)]%
        {KearnsNRW18}
\bibfield{author}{\bibinfo{person}{Michael~J. Kearns}, \bibinfo{person}{Seth
  Neel}, \bibinfo{person}{Aaron Roth}, {and} \bibinfo{person}{Zhiwei~Steven
  Wu}.} \bibinfo{year}{2018}\natexlab{}.
\newblock \showarticletitle{Preventing Fairness Gerrymandering: Auditing and
  Learning for Subgroup Fairness}. In \bibinfo{booktitle}{\emph{ICML}}.
\newblock


\bibitem[Kim et~al\mbox{.}(2020)]%
        {KimPP20}
\bibfield{author}{\bibinfo{person}{Andrey Kim}, \bibinfo{person}{Antonis
  Papadimitriou}, {and} \bibinfo{person}{Yuriy Polyakov}.}
  \bibinfo{year}{2020}\natexlab{}.
\newblock \showarticletitle{Approximate Homomorphic Encryption with Reduced
  Approximation Error}. In \bibinfo{booktitle}{\emph{{IACR} ePrint}}.
\newblock


\bibitem[Kleinberg et~al\mbox{.}(2017)]%
        {KleinbergMR17}
\bibfield{author}{\bibinfo{person}{Jon Kleinberg}, \bibinfo{person}{Sendhil
  Mullainathan}, {and} \bibinfo{person}{Manish Raghavan}.}
  \bibinfo{year}{2017}\natexlab{}.
\newblock \showarticletitle{Inherent Trade-Offs in the Fair Determination of
  Risk Scores}. In \bibinfo{booktitle}{\emph{ITCS}}.
\newblock


\bibitem[Kohavi(1996)]%
        {Kohavi96}
\bibfield{author}{\bibinfo{person}{Ron Kohavi}.}
  \bibinfo{year}{1996}\natexlab{}.
\newblock \showarticletitle{Scaling Up the Accuracy of Naive-Bayes Classifiers:
  {A} Decision-Tree Hybrid}. In \bibinfo{booktitle}{\emph{{KDD}}}.
\newblock


\bibitem[Krizhevsky and Hinton(2009)]%
        {Krizhevsky09}
\bibfield{author}{\bibinfo{person}{Alex Krizhevsky} {and}
  \bibinfo{person}{Geoffrey Hinton}.} \bibinfo{year}{2009}\natexlab{}.
\newblock \bibinfo{booktitle}{\emph{Learning multiple layers of features from
  tiny images}}.
\newblock \bibinfo{type}{{T}echnical {R}eport}.
\newblock


\bibitem[Kumar et~al\mbox{.}(2021)]%
        {KumarLF21}
\bibfield{author}{\bibinfo{person}{Aounon Kumar}, \bibinfo{person}{Alexander
  Levine}, {and} \bibinfo{person}{S. Feizi}.} \bibinfo{year}{2021}\natexlab{}.
\newblock \showarticletitle{Policy Smoothing for Provably Robust Reinforcement
  Learning}. In \bibinfo{booktitle}{\emph{ArXiv}}.
\newblock


\bibitem[Kurakin et~al\mbox{.}(2017)]%
        {KurakinGB2017}
\bibfield{author}{\bibinfo{person}{Alexey Kurakin}, \bibinfo{person}{Ian
  Goodfellow}, {and} \bibinfo{person}{Samy Bengio}.}
  \bibinfo{year}{2017}\natexlab{}.
\newblock \showarticletitle{Adversarial machine learning at scale}. In
  \bibinfo{booktitle}{\emph{ICLR}}.
\newblock


\bibitem[Laidlaw et~al\mbox{.}(2021)]%
        {LaidlawSF21}
\bibfield{author}{\bibinfo{person}{Cassidy Laidlaw}, \bibinfo{person}{Sahil
  Singla}, {and} \bibinfo{person}{Soheil Feizi}.}
  \bibinfo{year}{2021}\natexlab{}.
\newblock \showarticletitle{Perceptual Adversarial Robustness: Defense Against
  Unseen Threat Models}. In \bibinfo{booktitle}{\emph{ICLR}}.
\newblock


\bibitem[LeCun et~al\mbox{.}(2010)]%
        {LeCunCB10}
\bibfield{author}{\bibinfo{person}{Yann LeCun}, \bibinfo{person}{Corinna
  Cortes}, {and} \bibinfo{person}{CJ Burges}.} \bibinfo{year}{2010}\natexlab{}.
\newblock \showarticletitle{MNIST handwritten digit database}. In
  \bibinfo{booktitle}{\emph{ATT Labs}}.
\newblock


\bibitem[Lee et~al\mbox{.}(2021b)]%
        {LeeLLKKNC21}
\bibfield{author}{\bibinfo{person}{Eunsang Lee}, \bibinfo{person}{Joon{-}Woo
  Lee}, \bibinfo{person}{Junghyun Lee}, \bibinfo{person}{Young{-}Sik Kim},
  \bibinfo{person}{Yongjune Kim}, \bibinfo{person}{Jong{-}Seon No}, {and}
  \bibinfo{person}{Woosuk Choi}.} \bibinfo{year}{2021}\natexlab{b}.
\newblock \showarticletitle{Low-Complexity Deep Convolutional Neural Networks
  on Fully Homomorphic Encryption Using Multiplexed Convolutions}. In
  \bibinfo{booktitle}{\emph{{IACR} ePrint}}.
\newblock


\bibitem[Lee et~al\mbox{.}(2022)]%
        {LeeLKN22}
\bibfield{author}{\bibinfo{person}{Eunsang Lee}, \bibinfo{person}{Joon-Woo
  Lee}, \bibinfo{person}{Young-Sik Kim}, {and} \bibinfo{person}{Jong-Seon No}.}
  \bibinfo{year}{2022}\natexlab{}.
\newblock \showarticletitle{Optimization of Homomorphic Comparison Algorithm on
  RNS-CKKS Scheme}. In \bibinfo{booktitle}{\emph{IACR ePrint}}.
\newblock


\bibitem[Lee et~al\mbox{.}(2021a)]%
        {LeeKLCEDLLYKN21}
\bibfield{author}{\bibinfo{person}{Joon-Woo Lee}, \bibinfo{person}{HyungChul
  Kang}, \bibinfo{person}{Yongwoo Lee}, \bibinfo{person}{Woosuk Choi},
  \bibinfo{person}{Jieun Eom}, \bibinfo{person}{Maxim Deryabin},
  \bibinfo{person}{Eunsang Lee}, \bibinfo{person}{Junghyun Lee},
  \bibinfo{person}{Donghoon Yoo}, \bibinfo{person}{Young-Sik Kim}, {and}
  \bibinfo{person}{Jong-Seon No}.} \bibinfo{year}{2021}\natexlab{a}.
\newblock \showarticletitle{Privacy-Preserving Machine Learning with Fully
  Homomorphic Encryption for Deep Neural Network}. In
  \bibinfo{booktitle}{\emph{arXiv}}.
\newblock


\bibitem[Legislature(2018)]%
        {CCPA}
\bibfield{author}{\bibinfo{person}{California~State Legislature}.}
  \bibinfo{year}{2018}\natexlab{}.
\newblock \bibinfo{title}{California Consumer Privacy Act (CCPA)}.
\newblock
\newblock
\newblock
\shownote{\url{https://leginfo.legislature.ca.gov/faces/codes\_displayText.xhtml?division=3.\&part=4.\&lawCode=CIV\&title=1.81.5}}.


\bibitem[Li and Micciancio(2021)]%
        {Li2021}
\bibfield{author}{\bibinfo{person}{Baiyu Li} {and} \bibinfo{person}{Daniele
  Micciancio}.} \bibinfo{year}{2021}\natexlab{}.
\newblock \showarticletitle{On the Security of Homomorphic Encryption on
  Approximate Numbers}. In \bibinfo{booktitle}{\emph{EUROCRYPT}}.
\newblock


\bibitem[Liu et~al\mbox{.}(2017)]%
        {LiuJLA17}
\bibfield{author}{\bibinfo{person}{Jian Liu}, \bibinfo{person}{Mika Juuti},
  \bibinfo{person}{Yao Lu}, {and} \bibinfo{person}{N. Asokan}.}
  \bibinfo{year}{2017}\natexlab{}.
\newblock \showarticletitle{Oblivious Neural Network Predictions via MiniONN
  Transformations}. In \bibinfo{booktitle}{\emph{{ACM CCS}}}.
\newblock


\bibitem[Lou and Jiang(2019)]%
        {LouJ19}
\bibfield{author}{\bibinfo{person}{Qian Lou} {and} \bibinfo{person}{Lei
  Jiang}.} \bibinfo{year}{2019}\natexlab{}.
\newblock \showarticletitle{{SHE:} {A} Fast and Accurate Deep Neural Network
  for Encrypted Data}. In \bibinfo{booktitle}{\emph{NeurIPS}}.
\newblock


\bibitem[Lou and Jiang(2021)]%
        {LouJ21}
\bibfield{author}{\bibinfo{person}{Qian Lou} {and} \bibinfo{person}{Lei
  Jiang}.} \bibinfo{year}{2021}\natexlab{}.
\newblock \showarticletitle{{HEMET:} {A} Homomorphic-Encryption-Friendly
  Privacy-Preserving Mobile Neural Network Architecture}. In
  \bibinfo{booktitle}{\emph{ICML}}.
\newblock


\bibitem[Lou et~al\mbox{.}(2021)]%
        {SafeNET}
\bibfield{author}{\bibinfo{person}{Qian Lou}, \bibinfo{person}{Yilin Shen},
  \bibinfo{person}{Hongxia Jin}, {and} \bibinfo{person}{Lei Jiang}.}
  \bibinfo{year}{2021}\natexlab{}.
\newblock \showarticletitle{{\{}SAFEN{\}}et: A Secure, Accurate and Fast Neural
  Network Inference}. In \bibinfo{booktitle}{\emph{ICLR}}.
\newblock


\bibitem[Lu et~al\mbox{.}(2021)]%
        {LuHHMQ20}
\bibfield{author}{\bibinfo{person}{Wen{-}jie Lu}, \bibinfo{person}{Zhicong
  Huang}, \bibinfo{person}{Cheng Hong}, \bibinfo{person}{Yiping Ma}, {and}
  \bibinfo{person}{Hunter Qu}.} \bibinfo{year}{2021}\natexlab{}.
\newblock \showarticletitle{{PEGASUS:} Bridging Polynomial and Non-polynomial
  Evaluations in Homomorphic Encryption}. In \bibinfo{booktitle}{\emph{IEEE
  S\&P}}.
\newblock


\bibitem[Ma et~al\mbox{.}(2021)]%
        {Ma21}
\bibfield{author}{\bibinfo{person}{Xingjun Ma}, \bibinfo{person}{Yuhao Niu},
  \bibinfo{person}{Lin Gu}, \bibinfo{person}{Yisen Wang},
  \bibinfo{person}{Yitian Zhao}, \bibinfo{person}{James Bailey}, {and}
  \bibinfo{person}{Feng Lu}.} \bibinfo{year}{2021}\natexlab{}.
\newblock \showarticletitle{Understanding adversarial attacks on deep learning
  based medical image analysis systems}. In \bibinfo{booktitle}{\emph{Pattern
  Recognit.}}
\newblock


\bibitem[Madry et~al\mbox{.}(2018)]%
        {MadryMSTV18}
\bibfield{author}{\bibinfo{person}{Aleksander Madry},
  \bibinfo{person}{Aleksandar Makelov}, \bibinfo{person}{Ludwig Schmidt},
  \bibinfo{person}{Dimitris Tsipras}, {and} \bibinfo{person}{Adrian Vladu}.}
  \bibinfo{year}{2018}\natexlab{}.
\newblock \showarticletitle{Towards Deep Learning Models Resistant to
  Adversarial Attacks}. In \bibinfo{booktitle}{\emph{ICLR}}.
\newblock


\bibitem[Makary and Daniel(2016)]%
        {Makary16}
\bibfield{author}{\bibinfo{person}{Martin~A Makary} {and}
  \bibinfo{person}{Michael Daniel}.} \bibinfo{year}{2016}\natexlab{}.
\newblock \showarticletitle{Medical error{\textemdash}the third leading cause
  of death in the US}. In \bibinfo{booktitle}{\emph{BMJ}}.
\newblock


\bibitem[Mangaokar et~al\mbox{.}(2020)]%
        {Mangaokar20}
\bibfield{author}{\bibinfo{person}{Neal Mangaokar}, \bibinfo{person}{Jiameng
  Pu}, \bibinfo{person}{Parantapa Bhattacharya}, \bibinfo{person}{Chandan~K.
  Reddy}, {and} \bibinfo{person}{Bimal Viswanath}.}
  \bibinfo{year}{2020}\natexlab{}.
\newblock \showarticletitle{Jekyll: Attacking Medical Image Diagnostics using
  Deep Generative Models}. In \bibinfo{booktitle}{\emph{IEEE EuroSP}}.
\newblock


\bibitem[Masters et~al\mbox{.}(2019)]%
        {MastersHSCB19}
\bibfield{author}{\bibinfo{person}{Oliver Masters}, \bibinfo{person}{Hamish
  Hunt}, \bibinfo{person}{Enrico Steffinlongo}, \bibinfo{person}{Jack
  Crawford}, {and} \bibinfo{person}{Fl{\'{a}}vio Bergamaschi}.}
  \bibinfo{year}{2019}\natexlab{}.
\newblock \showarticletitle{Towards a Homomorphic Machine Learning Big Data
  Pipeline for the Financial Services Sector}. In
  \bibinfo{booktitle}{\emph{{IACR} ePrint}}.
\newblock


\bibitem[McNamara et~al\mbox{.}(2017)]%
        {McNamaraOW17}
\bibfield{author}{\bibinfo{person}{Daniel McNamara},
  \bibinfo{person}{Cheng~Soon Ong}, {and} \bibinfo{person}{Robert~C.
  Williamson}.} \bibinfo{year}{2017}\natexlab{}.
\newblock \showarticletitle{Provably Fair Representations}. In
  \bibinfo{booktitle}{\emph{arXiv}}.
\newblock


\bibitem[McNamara et~al\mbox{.}(2019)]%
        {mcnamara2019costs}
\bibfield{author}{\bibinfo{person}{Daniel McNamara},
  \bibinfo{person}{Cheng~Soon Ong}, {and} \bibinfo{person}{Robert~C.
  Williamson}.} \bibinfo{year}{2019}\natexlab{}.
\newblock \showarticletitle{Costs and Benefits of Fair Representation
  Learning}. In \bibinfo{booktitle}{\emph{AIES}}.
\newblock


\bibitem[Mishra et~al\mbox{.}(2020)]%
        {MishraLSZP20}
\bibfield{author}{\bibinfo{person}{Pratyush Mishra}, \bibinfo{person}{Ryan
  Lehmkuhl}, \bibinfo{person}{Akshayaram Srinivasan}, \bibinfo{person}{Wenting
  Zheng}, {and} \bibinfo{person}{Raluca~Ada Popa}.}
  \bibinfo{year}{2020}\natexlab{}.
\newblock \showarticletitle{Delphi: {A} Cryptographic Inference Service for
  Neural Networks}. In \bibinfo{booktitle}{\emph{{IACR} ePrint}}.
\newblock


\bibitem[Mohassel and Zhang(2017)]%
        {PaymanZ17}
\bibfield{author}{\bibinfo{person}{Payman Mohassel} {and}
  \bibinfo{person}{Yupeng Zhang}.} \bibinfo{year}{2017}\natexlab{}.
\newblock \showarticletitle{SecureML: A System for Scalable Privacy-Preserving
  Machine Learning}. In \bibinfo{booktitle}{\emph{IEEE S\&P}}.
\newblock


\bibitem[Mukherjee et~al\mbox{.}(2020)]%
        {MukherjeeYBS20}
\bibfield{author}{\bibinfo{person}{Debarghya Mukherjee},
  \bibinfo{person}{Mikhail Yurochkin}, \bibinfo{person}{Moulinath Banerjee},
  {and} \bibinfo{person}{Yuekai Sun}.} \bibinfo{year}{2020}\natexlab{}.
\newblock \showarticletitle{Two Simple Ways to Learn Individual Fairness
  Metrics from Data}. In \bibinfo{booktitle}{\emph{{ICML}}}.
\newblock


\bibitem[Papernot et~al\mbox{.}(2017)]%
        {PapernotAEGT17}
\bibfield{author}{\bibinfo{person}{Nicolas Papernot},
  \bibinfo{person}{Mart{\'{\i}}n Abadi}, \bibinfo{person}{{\'{U}}lfar
  Erlingsson}, \bibinfo{person}{Ian~J. Goodfellow}, {and}
  \bibinfo{person}{Kunal Talwar}.} \bibinfo{year}{2017}\natexlab{}.
\newblock \showarticletitle{Semi-supervised Knowledge Transfer for Deep
  Learning from Private Training Data}. In \bibinfo{booktitle}{\emph{{ICLR}}}.
\newblock


\bibitem[Parliament and Council(2016)]%
        {GDPR}
\bibfield{author}{\bibinfo{person}{European Parliament} {and}
  \bibinfo{person}{EU Council}.} \bibinfo{year}{2016}\natexlab{}.
\newblock \bibinfo{title}{The EU General Data Protection Regulation (GDPR)}.
\newblock
\newblock
\newblock
\shownote{\url{https://eur-lex.europa.eu/legal-content/EN/TXT/HTML/?uri=CELEX:32016R0679}}.


\bibitem[Paschali et~al\mbox{.}(2018)]%
        {Paschali18}
\bibfield{author}{\bibinfo{person}{Magdalini Paschali},
  \bibinfo{person}{Sailesh Conjeti}, \bibinfo{person}{Fernando Navarro}, {and}
  \bibinfo{person}{Nassir Navab}.} \bibinfo{year}{2018}\natexlab{}.
\newblock \showarticletitle{Generalizability vs. Robustness: Investigating
  Medical Imaging Networks Using Adversarial Examples}. In
  \bibinfo{booktitle}{\emph{MICCAI}}.
\newblock


\bibitem[Peychev et~al\mbox{.}(2022)]%
        {Peychev21}
\bibfield{author}{\bibinfo{person}{Momchil Peychev}, \bibinfo{person}{Anian
  Ruoss}, \bibinfo{person}{Mislav Balunovic}, \bibinfo{person}{Maximilian
  Baader}, {and} \bibinfo{person}{Martin~T. Vechev}.}
  \bibinfo{year}{2022}\natexlab{}.
\newblock \showarticletitle{Latent Space Smoothing for Individually Fair
  Representations}. In \bibinfo{booktitle}{\emph{ECCV}}.
\newblock


\bibitem[Reith et~al\mbox{.}(2019)]%
        {ReithST19}
\bibfield{author}{\bibinfo{person}{Robert~Nikolai Reith},
  \bibinfo{person}{Thomas Schneider}, {and} \bibinfo{person}{Oleksandr
  Tkachenko}.} \bibinfo{year}{2019}\natexlab{}.
\newblock \showarticletitle{Efficiently Stealing your Machine Learning Models}.
  In \bibinfo{booktitle}{\emph{WPES@CCS}}.
\newblock


\bibitem[Riazi et~al\mbox{.}(2020)]%
        {RiaziLPD20}
\bibfield{author}{\bibinfo{person}{M.~Sadegh Riazi}, \bibinfo{person}{Kim
  Laine}, \bibinfo{person}{Blake Pelton}, {and} \bibinfo{person}{Wei Dai}.}
  \bibinfo{year}{2020}\natexlab{}.
\newblock \showarticletitle{{HEAX:} An Architecture for Computing on Encrypted
  Data}. In \bibinfo{booktitle}{\emph{ASPLOS}}.
\newblock


\bibitem[Rouhani et~al\mbox{.}(2018)]%
        {RouhaniRK18}
\bibfield{author}{\bibinfo{person}{Bita~Darvish Rouhani},
  \bibinfo{person}{M.~Sadegh Riazi}, {and} \bibinfo{person}{Farinaz
  Koushanfar}.} \bibinfo{year}{2018}\natexlab{}.
\newblock \showarticletitle{Deepsecure: scalable provably-secure deep
  learning}. In \bibinfo{booktitle}{\emph{{DAC}}}.
\newblock


\bibitem[Ruoss et~al\mbox{.}(2020)]%
        {RuossBFV20}
\bibfield{author}{\bibinfo{person}{Anian Ruoss}, \bibinfo{person}{Mislav
  Balunovic}, \bibinfo{person}{Marc Fischer}, {and} \bibinfo{person}{Martin~T.
  Vechev}.} \bibinfo{year}{2020}\natexlab{}.
\newblock \showarticletitle{Learning Certified Individually Fair
  Representations}. In \bibinfo{booktitle}{\emph{NeurIPS}}.
\newblock


\bibitem[Salman et~al\mbox{.}(2019)]%
        {SalmanLRZZBY19}
\bibfield{author}{\bibinfo{person}{Hadi Salman}, \bibinfo{person}{Jerry Li},
  \bibinfo{person}{Ilya~P. Razenshteyn}, \bibinfo{person}{Pengchuan Zhang},
  \bibinfo{person}{Huan Zhang}, \bibinfo{person}{S{\'{e}}bastien Bubeck}, {and}
  \bibinfo{person}{Greg Yang}.} \bibinfo{year}{2019}\natexlab{}.
\newblock \showarticletitle{Provably Robust Deep Learning via Adversarially
  Trained Smoothed Classifiers}. In \bibinfo{booktitle}{\emph{NeurIPS}}.
\newblock


\bibitem[S{\'{e}}bert et~al\mbox{.}(2021)]%
        {SebertPZGS21}
\bibfield{author}{\bibinfo{person}{Arnaud~Grivet S{\'{e}}bert},
  \bibinfo{person}{Rafael Pinot}, \bibinfo{person}{Martin Zuber},
  \bibinfo{person}{C{\'{e}}dric Gouy{-}Pailler}, {and} \bibinfo{person}{Renaud
  Sirdey}.} \bibinfo{year}{2021}\natexlab{}.
\newblock \showarticletitle{{SPEED:} secure, PrivatE, and efficient deep
  learning}. In \bibinfo{booktitle}{\emph{ML}}.
\newblock


\bibitem[Sikeridis et~al\mbox{.}(2017)]%
        {SikeridisPRD17}
\bibfield{author}{\bibinfo{person}{Dimitrios Sikeridis},
  \bibinfo{person}{Ioannis Papapanagiotou}, \bibinfo{person}{Bhaskar~Prasad
  Rimal}, {and} \bibinfo{person}{Michael Devetsikiotis}.}
  \bibinfo{year}{2017}\natexlab{}.
\newblock \showarticletitle{A Comparative Taxonomy and Survey of Public Cloud
  Infrastructure Vendors}. In \bibinfo{booktitle}{\emph{arXiv}}.
\newblock


\bibitem[Singh et~al\mbox{.}(2019)]%
        {SinghGPV19}
\bibfield{author}{\bibinfo{person}{Gagandeep Singh}, \bibinfo{person}{Timon
  Gehr}, \bibinfo{person}{Markus P{\"u}schel}, {and} \bibinfo{person}{Martin
  Vechev}.} \bibinfo{year}{2019}\natexlab{}.
\newblock \showarticletitle{An abstract domain for certifying neural networks}.
  In \bibinfo{booktitle}{\emph{POPL}}.
\newblock


\bibitem[Szegedy et~al\mbox{.}(2013)]%
        {SzegedyZSBEGF2013}
\bibfield{author}{\bibinfo{person}{Christian Szegedy},
  \bibinfo{person}{Wojciech Zaremba}, \bibinfo{person}{Ilya Sutskever},
  \bibinfo{person}{Joan Bruna}, \bibinfo{person}{Dumitru Erhan},
  \bibinfo{person}{Ian Goodfellow}, {and} \bibinfo{person}{Rob Fergus}.}
  \bibinfo{year}{2013}\natexlab{}.
\newblock \showarticletitle{Intriguing properties of neural networks}. In
  \bibinfo{booktitle}{\emph{arXiv}}.
\newblock


\bibitem[Taori et~al\mbox{.}(2020)]%
        {TaoriDSCRS20}
\bibfield{author}{\bibinfo{person}{Rohan Taori}, \bibinfo{person}{Achal Dave},
  \bibinfo{person}{Vaishaal Shankar}, \bibinfo{person}{Nicholas Carlini},
  \bibinfo{person}{Benjamin Recht}, {and} \bibinfo{person}{Ludwig Schmidt}.}
  \bibinfo{year}{2020}\natexlab{}.
\newblock \showarticletitle{Measuring Robustness to Natural Distribution Shifts
  in Image Classification}. In \bibinfo{booktitle}{\emph{NeurIPS}}.
\newblock


\bibitem[Tatman and Kasten(2017)]%
        {TatmanK17}
\bibfield{author}{\bibinfo{person}{Rachael Tatman} {and}
  \bibinfo{person}{Conner Kasten}.} \bibinfo{year}{2017}\natexlab{}.
\newblock \showarticletitle{Effects of Talker Dialect, Gender \& Race on
  Accuracy of Bing Speech and YouTube Automatic Captions.}. In
  \bibinfo{booktitle}{\emph{INTERSPEECH}}.
\newblock


\bibitem[Tram{\`{e}}r et~al\mbox{.}(2020)]%
        {TramerCBM20}
\bibfield{author}{\bibinfo{person}{Florian Tram{\`{e}}r},
  \bibinfo{person}{Nicholas Carlini}, \bibinfo{person}{Wieland Brendel}, {and}
  \bibinfo{person}{Aleksander Madry}.} \bibinfo{year}{2020}\natexlab{}.
\newblock \showarticletitle{On Adaptive Attacks to Adversarial Example
  Defenses}. In \bibinfo{booktitle}{\emph{NeurIPS}}.
\newblock


\bibitem[Tram{\`{e}}r et~al\mbox{.}(2016)]%
        {TramerZJRR16}
\bibfield{author}{\bibinfo{person}{Florian Tram{\`{e}}r}, \bibinfo{person}{Fan
  Zhang}, \bibinfo{person}{Ari Juels}, \bibinfo{person}{Michael~K. Reiter},
  {and} \bibinfo{person}{Thomas Ristenpart}.} \bibinfo{year}{2016}\natexlab{}.
\newblock \showarticletitle{Stealing Machine Learning Models via Prediction
  APIs}. In \bibinfo{booktitle}{\emph{{USENIX}}}.
\newblock


\bibitem[Truong et~al\mbox{.}(2021)]%
        {TruongMWP20}
\bibfield{author}{\bibinfo{person}{Jean-Baptiste Truong},
  \bibinfo{person}{Pratyush Maini}, \bibinfo{person}{Robert~J. Walls}, {and}
  \bibinfo{person}{Nicolas Papernot}.} \bibinfo{year}{2021}\natexlab{}.
\newblock \showarticletitle{Data-Free Model Extraction}. In
  \bibinfo{booktitle}{\emph{CVPR}}.
\newblock


\bibitem[Turpin et~al\mbox{.}(2020)]%
        {AAMIBSI}
\bibfield{author}{\bibinfo{person}{Rob Turpin}, \bibinfo{person}{Emily Hoefer},
  \bibinfo{person}{Joe Lewelling}, {and} \bibinfo{person}{Pat Baird}.}
  \bibinfo{year}{2020}\natexlab{}.
\newblock \bibinfo{title}{Machine Learning AI in Medical Devices: Adapting
  Regulatory Frameworks and Standards to Ensure Safety and Performance}.
\newblock
\newblock


\bibitem[Viand et~al\mbox{.}(2021)]%
        {ViandJH21}
\bibfield{author}{\bibinfo{person}{Alexander Viand}, \bibinfo{person}{Patrick
  Jattke}, {and} \bibinfo{person}{Anwar Hithnawi}.}
  \bibinfo{year}{2021}\natexlab{}.
\newblock \showarticletitle{SoK: Fully Homomorphic Encryption Compilers}. In
  \bibinfo{booktitle}{\emph{IEEE S\&P}}.
\newblock


\bibitem[WA(2020)]%
        {SEALCrypto}
\bibfield{author}{\bibinfo{person}{Microsoft Research~Redmond WA}.}
  \bibinfo{year}{2020}\natexlab{}.
\newblock \bibinfo{title}{{M}icrosoft {SEAL} (release 3.6)}.
\newblock \bibinfo{howpublished}{{github.com/Microsoft/SEAL}}.
\newblock


\bibitem[Wang et~al\mbox{.}(2019)]%
        {Wang19}
\bibfield{author}{\bibinfo{person}{Guotai Wang}, \bibinfo{person}{Wenqi Li},
  \bibinfo{person}{Michael Aertsen}, \bibinfo{person}{Jan Deprest},
  \bibinfo{person}{S{\'{e}}bastien Ourselin}, {and} \bibinfo{person}{Tom
  Vercauteren}.} \bibinfo{year}{2019}\natexlab{}.
\newblock \showarticletitle{Aleatoric uncertainty estimation with test-time
  augmentation for medical image segmentation with convolutional neural
  networks}. In \bibinfo{booktitle}{\emph{Neurocomputing}}.
\newblock


\bibitem[Williams(2021)]%
        {Williams21}
\bibfield{author}{\bibinfo{person}{Ellison~Anne Williams}.}
  \bibinfo{year}{2021}\natexlab{}.
\newblock \showarticletitle{Unlocking Value With Privacy-Preserving Machine
  Learning}.
\newblock
\newblock
\shownote{\url{https://www.cpomagazine.com/data-privacy/unlocking-value-with-privacy-preserving-machine-learning/},
  accessed: 2022-05-02}.


\bibitem[Wong and Kolter(2021)]%
        {WongK21}
\bibfield{author}{\bibinfo{person}{Eric Wong} {and} \bibinfo{person}{J.~Zico
  Kolter}.} \bibinfo{year}{2021}\natexlab{}.
\newblock \showarticletitle{Learning perturbation sets for robust machine
  learning}. In \bibinfo{booktitle}{\emph{ICLR}}.
\newblock


\bibitem[Wong and Kolter(2018)]%
        {WongK18}
\bibfield{author}{\bibinfo{person}{Eric Wong} {and} \bibinfo{person}{Zico
  Kolter}.} \bibinfo{year}{2018}\natexlab{}.
\newblock \showarticletitle{Provable Defenses against Adversarial Examples via
  the Convex Outer Adversarial Polytope}. In \bibinfo{booktitle}{\emph{ICML}}.
\newblock


\bibitem[Xu et~al\mbox{.}(2020)]%
        {XuETAL20}
\bibfield{author}{\bibinfo{person}{Kaidi Xu}, \bibinfo{person}{Zhouxing Shi},
  \bibinfo{person}{Huan Zhang}, \bibinfo{person}{Yihan Wang},
  \bibinfo{person}{Kai{-}Wei Chang}, \bibinfo{person}{Minlie Huang},
  \bibinfo{person}{Bhavya Kailkhura}, \bibinfo{person}{Xue Lin}, {and}
  \bibinfo{person}{Cho{-}Jui Hsieh}.} \bibinfo{year}{2020}\natexlab{}.
\newblock \showarticletitle{Automatic Perturbation Analysis for Scalable
  Certified Robustness and Beyond}. In \bibinfo{booktitle}{\emph{NeurIPS}}.
\newblock


\bibitem[Yang et~al\mbox{.}(2020)]%
        {YangDHSRL20}
\bibfield{author}{\bibinfo{person}{Greg Yang}, \bibinfo{person}{Tony Duan},
  \bibinfo{person}{J.~Edward Hu}, \bibinfo{person}{Hadi Salman},
  \bibinfo{person}{Ilya~P. Razenshteyn}, {and} \bibinfo{person}{Jerry Li}.}
  \bibinfo{year}{2020}\natexlab{}.
\newblock \showarticletitle{Randomized Smoothing of All Shapes and Sizes}. In
  \bibinfo{booktitle}{\emph{ICML}}.
\newblock


\bibitem[Yeom and Fredrikson(2020)]%
        {yeom2020individual}
\bibfield{author}{\bibinfo{person}{Samuel Yeom} {and} \bibinfo{person}{Matt
  Fredrikson}.} \bibinfo{year}{2020}\natexlab{}.
\newblock \showarticletitle{Individual Fairness Revisited: Transferring
  Techniques from Adversarial Robustness}. In
  \bibinfo{booktitle}{\emph{IJCAI}}.
\newblock


\bibitem[Yurochkin et~al\mbox{.}(2020)]%
        {YurochkinBS20}
\bibfield{author}{\bibinfo{person}{Mikhail Yurochkin}, \bibinfo{person}{Amanda
  Bower}, {and} \bibinfo{person}{Yuekai Sun}.} \bibinfo{year}{2020}\natexlab{}.
\newblock \showarticletitle{Training individually fair {ML} models with
  sensitive subspace robustness}. In \bibinfo{booktitle}{\emph{{ICLR}}}.
\newblock


\end{thebibliography}

\message{^^JLASTREFERENCESPAGE \thepage^^J}

\ifincludeappendixx
	\clearpage
	\appendix
	\section{Argmax Approximation Details} We provide further details related to
\textsc{ArgmaxHE}, the approximation of \textsc{Argmax} used in \tool. First, in
\cref{app:params} we detail the procedure used to derive sound parameter
choices. Then, in \cref{app:reducing} we present an additional experiment that
investigates the impact of reducing the precision of the approximation (\ie
deliberately using unsound parameters) on the result. Last, in \cref{app:harm}
we discuss the idea of harmless violations, as introduced in
\cref{ssec:eval:warmup}.

\subsection{Parameter Choice Procedure} \label{app:params} In \cref{sec:error}
we derived an upper bound on the error probability of our system. In this
section, we describe a previously referenced procedure that jointly chooses the
condition parameters \mbox{$(Z_\textit{min}, Z_\textit{max}, D)$}, the \textsc{ArgmaxHE}
parameters \mbox{$(d_{q}^{(1)}$, $d_{p}^{(1)}$, $d_{q}^{(2)}$, $d_{p}^{(2)})$}
and the precision settings $(\phi^{(1)}, \psi^{(1)}, \phi^{(2)}, \psi^{(2)})$
for two \textsc{SgnHE} invocations, such that the obtained guarantees hold if
the range and difference conditions hold (up to the algorithmic error, the
probability of which is by construction at most $\alpha$).
   
\para{Choosing $Z_\textit{min}$ and $Z_\textit{max}$} To choose the range $[Z_\textit{min}, Z_\textit{max}]$
we propagate all examples $x$ from the training set through the network $f$ to
obtain the logit vectors $z = f(x)$, and keep $50$ examples with largest
$\max(z)$, $50$ with smallest $\min(z)$ and $100$ with the largest $\max(z) -
\min(z)$. On these $200$ examples we run two PGD adversarial attacks
\cite{MadryMSTV18} with $\ell_2$ radius $R=2$, and $100$ steps of size $0.01$, to
obtain adversarial examples $\hat{x}$ from $x$ that maximize (resp. minimize)
the extreme $z_i$. Next, we $n$ times sample and add Gaussian noise
$\varepsilon_i \sim \mathcal{N}(0, \sigma)$ to each $\hat{x}$ and note the
largest and the smallest observed value of $z_i \in z = f(\hat{x} +
\varepsilon_i)$. Finally, we add a small margin to the obtained result to obtain
the final values for $[Z_\textit{min}, Z_\textit{max}]$.

\para{Ensuring the \textsc{SgnHE} Input Requirement} Before proceeding, we comment on
the basic input requirement of \textsc{SgnHE}, namely that its inputs are in $[-1,
1]$. To ensure this requirement for the first invocation
(\cref{lin:argmax:sign-1} in \cref{alg:argmax}), we normalize all logits:
\begin{equation*}
    \hat{z}_i = (z_i - Z_\textit{min}) / (Z_\textit{max} - Z_\textit{min}).
\end{equation*}
In practice, this is done by modifying the weight of the last layer of the
neural network. Note that under the range condition $z_i \in [Z_\textit{min}, Z_\textit{max}]$
(\cref{sec:error}), it is ensured that the normalized logits are in $[0, 1]$,
meaning that for all $i \neq j$ it is $\hat{z}_i -\hat{z}_j \in [-1, 1]$,
satisfying the requirement of the first invocation of \textsc{SgnHE}, invoked
on logit differences. The input requirement of the second invocation
(\cref{lin:argmax:sign-2}) holds by construction, as the input $\mathit{scores}$
is guaranteed to contain values in $[-1, 1/(2c-2)]$ (and trivially $c \geq 2$).

\para{Setting $d_q^{(1)}$ and $D$} We proceed to choose the parameters of
\textsc{ArgmaxHE} used in \textsc{SgnHE} invocations. To this end, we
extensively use the Lemma 3 from~\cite{CheonKK20} to find the minimal
$d_q(\phi)$ for a given input guarantee $\phi$, where smaller $d_q$ requires a
larger $\phi$. Similarly, we use Lemma 4 from the same paper to find the minimal
$d_p(\psi)$ for the required output precision $\psi$, where for larger $\psi$ we
need a larger $d_p$.

As the first \textsc{SgnHE} is applied to logit differences we have that
$\phi^{(1)} = D / (Z_\textit{max} - Z_\textit{min})$, where $D$ is smallest absolute
difference of two logits before normalization, per the difference condition
(\cref{sec:error}). This introduces a tradeoff when choosing $d_q^{(1)}$.
Namely, reducing $d_q^{(1)}$ (to reduce multiplicative depth and reduce latency)
requires ensuring a larger $D$ and makes it more probable that the difference
condition is violated. We choose $d_q^{(1)}=6$ which implies a particular value
of $D$ in each of our applications. 

\para{Setting $d_p^{(1)}$ and $d_q^{(2)}$} The parameters $d_p^{(1)}$ and
$d_q^{(2)}$ are directly dependent on each other. For fixed $\psi^{(1)}$, after
normalization (\cref{lin:argmax:scores} in \cref{alg:argmax}) the smallest
absolute value of an element of $\mathit{scores}$, the input to second
\textsc{SgnHE}, can be calculated as
\begin{equation*}
    \phi^{(2)} = (1 - (c-1)2^{-\psi^{(1)}}) / (2c-2).
\end{equation*}
Increasing $\psi^{(1)}$ (by increasing $d_p^{(1)}$) leads to larger
$\phi^{(2)}$, allowing smaller $d_q^{(2)}$, inducing a tradeoff. For the trivial
condition $\phi^{(2)}>0$ we need $\psi^{(1)} > \log(c-1)$. Similarly, we see
that $\phi^{(2)} \leq 1/(2c-2)$. 

We could do a linear search to find the minimal $d_p^{(1)}+d_q^{(2)}$. However,
in all cases we consider the two stated conditions imply $d_p^{(1)} \geq 1$ and
$d_q^{(2)} \geq 2$, and we can see that $d_p^{(1)}=1, d_q^{(2)}=2$ are in fact
simultaneously achievable, guaranteeing \mbox{$\psi^{(1)} \approx 6$} and in
turn \mbox{$\phi^{(2)} \approx 0.05$}. However, $d_q^{(2)}=2$ supports
$\phi^{(2)} \geq 0.0423$, providing additional room for error.

\para{Setting $d_p^{(2)}$} Randomized smoothing, used to derive robustness and
fairness guarantees in \cref{sec:smoothing}, concludes by summing up $n$ outputs
of the second \textsc{SgnHE} (divided by two) and expects the client to
correctly round up the sums to the nearest integer. Therefore, we require $n \cdot 2^{-\psi^{(2)}} / 2 < 0.5$, \ie $\psi^{(2)} > \log(n)$. We can check that for
$d_p^{(2)}=1$ we can afford $n \leq 64$, but $d_p^{(2)}=2$ supports $n \geq
10^6$, which is more than sufficient for practical cases. $\psi^{(2)}$ that is
ensured by this choice is $26$, implying again that for realistic choices of $n$
there is a large margin preventing rounding errors.

\subsection{Using Less Polynomial Evaluations} \label{app:reducing}

\begin{table}[t]\centering

    \caption{MockSEAL evaluation of private inference with robustness guarantees
    on the MNIST dataset with less polynomial evaluations within the
    \textsc{SgnHE} approximation. We show consistency with the non-private
    evaluation (in \%) at count and result level.} 
    \label{table:reducing}
    
    \newcommand{\twocol}[1]{\multicolumn{2}{c}{#1}}
    \newcommand{\fivecol}[1]{\multicolumn{5}{c}{#1}}
    \renewcommand{\arraystretch}{1.2}

    \begin{tabular}{@{}ccccrr@{}} \toprule
    $d_q^{(1)}$ & $d_p^{(1)}$ & $d_q^{(2)}$ & $d_p^{(2)}$ & CountsOK & ResultOK  \\ \cmidrule{1-6}
    6 & 1 & 2 & 2 & 100.00 & 100.00 \\
    \midrule 
    6 & 1 & 2 & 1 & 100.00 & 99.99 \\
    6 & 1 & 2 & 0 & 0.00 & 0.00 \\
    6 & 1 & 1 & 2 & 5.97 & 1.89 \\
    6 & 1 & 0 & 2 & 0.00 & 0.00 \\
    6 & 0 & 2 & 2 & 0.00 & 13.92 \\
    5 & 1 & 2 & 2 & 99.93 & 99.99 \\
    5 & 1 & 2 & 1 & 99.88 & 99.98 \\
    4 & 1 & 2 & 2 & 99.20 & 99.97 \\
    4 & 1 & 2 & 1 & 98.66 & 99.92 \\
    3 & 1 & 2 & 2 & 89.63 & 99.77 \\
    3 & 1 & 2 & 1 & 87.49 & 99.52 \\
    2 & 1 & 2 & 2 & 60.93 & 97.15 \\
    2 & 1 & 2 & 1 & 58.69 & 96.02 \\
    1 & 1 & 2 & 2 & 3.00 & 25.48 \\
    1 & 1 & 2 & 1 & 2.53 & 24.85 \\
    \bottomrule  
    \end{tabular}
\end{table}
  
The parameter values determining the number of polynomial evaluations within the
\textsc{SgnHE} approximation that we describe in \appref{app:params} and use in
all our main experiments in \cref{sec:eval} are carefully chosen to ensure no
errors caused by the approximation under stated conditions, and enable rigorous
upper bounding of the error probability~(\cref{sec:error}). In this section, we
explore how reducing the number of evaluations below the theoretically sound
values affects the results of the algorithm in practice.

\para{Setup} To investigate this, we repeat the \toolR experiments with the main
MNIST model from \cref{ssec:eval:smoothing} using MockSEAL with various values
of $d_{q}^{(1)}$, $d_{p}^{(1)}$, $d_{q}^{(2)}$, $d_{p}^{(2)}$. We show the
\emph{CountsOK} and \emph{ResultOK} columns, the percentage of examples where
the FHE evaluation obtained counts consistent with the non-private evaluation,
and where the final result (``abstain'' or ``guaranteed class $k$'') was
consistent. Our baseline are the theoretically sound choices of $d_q^{(1)}=6,
d_p^{(1)}=1, d_q^{(2)}=2, d_p^{(2)}=2$ for which the result for all $100\%$ of
examples is consistent.

\para{Results} The results are presented in \cref{table:reducing}. We can notice
that reducing $d_p^{(1)}$ or $d_q^{(2)}$ is impossible without greatly
sacrificing the quality of the results. On the other hand, we see that it is
possible to reduce $d_p^{(2)}$ to $1$ (but not to $0$), as well as significantly
reduce $d_q^{(1)}$ without a large effect on the result. While the percentage of
consistent counts slowly degrades as we decrease the parameters, the percentage
of consistent final results is more slowly affected, again demonstrating the
ability of the aggregation of predictions to absorb inaccuracies. 

While it is most often crucial to have formal guarantees given by the error
probability bounds, we can conclude that when this is not the case, using
significantly smaller parameters (thus significantly reducing depth and in turn
latency) is possible in practice without greatly harming the results.     

\subsection{Harmless Violations} \label{app:harm} Here we provide more details
on the distinction between harmless and harmful violations introduced in
\cref{ssec:eval:warmup}.

Recall from \cref{sec:error} that \textsc{SgnHE} requires each logit vector $z
\in \mathbb{R}^c$ to satisfy
\begin{itemize} 
    \item $\forall i\colon\ z_i \in [Z_\textit{min}, Z_\textit{max}]\ $ (\emph{range
    condition}), and
    \item $\forall i \neq j\colon\ |z_i-z_j| \geq D\ $ (\emph{difference
    condition}),
\end{itemize}
for some chosen constants $Z_\textit{min}, Z_\textit{max}, D$.   
As long as it holds that \mbox{$\max(z) - \min(z) \leq Z_\textit{max} - Z_\textit{min}$} and
$z_1 - z_2 \geq D$, where $z_1$ and $z_2$ are the two largest logits, the result
will not be inconsistent. We call violations that satisfy these conditions
\emph{harmless}. We next illustrate this behavior on the example of difference
violations, as their analysis is more involved.

Intuitively, harmless difference violations exist as invalidating the input
requirement of the first invocation of \textsc{SgnHE} (\cref{lin:argmax:sign-1}
in \cref{alg:argmax}) can in some cases still lead to the input requirement for
the second invocation being satisfied (\cref{lin:argmax:sign-2}), implying that
the final result of \textsc{ArgmaxHE} will not be affected.

More formally, as derived in \appref{app:params}, the difference condition
implies a certain input requirement parameter $\phi^{(1)}$ for the first
\textsc{SgnHE} invocation (\cref{lin:argmax:sign-1} in \cref{alg:argmax}). From
there, we carefully derived $d_q^{(1)}$ and $\psi^{(1)}$ such that
\textsc{SgnHE} is provably $(\phi,\psi)$-close, as defined in
\cref{sec:argmaxhe}.
This allows us, assuming no violations, to reason about the values of
$\mathit{scores}$ and transform its first $c$ slots to \mbox{$[-1, \ldots,
-1/(2c-2), 1/(2c-2)]$} (\cref{lin:argmax:scores}), before invoking the second
\textsc{SgnHE} (\cref{lin:argmax:sign-2}).

If there is a harmless violation, for example $z_{c} - z_{c-1} > -D$ (where
$z_{c}$ and $z_{c-1}$ are the smallest and the second smallest logit in~$z$),
the input requirement $\phi^{(1)}$ is violated, making the $(\phi,\psi)$-close
guarantee invalid. However, while the output of the first \textsc{SgnHE} for the
entry in $\mathit{diff}$ corresponding to $z_{c} - z_{c-1}$ is not anymore
guaranteed to be close to $-1$ it will still be in the range $[-1, 0]$. Thus,
the element of $\mathit{scores}$ corresponding to $z_{c}$ will not be
approximately $-1$ (after \cref{lin:argmax:scores}) as before, but will be
provably below $-1/(2c-2)$. This implies that the behavior of the second
\textsc{SgnHE} will remain unchanged and will map that element to $-1$ with
usual approximation error. An analogous argument can be made for other
difference violations, showing that only $z_1-z_2 < D$ is a harmful difference
violation that can affect the result of \textsc{ArgmaxHE}.

\section{Inaccuracies of Soft Counting} \label{app:softcount} 
As outlined in \cref{ssec:eval:warmup}, one reason for an inconsistency between
private and non-private evaluations is the employed soft preliminary counting
heuristic. While this heuristic can cause an inconsistency by inaccurately
predicting the most likely class, it can never lead to an error, \ie an unsound
guarantee.

Recall that the \textsc{Certify} algorithm for randomized smoothing first uses a
heuristic $H$ to guess the top class $k$, which is most likely to be predicted
(see Lines 13\&14 in Algorithm 1). With high probability, \textsc{Certify} will
(i) abstain if the guess was wrong or, (ii) if the guess was correct, abstain or
return the class and a certification radius. In \tool, we use a different
heuristic $H’$, but maintain the same properties: with high probability, if the
result of $H’$ is wrong, the algorithm will abstain and else predict based on
the robustness of the underlying model. Therefore, using $H’$ instead of $H$
does not lead to unsoundness. However, it is possible for $H$ and $H’$ to pick
a different class $k$ for the same input, in which case one heuristic would
predict $k$ while the other would abstain, implying an {inconsistency}. As
we show in \cref{table:results} this is rare in practice.

\section{Full Experimental Details} \label{app:eval} Here we list all details of
our evaluation that were omitted from \cref{sec:eval}.

In all \toolR experiments we use the following parameters for \textsc{ArgmaxHE}:
\mbox{$d_q^{(1)}=6, d_p^{(1)}=1, d_q^{(2)}=2, d_p^{(2)}=2$}, use the largest
batch size supported, $B=32$, and set $\alpha=0.001$. For MNIST, we use the
randomized smoothing parameters $\sigma=0.5$, $n=128$, $n_0=32$,
$\tau=0.76=0.75+\zeta$ where $\zeta=0.01$. For CIFAR, we use $\sigma=0.25$,
$n=64$, $n_0=8$, and as before, $\tau=0.76$.
 
For the \toolF (Adult dataset) experiment presented in
\cref{ssec:eval:smoothing}, we use \mbox{$d_q^{(1)}=6, d_p^{(1)}=0, d_q^{(2)}=0,
d_p^{(2)}=2$}. Setting \mbox{$d_p^{(1)}=d_q^{(2)}=0$} is enabled by the special
case of binary classification ($c=2$) which is common in fairness problems. We
use $\alpha = 0.001$, and set the maximum $B=1024$ and \mbox{$n=n_0=1024$}
enabled by the setting (1 batch each for the preliminary and the main count).
The similarity constraint $\theta$ is manually set by inspecting the attributes,
\eg we set $\theta_{i}=2$ for the age attribute, and $\theta_{i}=1$ for the
education level. The $\Sigma$ covariance matrix obtained from $\theta$ is used
to add Gaussian noise to batches of examples during training. 
\balance
Note that we can obtain an equivalent formulation of our problem (see
\cref{eq:robustness2}) by multiplying $\Sigma$ by a constant factor $\kappa$ and
dividing $R = \sqrt{1/L}$ by the same factor, which introduces a tradeoff. While
smaller $R$ leads to smaller $\tau$ and is thus easier to certify, it introduces
$\Sigma$ with larger total variation. This in turn increases the noise that is
used in both training and certification, making it harder to obtain correct
predictions for samples used in the process of randomized smoothing. In our
implementation we rescale $\Sigma$ to total variation of $\kappa d$, where we
try several values of $\kappa$ and choose $\kappa = 0.5$, based on the
accuracy-certifiability tradeoff. This leads to $\tau = \Phi(\sqrt{1/L}) + \zeta
= 0.719$ (as $\zeta=0.01$ as before).

\fi

\message{^^JLASTPAGE \thepage^^J}

\end{document}